\documentclass[10pt,twocolumn,letterpaper]{article}

\usepackage{cvpr}
\usepackage{times}
\usepackage{epsfig}
\usepackage{amsmath}
\usepackage{amssymb}
\usepackage{graphicx}
\usepackage{mathtools}
\usepackage{color}
\usepackage{algorithm,algcompatible,lipsum}
\usepackage{csvsimple}
\usepackage{amsthm}
\usepackage{comment}
\usepackage{caption}
\usepackage{subcaption}
\usepackage{array,multirow,graphics}

\usepackage[pagebackref=true,breaklinks=true,letterpaper=true,colorlinks,bookmarks=false]{hyperref}

\newcommand{\argmin}{\arg\!\min}

\def\l{{\ell}}          
\def\L{{\mathcal{L}}}   
\def\w{{w}}             
\def\x{{x}}             
\def\X{{X}}   
\def\y{{y}}             
\def\Y{{Y}}   
\def\R{\mathbb{R}}

\def\G{{G}}
\def\c{{c}}
\def\b{{\beta}}

\graphicspath{{./figures/}}
\DeclareGraphicsExtensions{.pdf,.jpeg,.png,.jpg}

\usepackage{grffile} 

\usepackage{afterpage} 

\usepackage[keeplastbox]{flushend}  

\usepackage{enumitem}

\usepackage{float} 

\usepackage{url} 

\usepackage{color}
\usepackage[dvipsnames]{xcolor}
\definecolor{tangelo}{rgb}{0.80, 0.2, 0.0}

\newcolumntype{L}[1]{>{\raggedright\let\newline\\\arraybackslash\hspace{0pt}}m{#1}}
\newcolumntype{C}[1]{>{\centering\let\newline\\\arraybackslash\hspace{0pt}}m{#1}}
\newcolumntype{R}[1]{>{\raggedleft\let\newline\\\arraybackslash\hspace{0pt}}m{#1}}

\cvprfinalcopy 

\def\cvprPaperID{3271} 

\ifcvprfinal\fi
\begin{document}

\title{%
Block-Cyclic Stochastic Coordinate Descent for Deep Neural Networks}

\author{Kensuke Nakamura\\
Computer Science Department\\
Chung-Ang University\\
{\tt\small kensuke@image.cau.ac.kr}
\and
Stefano Soatto\\
Computer Science Department\\
University of California Los Angeles\\
{\tt\small soatto@cs.ucla.edu}
\and
Byung-Woo Hong$^*$\\
Computer Science Department\\
Chung-Ang University \\
{\tt\small hong@cau.ac.kr}
}

\maketitle
%
%
\begin{abstract}
We present a stochastic first-order optimization algorithm, named BCSC, that adds a cyclic constraint to stochastic block-coordinate descent. It uses different subsets of the data to update different subsets of the parameters, thus limiting the detrimental effect of outliers in the training set. Empirical tests in benchmark datasets show that our algorithm outperforms state-of-the-art optimization methods in both accuracy as well as convergence speed. The improvements are consistent across different architectures, and can be combined with other training techniques and regularization methods. 
\end{abstract}
%
%

{\let\thefootnote\relax\footnote{{ \small * Corresponding author.}}}

\section{Introduction}
The two workhorses of Deep Learning, responsible for much of the remarkable progress in traditionally challenging Computer Vision problems, are SGD (stochastic gradient descent) and GSD (graduate student descent). The latter has produced an ever-growing body of neural network architectures, starting from basic shallow convolutional ones~\cite{lecun1998gradient} to non-Markov ones~\cite{he2016deep,he2016identity, Balduzzi2017shattered}, and still growing deeper~\cite{hu2017squeeze,chen2017dual,huang2017densely}. The former has been the subject of intense scrutiny, despite its simplicity, both in terms of unraveling the mysteries behind its unreasonable effectiveness, as well as fostering a cottage industry of modifications and improvements. Our work is squarely in the latter vein. 

SGD~\cite{robbins1951stochastic, rumelhart1988learning, zhang2004solving} is a simple variant of classical gradient descent where the stochasticity comes from employing a random subset of the measurements (mini-batch) to compute the gradient at each step of descent. This has computational cost of $\mathcal{O}(1)$ in the total example size, that is usually in the tens of thousands to millions. It also has implicit regularization effects, making it suited for highly non-convex loss functions, such as those entailed in training deep networks for classification. 

The entire process is sensitive to outlier data such as erroneous labeling in the training set, as each mini-batch affects the update of the entire set of parameters. 
The mini-batch size is usually small, thus the relative impact of an outlier can be large compared to the full batch gradient.
There are a number of techniques such as adaptive learning rate, regularization, and some gradient descent designed for weakening the impact of outliers, but they aim to normalize the variation of mini-batches and cannot manipulate training outliers explicitly.
Stochastic methods of such as randomized block coordinate descent (SBC)~\cite{wang2014randomized, zhao2014accelerated, wan2013regularization}, on the other hand, trade off accuracy with robustness to noise.
Our objective is to develop an accurate optimization algorithm for deep learning that is not subject to such a strict tradeoff.

In the proposed algorithm, named BCSC, we leverage randomized methods based on  stochastic randomized block coordinate descent~\cite{wang2014randomized, zhao2014accelerated, wan2013regularization}, but introduce a cyclic constraint in the selection of both measurements and model parameters, so that different mini-batches of data are used to update different subsets of the unknown parameters.
We perform numerical experiments using neural networks from shallow to recently developed deeper models based on popular benchmark sets, and demonstrate that our algorithm consistently outperforms the state-of-the-art optimization techniques for all the network models under consideration.

In Sect.~\ref{sec:related_work} we place our contribution in context, and provide the problem of interest and relevant algorithms in Sect.~\ref{sec:preliminary}. The technical details on the proposed algorithm are presented in Sect.~\ref{sec:method}.  In Sect.~\ref{sec:experiments} we report experiments to compare with the state-of-the-art, and discuss limitations and potential extensions in Sect.~\ref{sec:discussion}.
%
%
\section{Related Work} \label{sec:related_work}
%
%
{\bf Adaptive step size methods}
\hspace{5pt}
In SGD, the current parameter estimate is updated by subtracting the (approximate) gradient multiplied by a factor, the {\em learning rate}. Since SGD does not converge to a point estimate, the learning rate usually decreases over iterations monotonically to reduce fluctuation. While it is still common in practice to modulate the learning rate based on a fixed schedule, several adaptive learning functions have been studied to automate the scheduling~\cite{george2006adaptive}.
Some of the best known methods include AdaGrad~\cite{duchi2011adaptive} and AdaDelta~\cite{schaul2013no, zeiler2012adadelta}. They reduce the learning rate by accumulating the gradient of the loss function globally~\cite{duchi2011adaptive} or parameter-wise~\cite{schaul2013no, zeiler2012adadelta}.
For the adaptive scheduling of the learning rate, the interpolation with a random sampling technique has been used to compute the step size~\cite{tan2016barzilai, de2016big}.
In an effort to reduce the variance of gradients, adaptively changing the mini-batch size has been introduced in~\cite{de2016big}.
Our approach is not directly aimed at acceleration, and can be used in conjunction with an adaptive step-size selection. However, as we will show empirically, it outperforms adaptive step size methods in terms of both convergence speed and overall accuracy. 
%
%

{\bf Regularization methods}
\hspace{5pt}
There are a number of ways to impose regularity to the model in order to improve generalization for better prediction, among which are data augmentation~\cite{an1996effects, simonyan2014very}, batch normalization~\cite{ioffe2015batch, hoffer2017train}, or dropout~\cite{hinton2012improving, srivastava2014dropout, wan2013regularization, huang2016deep}. 
One can also incorporate regularization in the network architectures, including pooling~\cite{krizhevsky2012imagenet}, maxout~\cite{goodfellow2013maxout}, or skip connections~\cite{long2015fully, huang2017densely}.
There is also an explicit regularization that is integrated with the objective function with classical weight decay~\cite{plaut1986experiments, lang1990dimensionality}, lasso~\cite{tibshirani1996regression}, group lasso~\cite{yuan2006model}, or Hessian~\cite{rifai2011adding}.
Our method acts in concert, not in alternative, to other forms of regularization. 
%
%

{\bf Variants of gradient descent}
\hspace{5pt}
Stochastic average gradient (SAG)~\cite{Roux2012} calculates the gradient using a randomly-chosen subset of the examples and then averages their gradients in the estimation of the full gradient.
Stochastic variance reduced gradient (SVRG)~\cite{johnson2013accelerating} considers the inherent variance of the gradient or the difference between the gradients of a mini-batch and the full gradient. 
Both SAG~\cite{Roux2012} and SVRG~\cite{johnson2013accelerating} are approximations of the standard gradient and would be subject to its same limitations in large scale optimization problems for non-convex objective functions.
A variety of first-order stochastic algorithms have been developed for parallel computation~\cite{zhang2015deep} or proximal operators~\cite{duchi2009efficient}.
Similar to SGD that randomly selects subsets of data, stochasticity has been applied to select subsets of parameters to update by randomized block coordinate descent (BCD)~\cite{nesterov2012efficiency, richtarik2014iteration}.  
Such a technique has been used to train neural networks in~\cite{liu2016parallel} utilizing parallel computation.

Our algorithm is closely related to stochastic (randomized) block coordinate descent (SBC)~\cite{wang2014randomized, zhao2014accelerated, wan2013regularization}, 
which randomly chooses both parameters and examples in the optimization procedure.
However, when the number of parameters is in the millions, there is a tradeoff between accuracy and robustness to outliers.
To mitigate this issue, we introduce a cyclic procedure such that a parameter is updated only once with each sample within an epoch.
This is, however, different from classical cyclic coordinate descent~\cite{saha2010finite}, since we consider mini-batches of both the data and the parameters. 
Furthermore, our goal is not to approximate the full gradient, as in~\cite{wang2014randomized, zhao2014accelerated}. Instead, we aim to modify the stochastic procedure to achieve faster convergence and better regularization, hence better accuracy.
%
%
\section{Preliminaries} \label{sec:preliminary}
Let $\{ (\x_1, \y_1), (\x_2, \y_2), \cdots, (\x_n, \y_n) \}$ be a set of training data where $\x_i \in \X$ is an input, typically an image, and $\y_i \in \Y$ is an output, typically a label. 
Let $h_\w : \X \rightarrow \Y$ be a prediction function with the associated model parameter $\w = ( \w_1, \w_2, \cdots, \w_m ) \in \R^m$ where the dimension of the feature space is $m$. 
The discrepancy between the predicted output $h_\w(\x_i)$ and the true output $\y_i$ is measured by a loss function $\l( h_\w(\x_i), \y_i )$ for each training sample $(\x_i, \y_i)$. 
The goal is to find optimal parameters $\w^*$ that are typically obtained by minimizing the empirical loss $\L(\w)$ on the dataset $\{ (\x_1, \y_1), (\x_2, \y_2), \cdots, (\x_n, \y_n) \}$:
\begin{align}
\L(\w) &= \frac{1}{n} \sum_{i=1}^n \l( h_\w(\x_i), \y_i) = \frac{1}{n} \sum_{i=1}^n f_i(\w), \label{eq:objective}\\
\w^* &= \argmin_\w \L(\w), \label{eq:optimal}
\end{align}
where the loss incurred by the parameter $\w$ with sample $(\x_i, \y_i)$ is denoted by $f_i(\w) \coloneqq \l( h_\w(\x_i), \y_i)$. 
%
%
\subsection{Stochastic Gradient Descent} \label{sec:sgd}
The minimization of $\L(\w)$ in Eq.~\eqref{eq:objective}, assuming  $f_i(\w)$ is differentiable, involves the computation of the gradient for a large number $n$ of training data.
Stochastic gradient descent (SGD)~\cite{robbins1951stochastic,rumelhart1988learning,zhang2004solving} achieves the dual objective of reducing the computational load as well as improving generalization due to the implicit regularization effect. 
The stochastic process of sampling subsets of data at each iteration leads to regularization in the estimation of the gradient for the expected loss.
Let $\chi = \{ 1, 2, \cdots, n \}$ be the index set of the training data and $\b \subset \chi$ be its random subset, called the mini-batch. SGD updates an initial estimate (typically random) of the weights recursively at each iteration $t$ via 
\begin{align}
\w^{(t+1)} \coloneqq \w^{(t)} - \eta^{(t)} \frac{1}{| \b^{(t)} |} \sum_{i \in \b^{(t)}} \nabla f_{i}(\w^{(t)}),
\end{align}
where $\eta^{(t)}$ is a positive scalar, called learning rate. Manual scheduling of the learning rate is typical, although adaptive scheduling schemes based on the gradient or the iteration are also considered~\cite{duchi2011adaptive, zeiler2012adadelta, schaul2013no}. 
%
%
\subsection{Random Coordinate Descent} \label{sec:rcd}
In the optimization of deep neural networks, it is often required to compute loss function $\{ f_i(\w) \}_{i = 1}^n$ with respect to a large number $m$ of parameters $\w \in \R^m$ in addition to dealing with a large number $n$ of data.
Randomized block coordinate descent (BCD)~\cite{nesterov2012efficiency,richtarik2014iteration} selects a subset $\c$ from the index set $\{ 1, 2, \cdots, m \}$ of the feature space uniformly at random and computes gradients $\nabla_{\w_{\c}} \L(\w)$ restricted to the selected subset $\w_{\c}$ of the coordinates using the set of loss functions $\{ f_i \}_{i = 1}^n$ on the whole data set.
Then, the only selected parameters $\w_\c$ are updated based on the gradient $\nabla_{\w_{\c}} \L(\w)$.
The BCD algorithm proceeds at each iteration $t$ via
\begin{align}
\w_{\c^{(t)}}^{(t+1)} \coloneqq \w_{\c^{(t)}}^{(t)} - \eta^{(t)} \frac{1}{n} \sum_{i = 1}^{n} \nabla_{\w_{\c^{(t)}}} f_{i}(\w^{(t)}).
\end{align}
%

%
\subsection{Stochastic Random Coordinate Descent} \label{sec:srcd}

It is natural to consider combining the use of random mini-batches of data as done by SGD in Sect.~\ref{sec:sgd} with random subsets of coordinates as done by BCD in Sect.~\ref{sec:rcd}. The resulting algorithm, called stochastic random block coordinate descent (SBC)~\cite{wang2014randomized,zhao2014accelerated,wan2013regularization}, selects subsets of the training data uniformly at random and computes the gradient of the objective function with respect to a randomly chosen subset of the parameters:
\begin{align}
\w_{\c^{(t)}}^{(t+1)} \coloneqq \w_{\c^{(t)}}^{(t)} - \eta^{(t)} \frac{1}{| \b^{(t)} |} \sum_{i \in \b^{(t)} } \nabla_{\w_{\c^{(t)}}} f_{i}(\w^{(t)}).
\end{align}
While it is reasonable to expect that the regularizing effects of mini-batching would compound the computational benefits of block-descent, it is less obvious that connecting the random selection so that different sets of data are used to update different set of parameters would be beneficial. Enter our proposed algorithm, which we derive in the following section.
%
%
\section{Block-Cyclic Stochastic Coordinate Descent} \label{sec:method}
The essential motivation of our proposed algorithm is to combine the two types of algorithms, SGD and BCD, in such a way that SGD is designed to feed random subsets of data in the computation of gradient and BCD is designed to select random subsets of parameters to update. 
The combination of the two stochastic processes allows to use different subsets of data to update different subsets of parameters.
We also introduce a constraint that allows the algorithm to end up using all the training example data to update each of model parameters and updating all the parameters at each epoch.
We call the resulting algorithm block-cyclic stochastic coordinate descent (BCSC), which entails a doubly-stochastic process with randomization of both mini-batches of data and parameter blocks based on the cyclic block structure.
%
%
\subsection{Cyclic Block Structure} \label{sec:block}
We model the block structure of coordinates by decomposing the feature space $\R^m$ into $M$ subspaces.
Let $P$ be a random permutation of the $m \times m$ identity matrix and $P = [ P_1 | P_2 | \cdots | P_M ]$ be a decomposition of $P$ into a set of $M$ column blocks with $P_j$ of size $m \times m_j$, where $\sum_{j = 1}^M m_j = m$. 
For a random selection of the elements from a feature vector with all the other elements being zero, we define a random selection matrix $Q_j$ with size $m \times m$ for a column block $P_j$ of the permutation matrix $P$ as follows:
$
Q_j = \left[
P_j | O_j
\right], 
$
where $O_j$ is a zero matrix with size of $m \times (m - m_j)$. For a given feature vector $\w \in \R^m$, it can be uniquely written as $\w = \sum_{j = 1}^M Q_j Q_j^T \w$.
The iterative selection of a parameter block $\w_{[j]} = P_j^T \w$ from the elements in $\w$ over $j$ considers all the elements in $\w$ exhaustively with being mutually disjoint across blocks.   
%
%
\subsection{Dual Cyclic Stochastic Process} \label{sec:dual}
In the optimization procedure, one can consider a single stochastic process in the selection of mini-batch $\b^{(t)}$ with a given cyclic block structure $P = [ P_1 | P_2 | \cdots | P_M ]$ for a random grouping of elements in parameter vector $\w$, namely random block coordinate descent (RBC) algorithms, where the same mini-batch $\b^{(t)}$ is used to update all the sequential blocks of parameters $\w_{[j]} = P_j^T \w$ in an iterative way. 
The RBC algorithm iterates over each $j$ with a fixed $t$ as follows:
\begin{align}
\G^{(t, j)} &\coloneqq \frac{1}{| \b^{(t)} |} \sum_{i \in \b^{(t)}} \nabla f_i(\w^{(t, j)}),\\
\w^{(t, j+1)} &\coloneqq \w^{(t, j)} - \eta^{(t)} Q_j^T \G^{(t, j)},
\end{align}
where $\G^{(t, j)}$ denotes the gradient of the objective function based on a mini-batch $\b^{(t)}$, and it is assumed that $\cup_t \b^{(t)}$ is the whole index set $\chi$ of the training data and mini-batches are mutually disjoint $\b^{(t)} \cap \b^{(s)} = \varnothing$ if $t \neq s$.
However, this approach is ineffective in the presence of outliers that may corrupt the estimation of the gradient for the entire set of parameters. 

The algorithm we propose, called block-cyclic stochastic coordinate descent (BCSC), is developed based on the dual cyclic stochastic process within the selection of both mini-batch from the training set and coordinate block from the parameters. It is designed to ensure that each random block $\w_{[j]} = P_j^T \w$ of the parameters $\w$ is updated following the independent stochastic selection of mini-batch $\b^{(t)}$.
In addition, each element in the training data ends up being used to update all the parameters within an epoch.
Our BCSC algorithm proceeds with the dual stochastic process to select both $\b^{(t, j)}$ and $Q_j$ as follows:
\begin{align}
\G^{(t, j)} &\coloneqq \frac{1}{| \b^{(t, j)} |} \sum_{i \in \b^{(t, j)}} \nabla f_i(\w^{(t, j)}),\\
\w^{(t, j+1)} &\coloneqq \w^{(t, j)} - \eta^{(t)} Q_j^T \G^{(t, j)}, \label{eq:bcsc_blockupdate}
\end{align}
subject to 
$\cup_t \b^{(t, j)}$ is the whole index set of the training data and $\b^{(t, j)} \cap \b^{(s, j)} = \varnothing$ if $t \neq s$ at fixed $j$.
Note that $P$ is randomly generated at every epoch and the index sets $\{ \chi_j \}_{j = 1}^M$ of the training data are also randomly shuffled at every epoch.
%
%
\subsection{Our Proposed Algorithm} \label{sec:algorithm}

The central idea of the algorithm we propose  is to use different subsets of data (mini-batches) to update different subsets (blocks) of parameters. 
This is illustrated in Fig.~\ref{fig:concept}, where the same mini-batch is used to update all the parameters in SGD (left) and different mini-batches are used to update different blocks of parameters in BCSC (right).

More details are described in Algorithm~\ref{algorithm:ours} where $M$ is a given number of partitions in the parameters.
The algorithm proceeds with the initialization for the $M$ index sets $\{ \chi_j \}_{j = 1}^M$ of training data and for the permutation matrix $P$ at each epoch. 
Then, different mini-batches $\b^{(t, j)}$ are taken from the data $\chi_j$ to update different blocks $\w_{[j]} = P_j^T \w$ of parameters $\w$ followed by the update of the index set $\chi_j$ by excluding the mini-batch $\b^{(t, j)}$ from $\chi_j$.

%
%
%
%
\def\fh{55pt}
\def\sp{5pt}
\begin{figure}[htb]
\centering
\begin{tabular}{c@{}c}
\includegraphics[height=\fh]{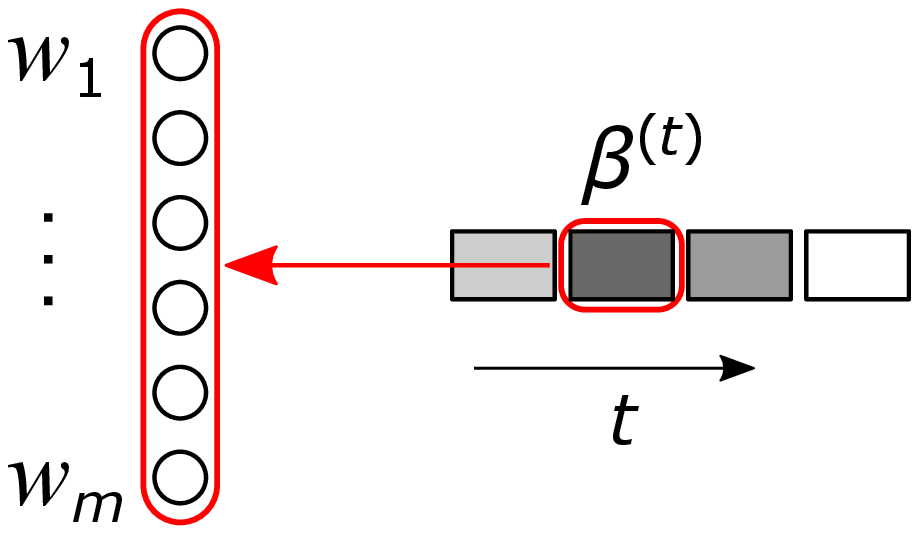} &
\includegraphics[height=\fh]{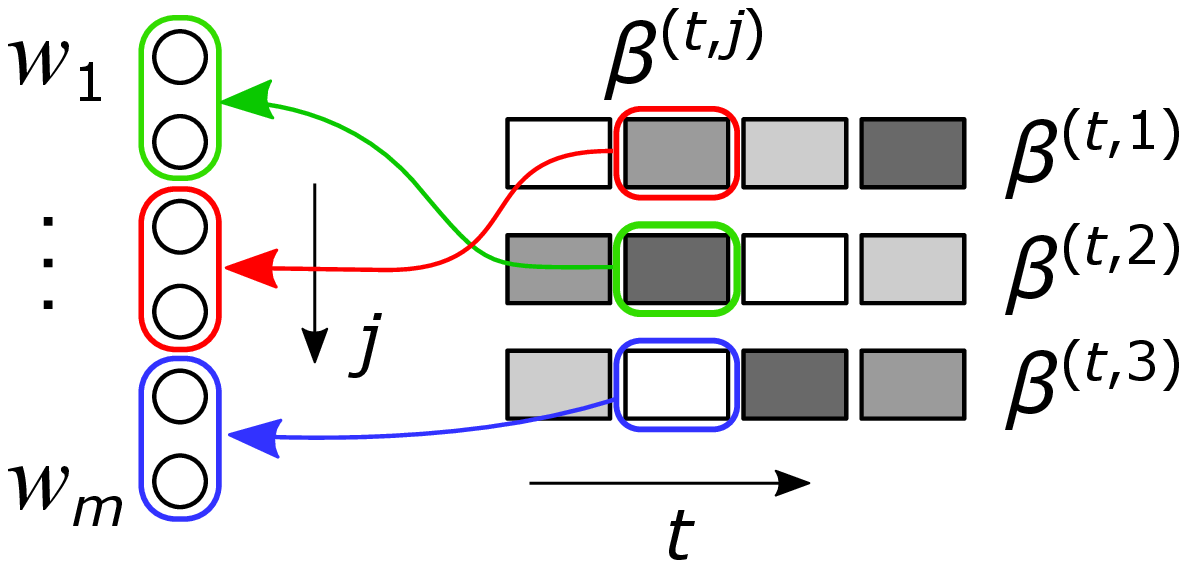} \\
SGD & BCSC (Ours)
\end{tabular}
\caption{{\bf Illustration of the proposed algorithm} SGD simultaneously updates all the parameters $(\w_1, \w_2, \cdots, \w_m)$ using the same mini-batch $\b^{(t)}$. 
On the other hand, our BCSC uses different mini-batches $\b(t, j)$ to update different blocks $\w_{[j]} = P_j^T w$ of parameters $\w$.}
\label{fig:concept}
\end{figure}
%
%

%
%
%
\begin{algorithm}[htb]
\caption{\small Block-Cyclic Stochastic Coordinate Descent}
\label{algorithm:ours}
\begin{algorithmic}
\small
	\FORALL {epoch}
      \STATE $\{ \chi_j \}_{j=1}^M$ : $M$ index sets $\chi_j$ of data by random shuffling.
      \STATE $P = [P_1 | P_2 | \cdots | P_M]$ : random permutation matrix.
      \FORALL {$t$ : index for mini-batch} 
        \FORALL {$j$ : index for parameter block}
            \STATE Take mini-batch $\b^{(t, j)}$ from $\chi_j$. 
            \STATE Take parameter block $\w_{[j]} = P_j^T \w$ using $P_j$. 
            \STATE Compute gradient of the loss to $\w_{[j]}$ using $\{f_i\}_{i \in \b^{(t, j)}}$.
            \STATE Update parameter block $\w_{[j]}$ using Eq.~\eqref{eq:bcsc_blockupdate}.
            \STATE Update index set $\chi_j \coloneqq \chi_j \setminus \b^{(t, j)}$.
        \ENDFOR
      \ENDFOR
    \ENDFOR
\end{algorithmic}
\end{algorithm}
%
%
%
%
\section{Experimental results} \label{sec:experiments}
%
We provide quantitative and qualitative evaluation of our algorithm in comparison to the state-of-the-art optimization algorithms on the datasets including MNIST~\cite{lecun1998gradient}, Cifar10~\cite{krizhevsky2009learning} and Cifar100~\cite{krizhevsky2009learning}. MNIST consists of $60,000$ training and $10,000$ testing images with $10$ labels. Cifar10 and Cifar100 are more challenging datasets that consist of $50,000$ training and $10,000$ test data with $10$ and $100$ labels, respectively.

In order to provide better understanding on the effectiveness and robustness of our algorithm, we consider a variety of neural networks raging from simple to deep and wide models; LeNet4~\cite{lecun1998gradient}, VGG19~\cite{simonyan2014very}, GoogLeNet~\cite{szegedy2015going}, ResNet18~\cite{he2016deep,he2016identity}, ResNeXt29~\cite{xie2016aggregated}, MobileNet~\cite{howard2017mobilenets}, ShuffleNet~\cite{zhang2017shufflenet}, SENet18~\cite{hu2017squeeze}, DPN92~\cite{chen2017dual}, and DenseConv~\cite{huang2017densely}.

The performance of our BCSC algorithm is compared with other state-of-the-art optimization algorithms including AdaGrad (AG)~\cite{duchi2011adaptive}, AdaDelta (AD)~\cite{zeiler2012adadelta, schaul2013no}, stochastic gradient descent (SGD), stochastic randomized block-coordinate descent (SBC)~\cite{wang2014randomized, zhao2014accelerated,wan2013regularization}, and randomized block-coordinate descent (RBC) in Sect.~\ref{sec:dual}.
For each experiment we provide the learning curves that consist of the training loss, the test loss, and the test accuracy. In addition, the standard deviation of the training loss computed from the mini-batches within each epoch is also presented.
The learning curves are shown in colors; training loss in blue, test loss in red, and test accuracy in green, and they are plotted in log scale. 
The percentile loss and the percentile accuracy are displayed with respect to the left vertical axis and the percentile accuracy is displayed with respect to the right vertical axis. 
As quantitative comparison, the test accuracy is computed within the first half epochs, the last half epochs, all the epochs, and the final epoch.

For the selection of the hyper-parameters associated with the optimization algorithms, we use the customary values; mini-batch size is $128$, momentum is $0.9$, weight decay is $5 \times 10^{-4}$, and the total number of epochs is $200$. 
For the learning rate, we employ a manual scheduling that is known to be effective; $\eta = 0.1$ for epochs $1-100$, $0.01$ for epochs $101-150$, and $0.001$ for epochs $151-200$, so that the staircase effect appears in the learning curve in which it is noted that the horizontal axis for epoch iteration is in log-scale.
These values are applied to all the algorithms throughout the experiments unless mentioned otherwise. 
For a fair comparison, the same values are used for the common hyper-parameters among the algorithms.
%
%
%
%
%
\def\fw{75pt}
\def\sp{5pt}
\begin{figure*}[htb]
\centering
\begin{tabular}{c@{}c@{}c@{}c@{}c}
\small
\includegraphics[height=\fw]{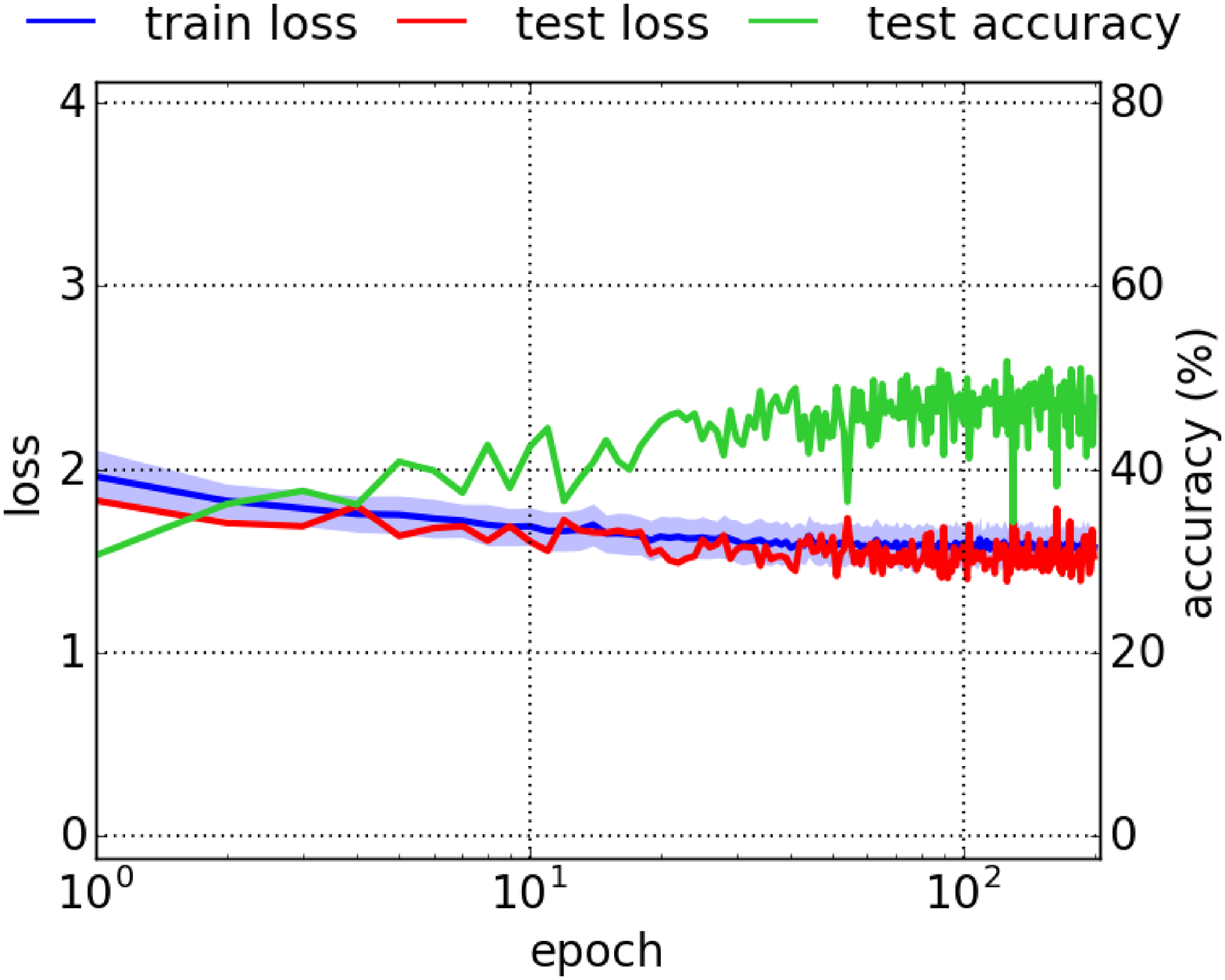} &
\includegraphics[height=\fw]{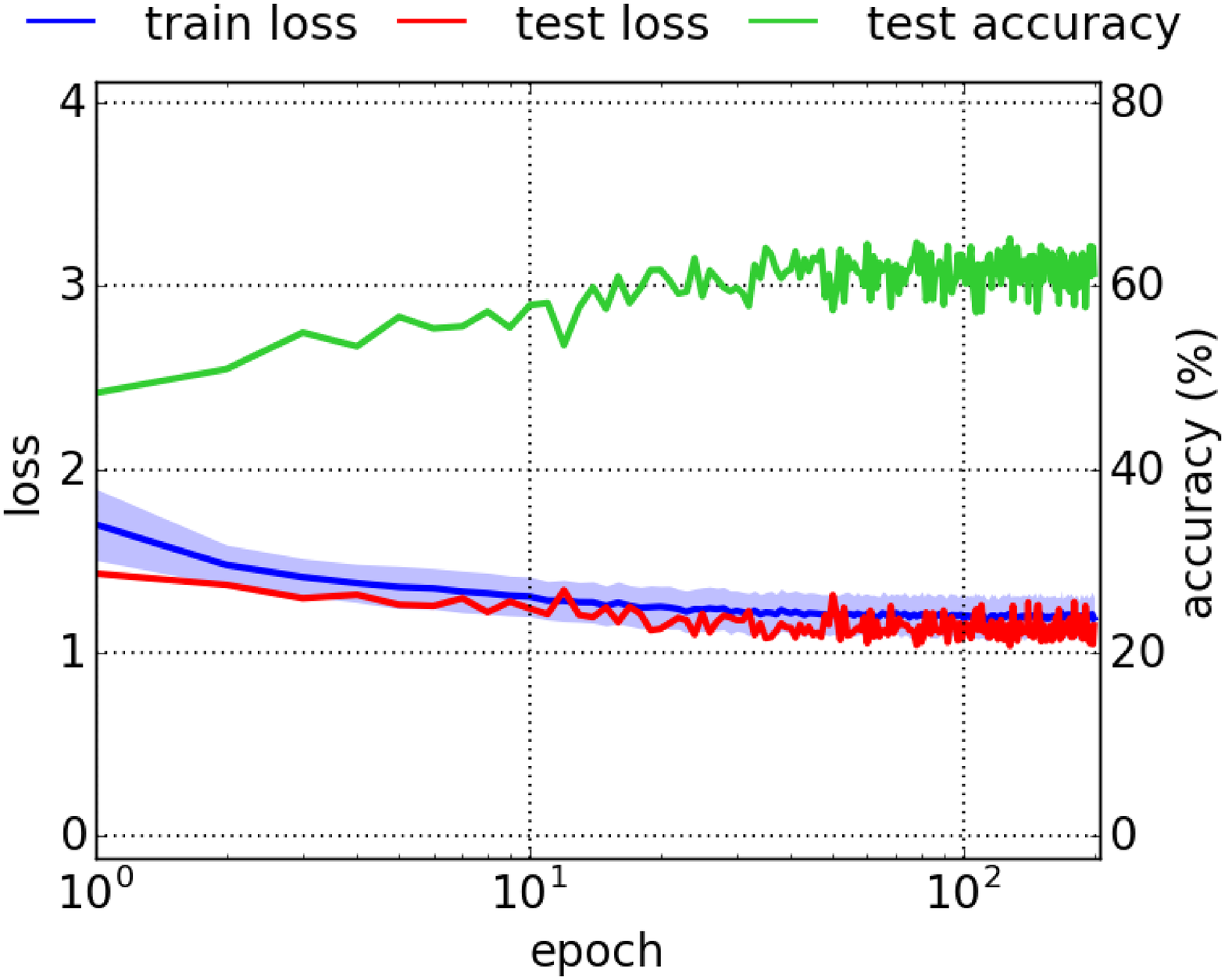} &
\includegraphics[height=\fw]{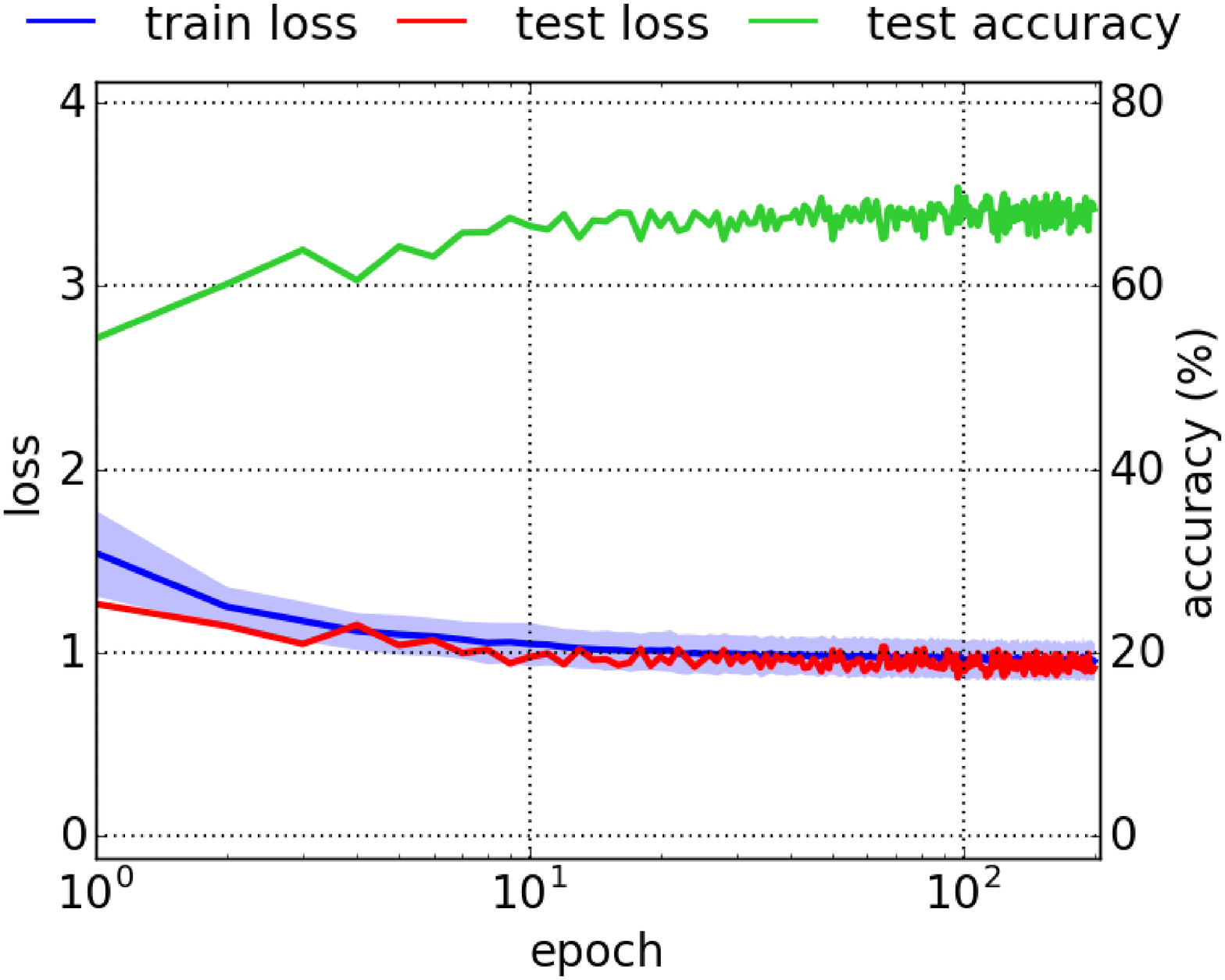} &
\includegraphics[height=\fw]{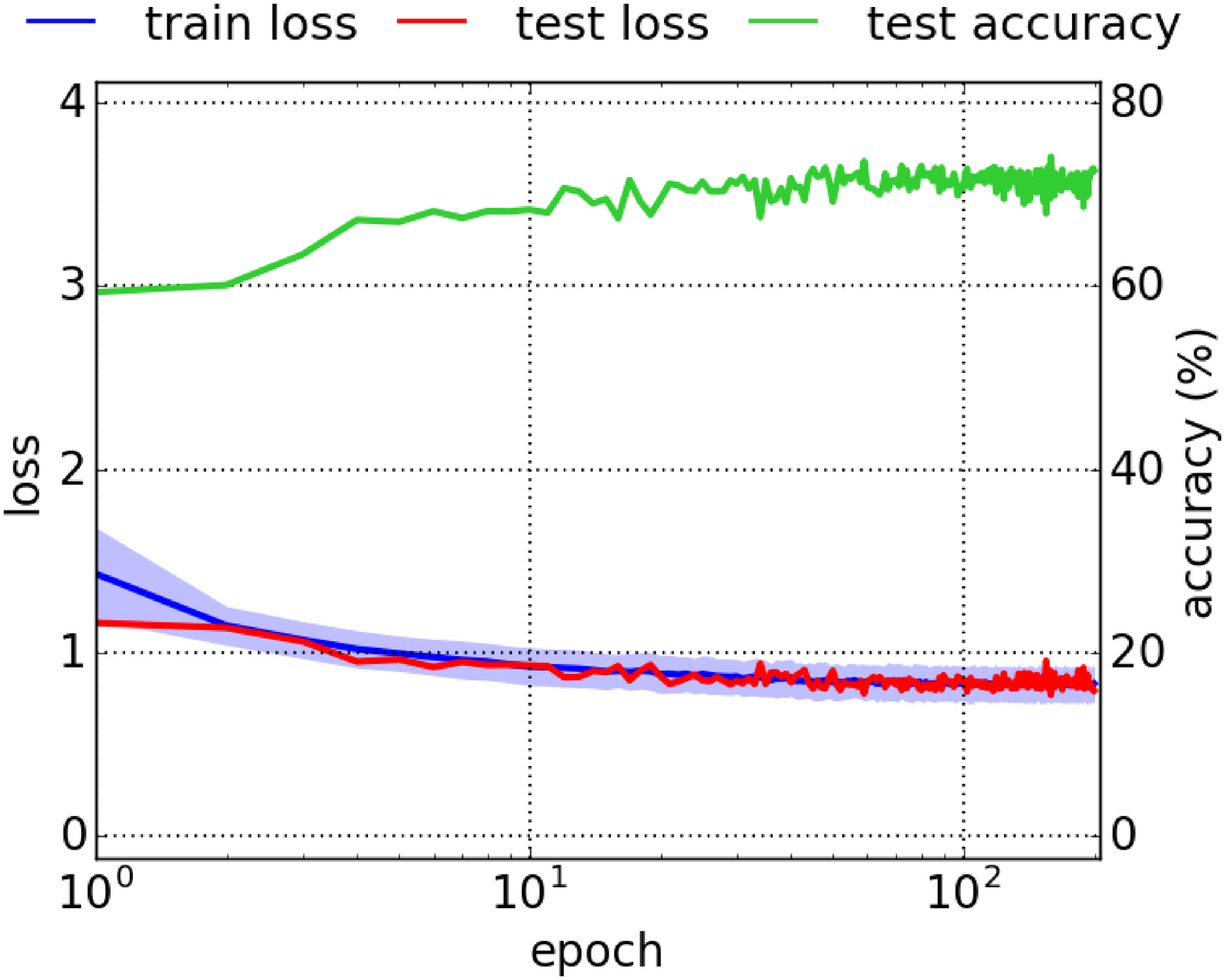} &
\includegraphics[height=\fw]{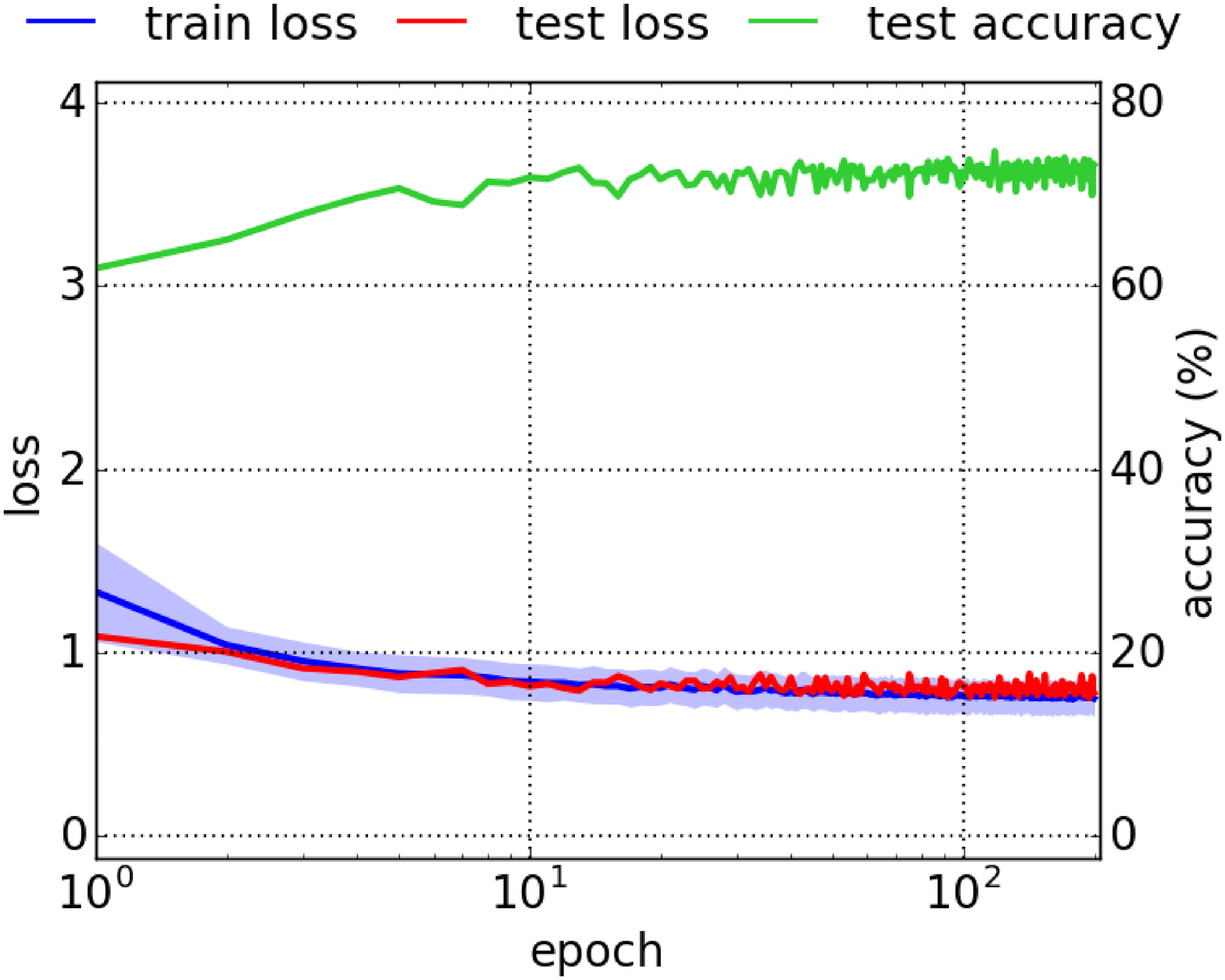} \\
$M=1$ (SGD) & $M=2$ & $M=4$ & $M=8$ & $M=16$ \\
\multicolumn{5}{c}{(a) LeNet4~\cite{lecun1998gradient}} \\
\\
\includegraphics[height=\fw]{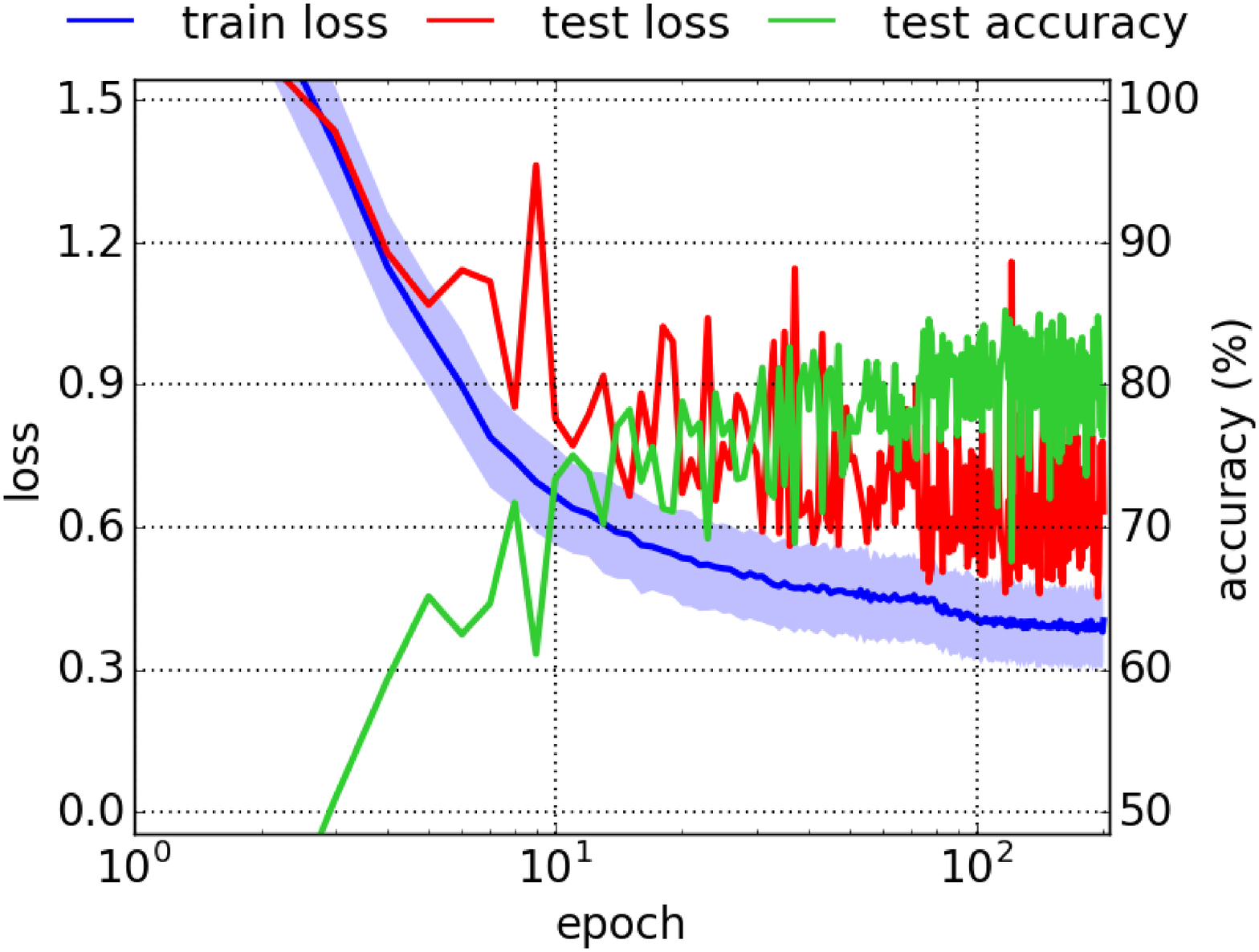} &
\includegraphics[height=\fw]{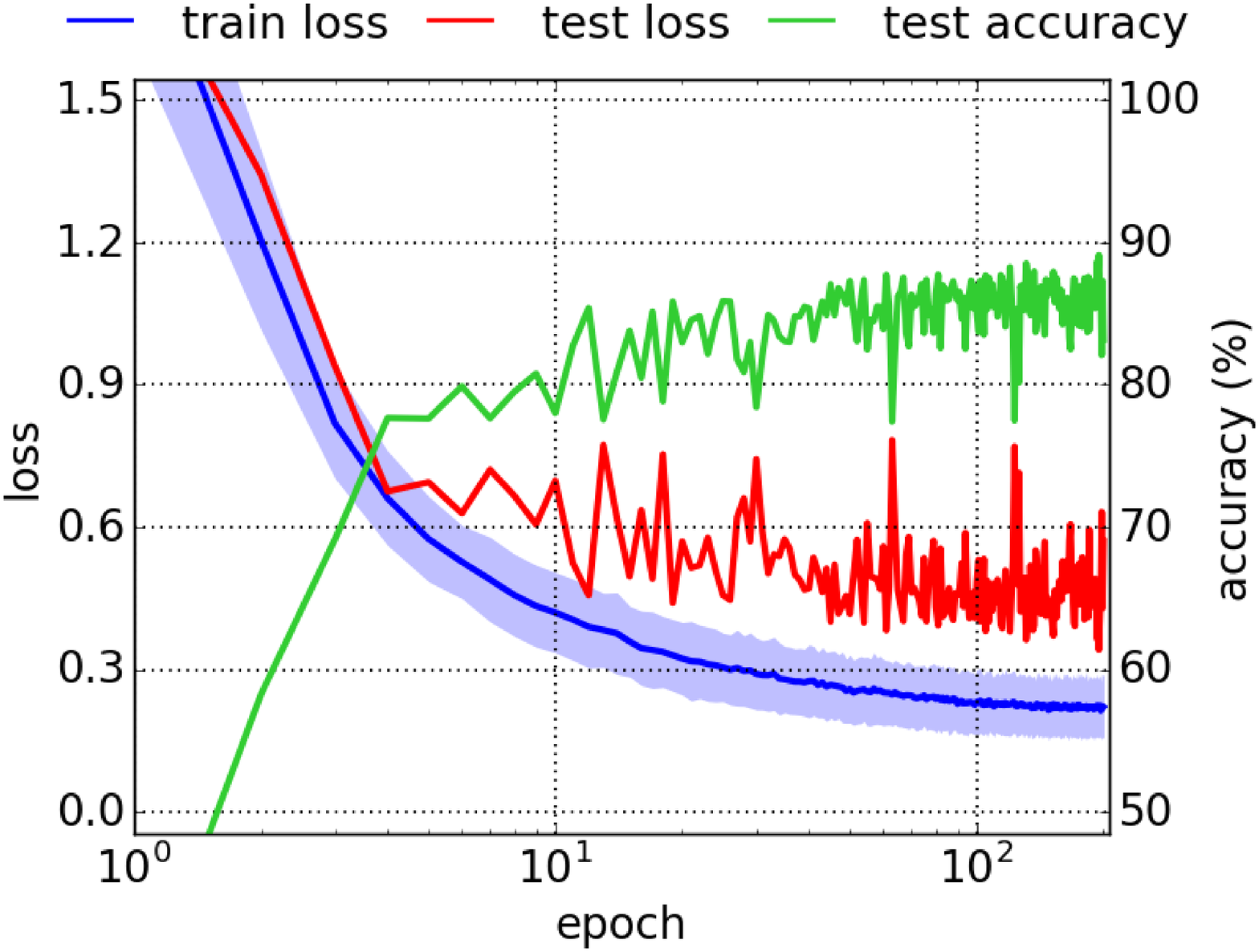} &
\includegraphics[height=\fw]{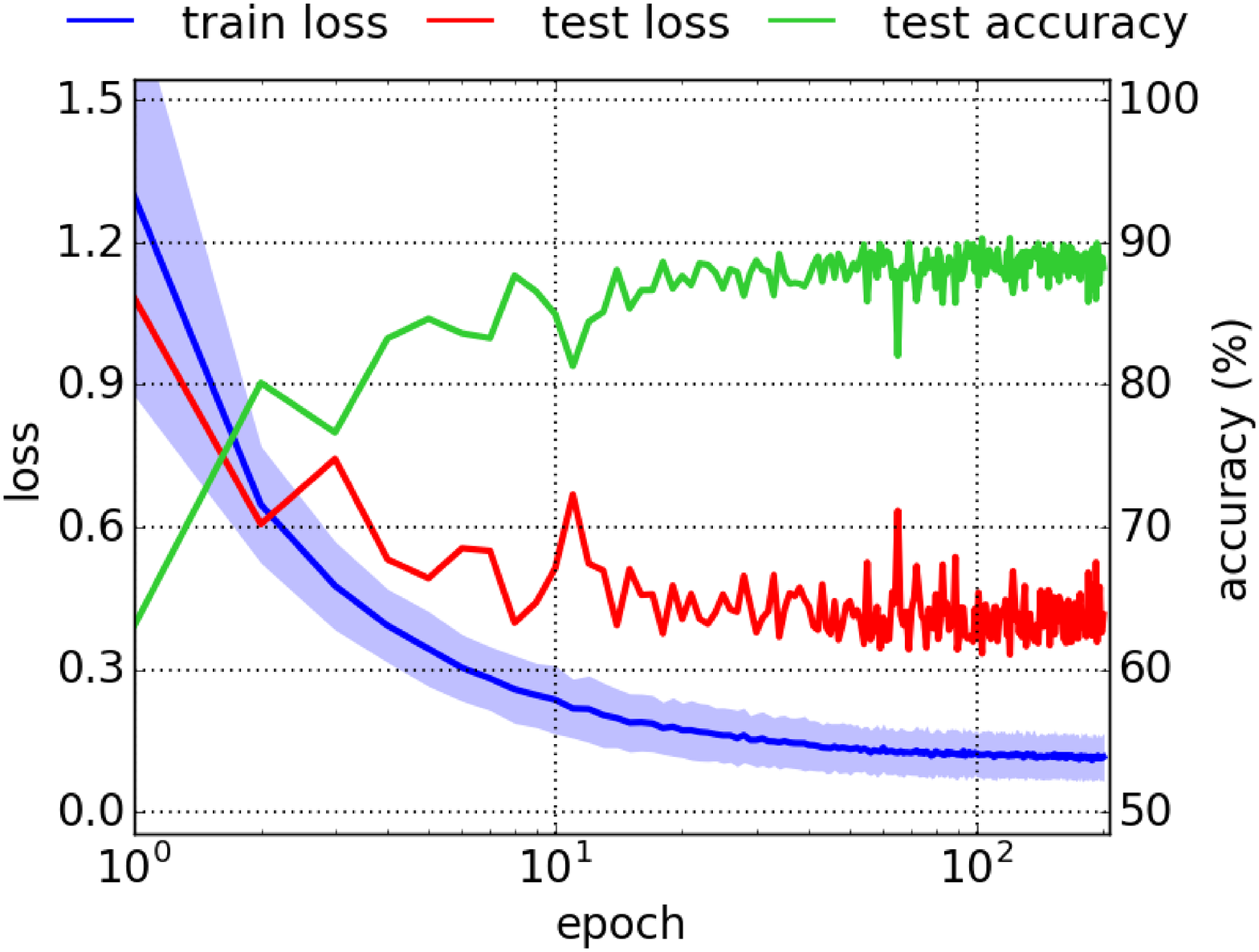} &
\includegraphics[height=\fw]{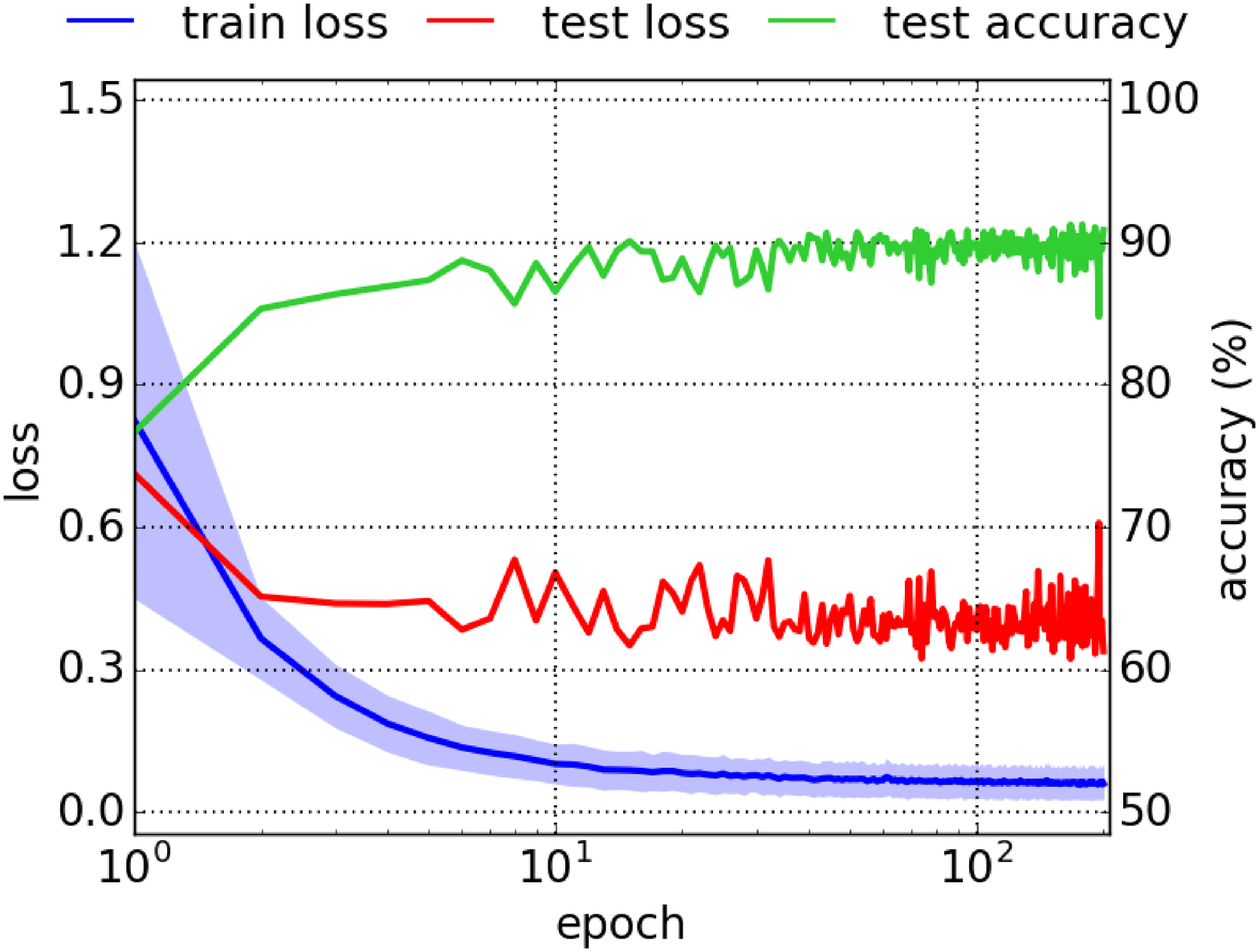} &
\includegraphics[height=\fw]{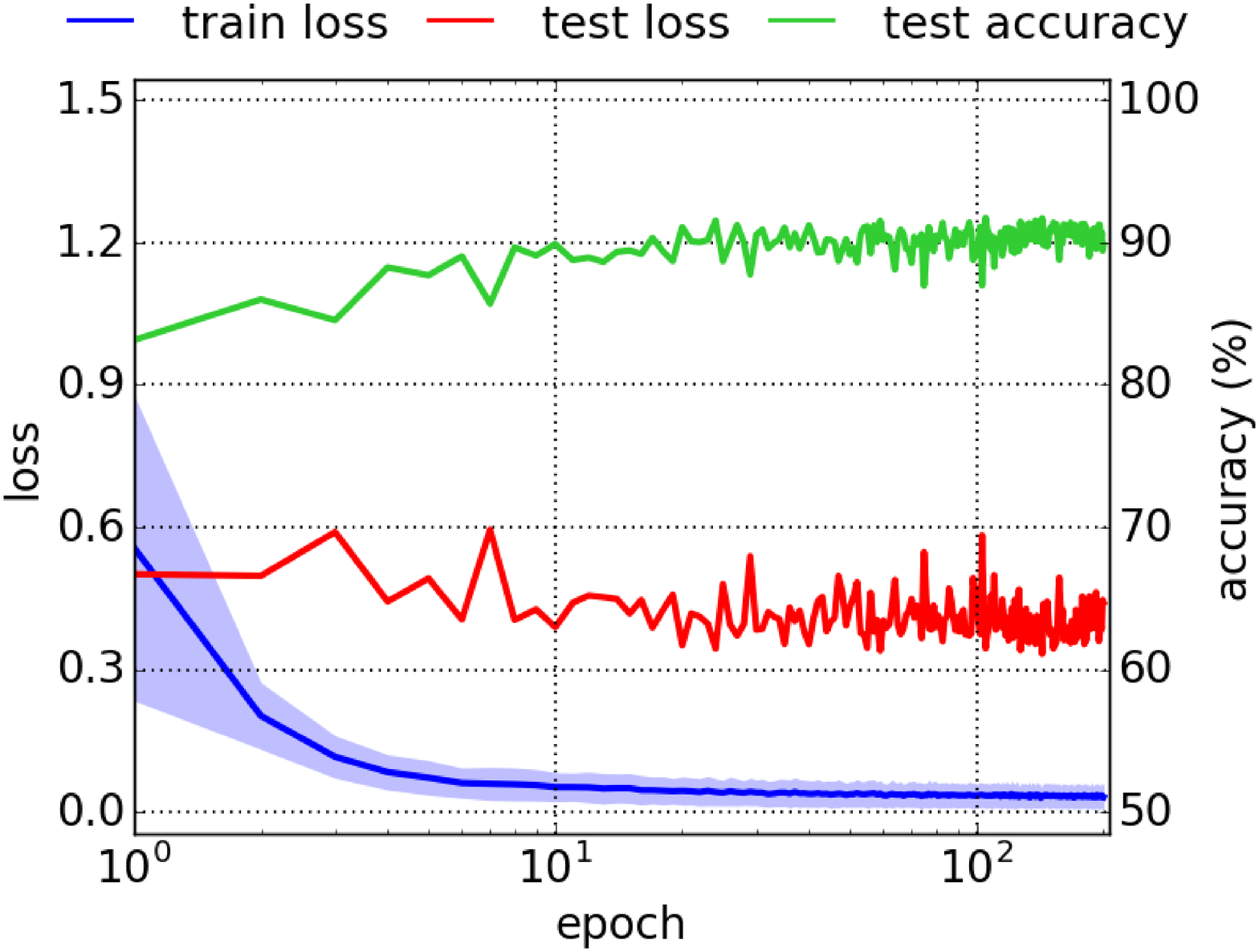}  \\
$M=1$ (SGD) & $M=2$ & $M=4$ & $M=8$ & $M=16$ \\
\multicolumn{5}{c}{(b) VGG19~\cite{simonyan2014very}} \\
\\
\includegraphics[height=\fw]{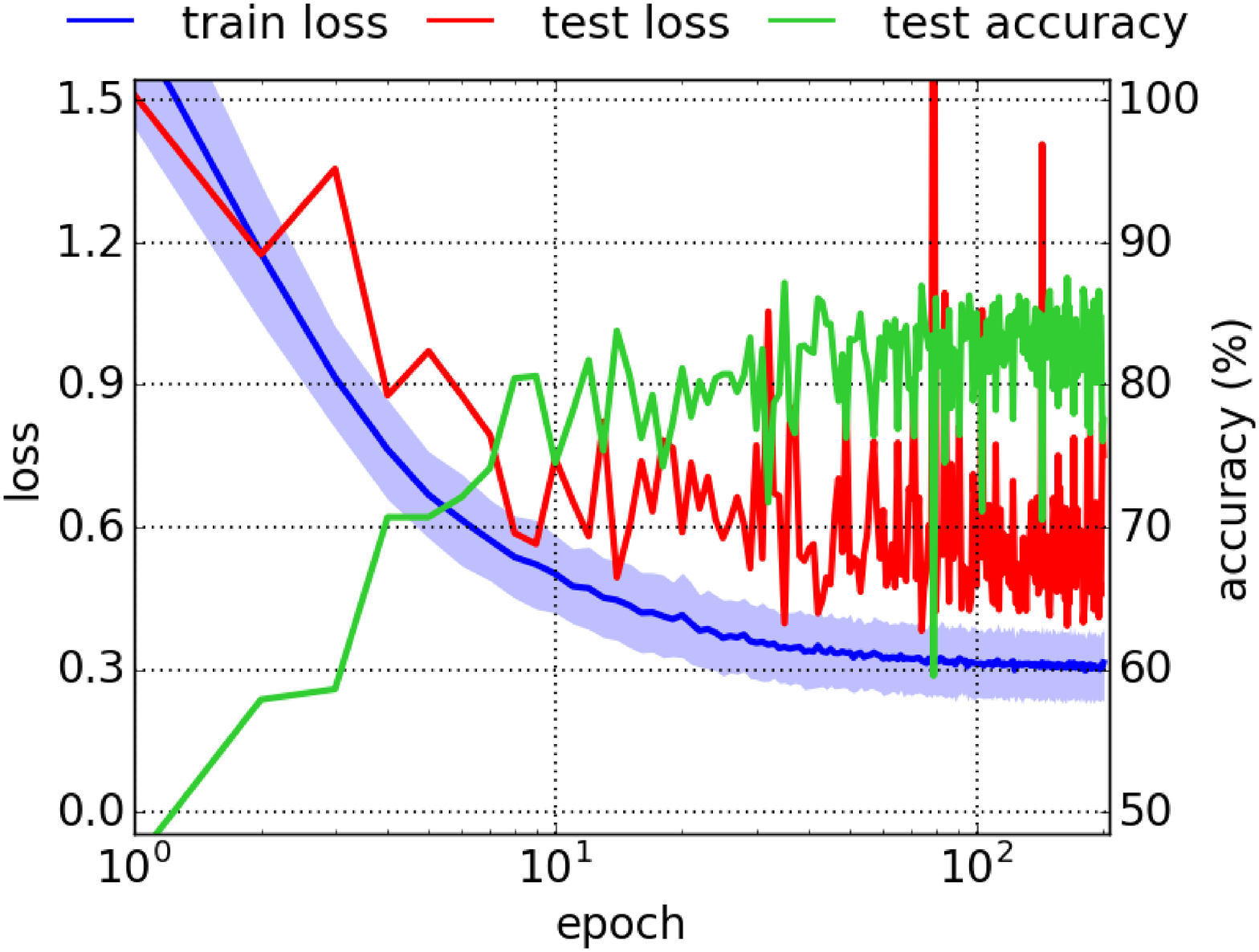} &
\includegraphics[height=\fw]{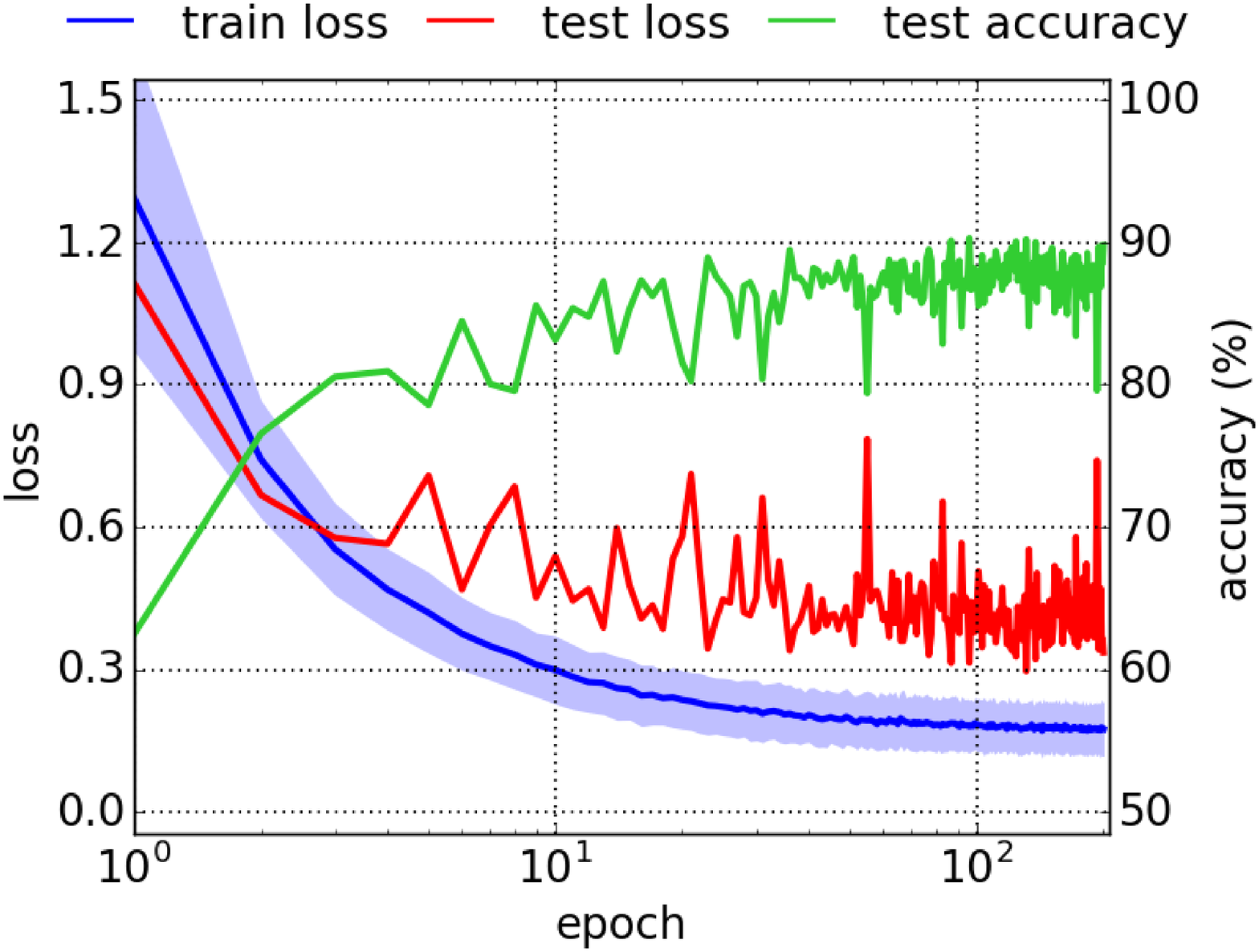} &
\includegraphics[height=\fw]{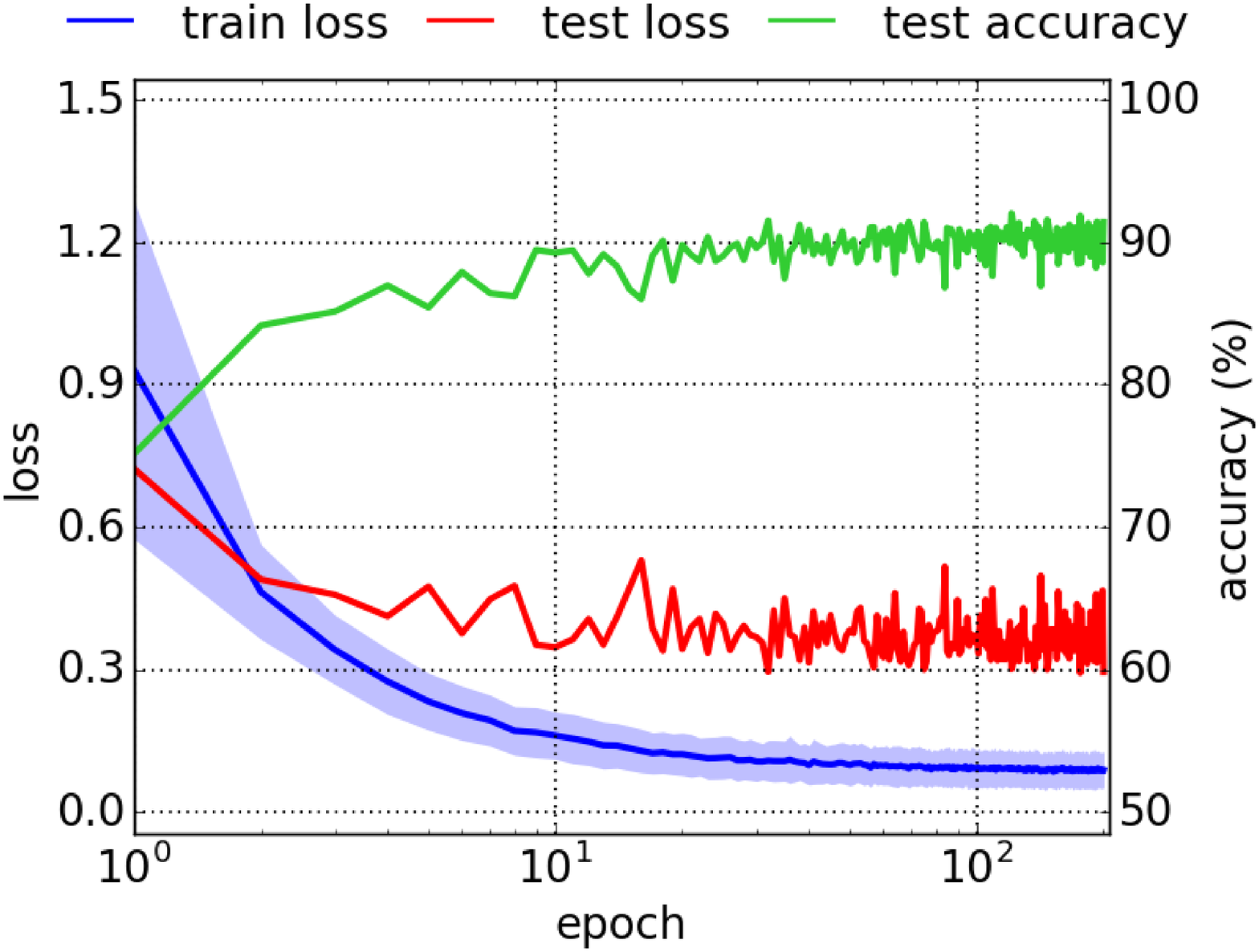} &
\includegraphics[height=\fw]{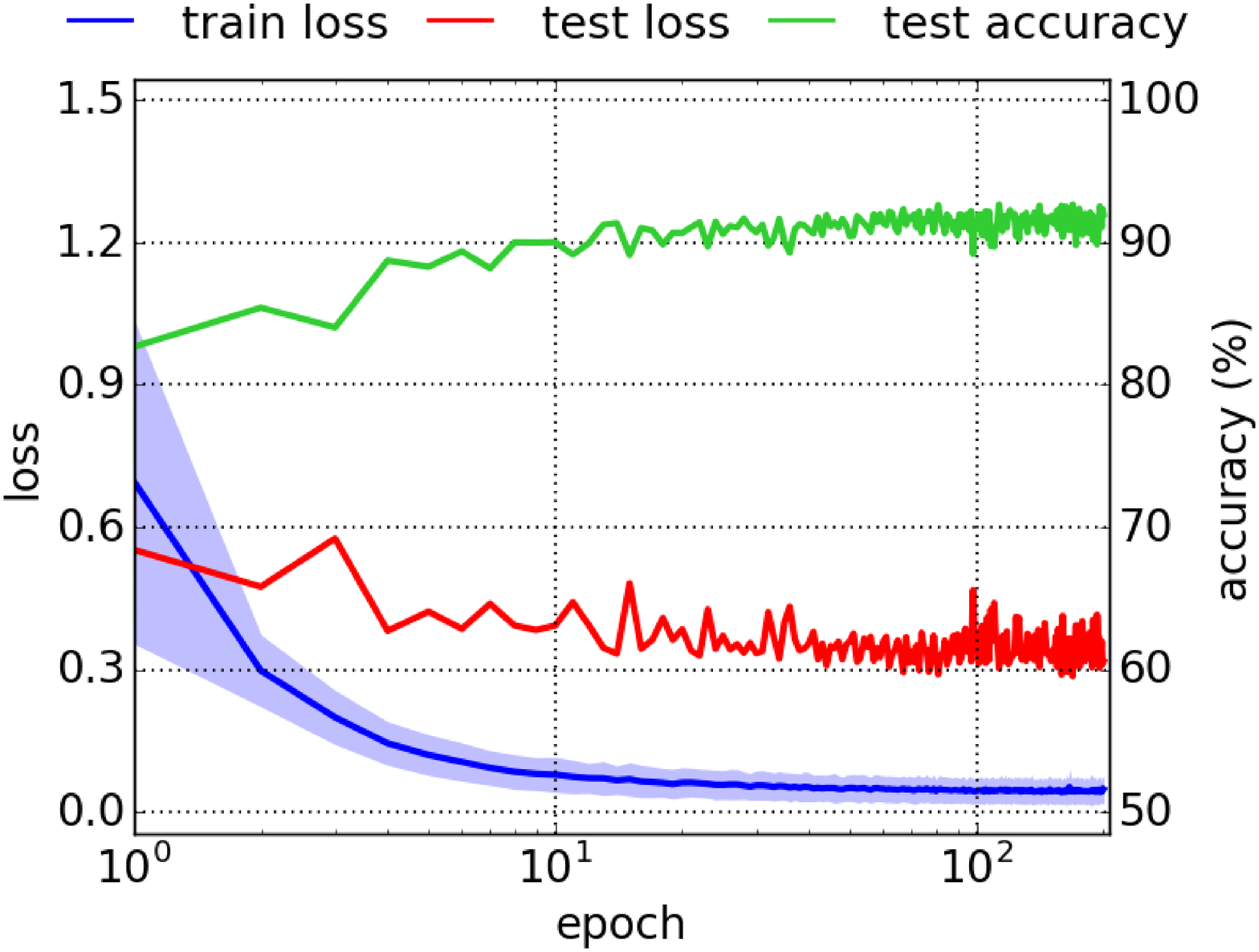} &
\includegraphics[height=\fw]{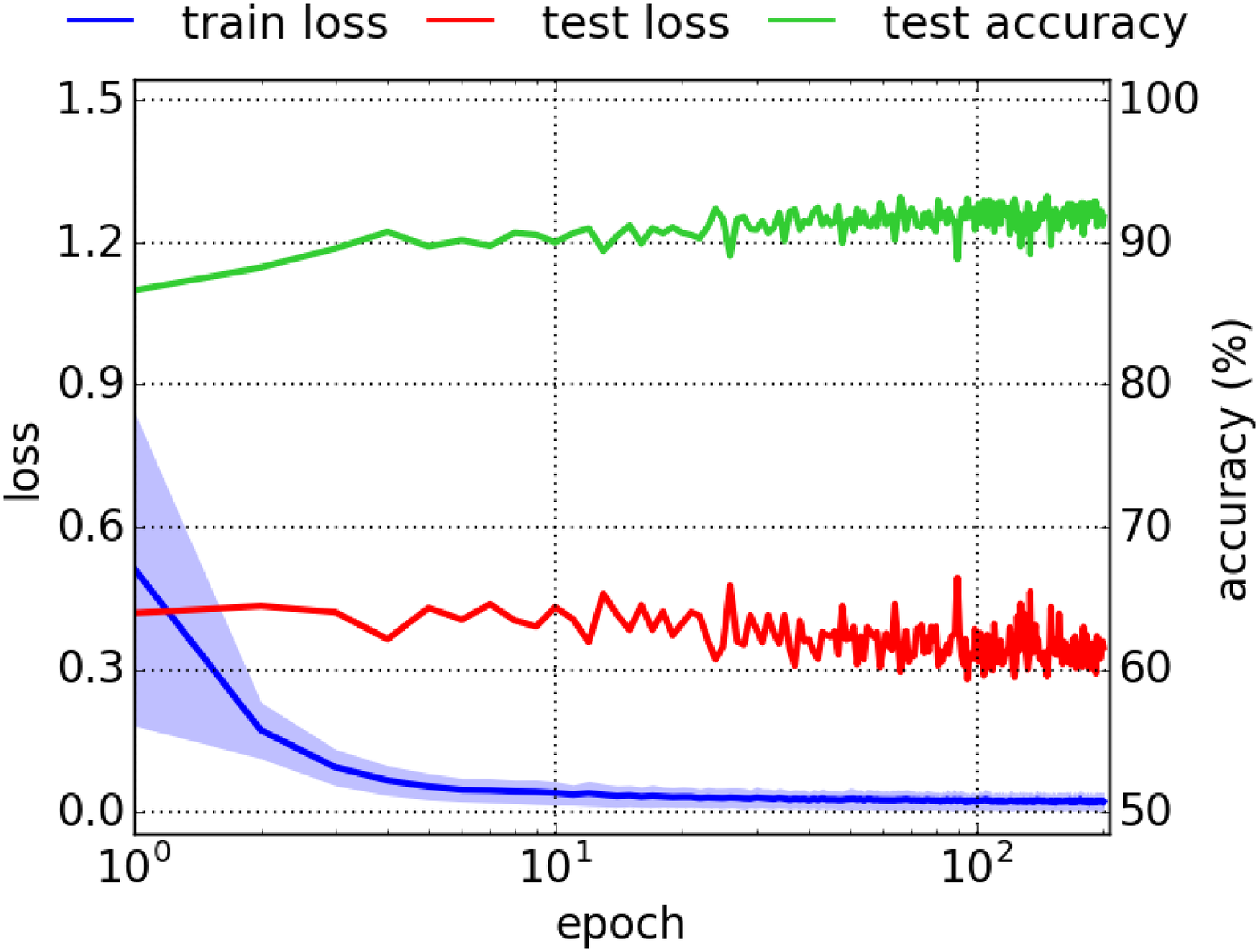} \\
$M=1$ (SGD) & $M=2$ & $M=4$ & $M=8$ & $M=16$ \\
\multicolumn{5}{c}{(c) ResNet18~\cite{he2016deep, he2016identity}} 
\end{tabular}%
\caption{{\bf Effect of the number of parameter blocks $M$} Learning curves optimized by our BCSC with varying number of parameter blocks $M$ on Cifar10. BCSC with $M = 1$ is equivalent to SGD.
}
\label{fig:result_on_M}
\end{figure*}
%
%

{\bf Effectiveness of the number of parameter blocks}
\hspace{5pt}
We initially design the experiment to validate the behavior of our algorithm as a function of the number of parameter blocks $M$. 
Thus, we compare our BCSC with varying $M = 2, 4, 8, 16$ against SGD ($M = 1$) using the models LeNet4, VGG19, and ResNet18, on the Cifar10 dataset.
In this experiment, we use a fixed learning rate of $\eta = 0.1$ across all the epochs to better understand the behavior of BCSC in comparison to SGD. 
The learning curves obtained from different network models are presented with varying number of parameter blocks $M$ in Fig.~\ref{fig:result_on_M} where it is clearly observed that both training and testing losses (red and blue lines) are significantly improved with increasing $M$ in particular with deeper network models where the number of parameters is large, resulting in a notable improvement of the test accuracy (green line).
It is also observed that the convergence speed becomes faster and the variation of the training loss decreases earlier with increasing $M$.
%
%
%
%
\begin{table}[hbt] 
	\centering
	\caption{Test accuracy with varying degree of outliers (\%)}
	{\small
	\setlength\tabcolsep{3pt} 
	\begin{tabular}{ l | c c c c }
	\hline
	Training Outlier (\%) 	& 0 	& 5 	& 10 	& 15\\
	\hline
	SGD	 ($M = 1$)	& 57.91 & 53.43 & 52.80 & 51.79\\ 
	BCSC ($M = 2$) 	& 67.80 & 66.31 & 64.45 & 64.98\\
	BCSC ($M = 4$) 	& 71.72 & 71.32 & 70.73 & 70.12\\
	BCSC ($M = 8$) 	& {\bf 73.88} & {\bf 73.64} & {\bf 73.56} & {\bf 73.21}\\
	\hline
	\end{tabular}
	} 
	\label{tab:accuracy_cifar10_outlier}
\end{table}
%
%

{\bf Robustness to training outliers}
\hspace{5pt}
To demonstrate the robustness of our BCSC to training outliers in comparison to SGD, we compute test accuracy with parameter blocks $M$ = $2, 4, 8$ in the presence of arbitrarily corrupted training data with varying rate of outliers from $0$\% (original), $5$\%, $10$\%, $15$\% based on the model LeNet4 with the Cifar10 dataset. 
Table~\ref{tab:accuracy_cifar10_outlier} presents the average testing accuracy over epoch and clearly demonstrates the relative benefit of using our BCSC with increasing $M$ against the training outliers. 
SGD is shown to be more sensitive to outliers, whereas BCSC is essentially unaffected up to the percentage of outliers tested.
%
%
%
%
\def\fw{70pt}
\def\sp{5pt}
\begin{figure*}[htb]
\centering
\begin{tabular}{c@{}c@{}c@{}c}
\includegraphics[height=\fw]{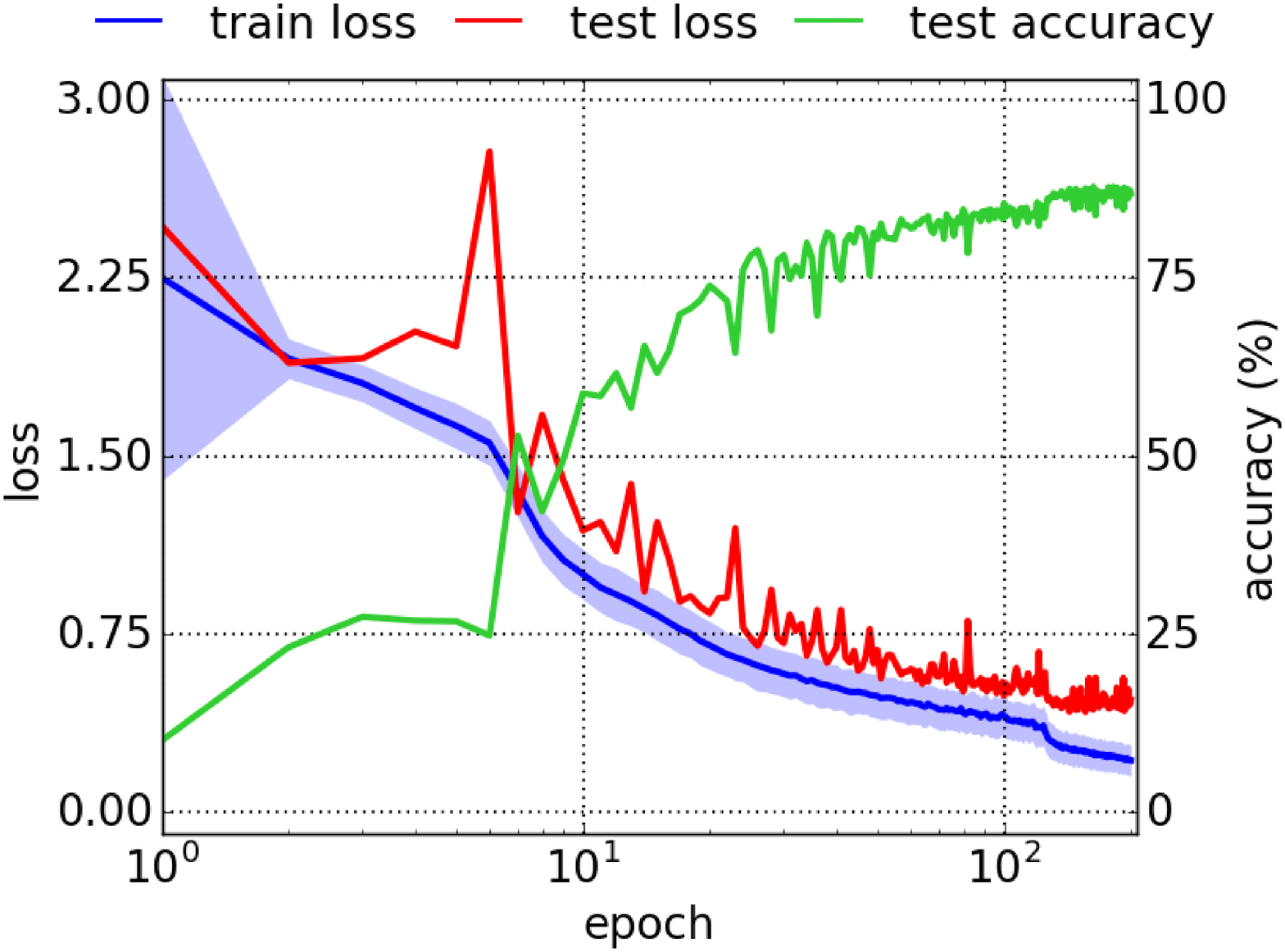} & 
\includegraphics[height=\fw]{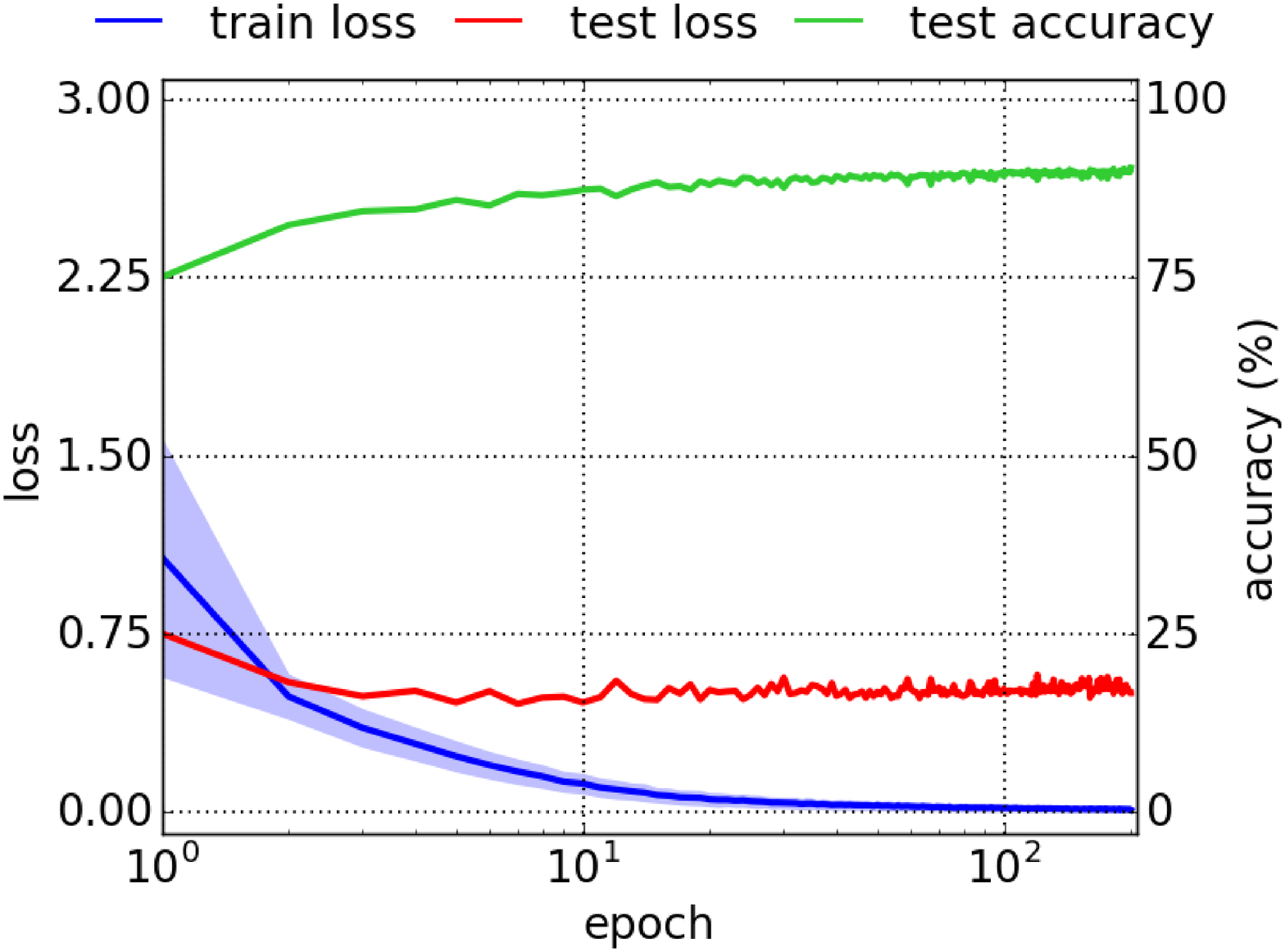} & 
\includegraphics[height=\fw]{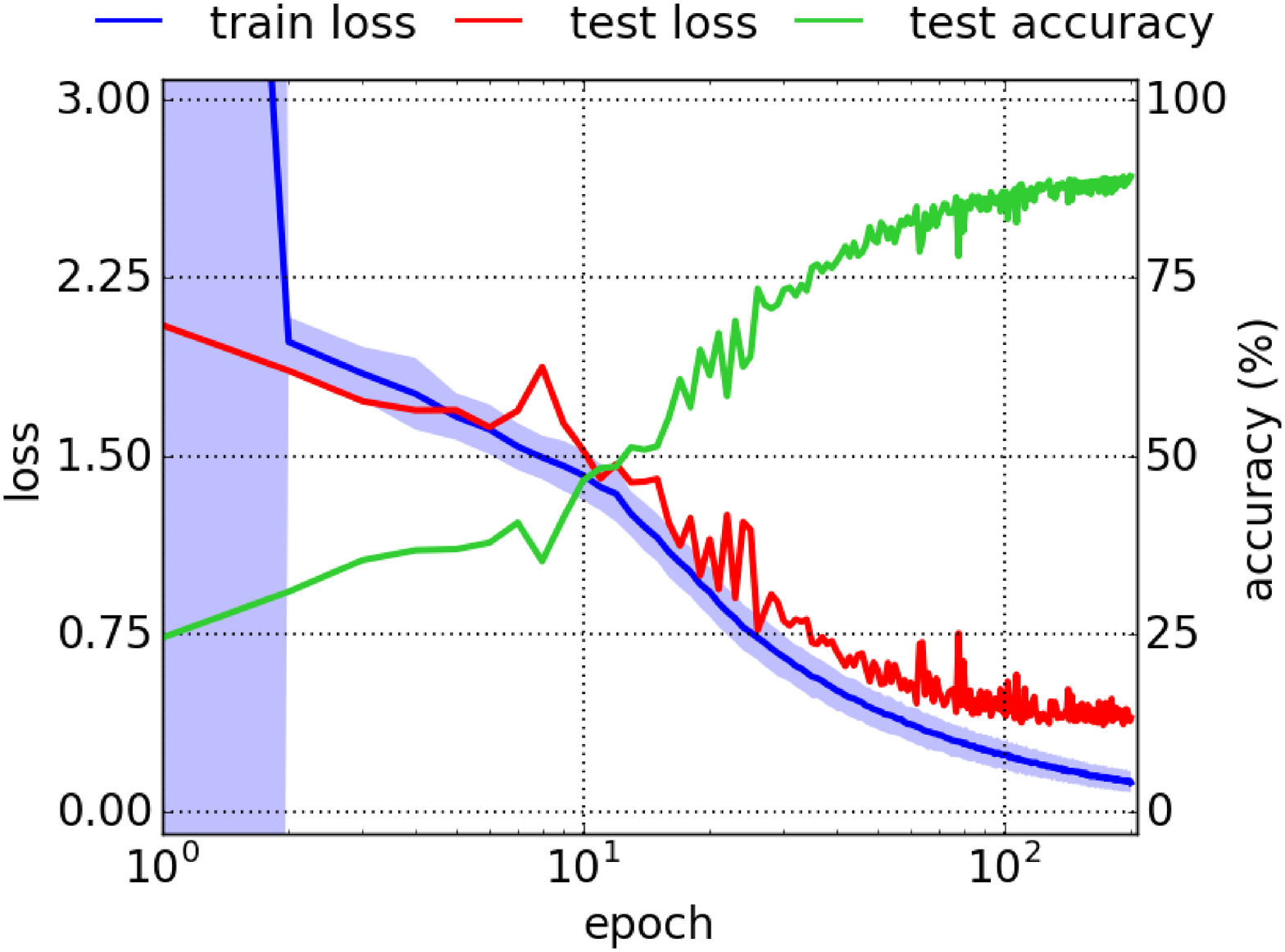} & 
\includegraphics[height=\fw]{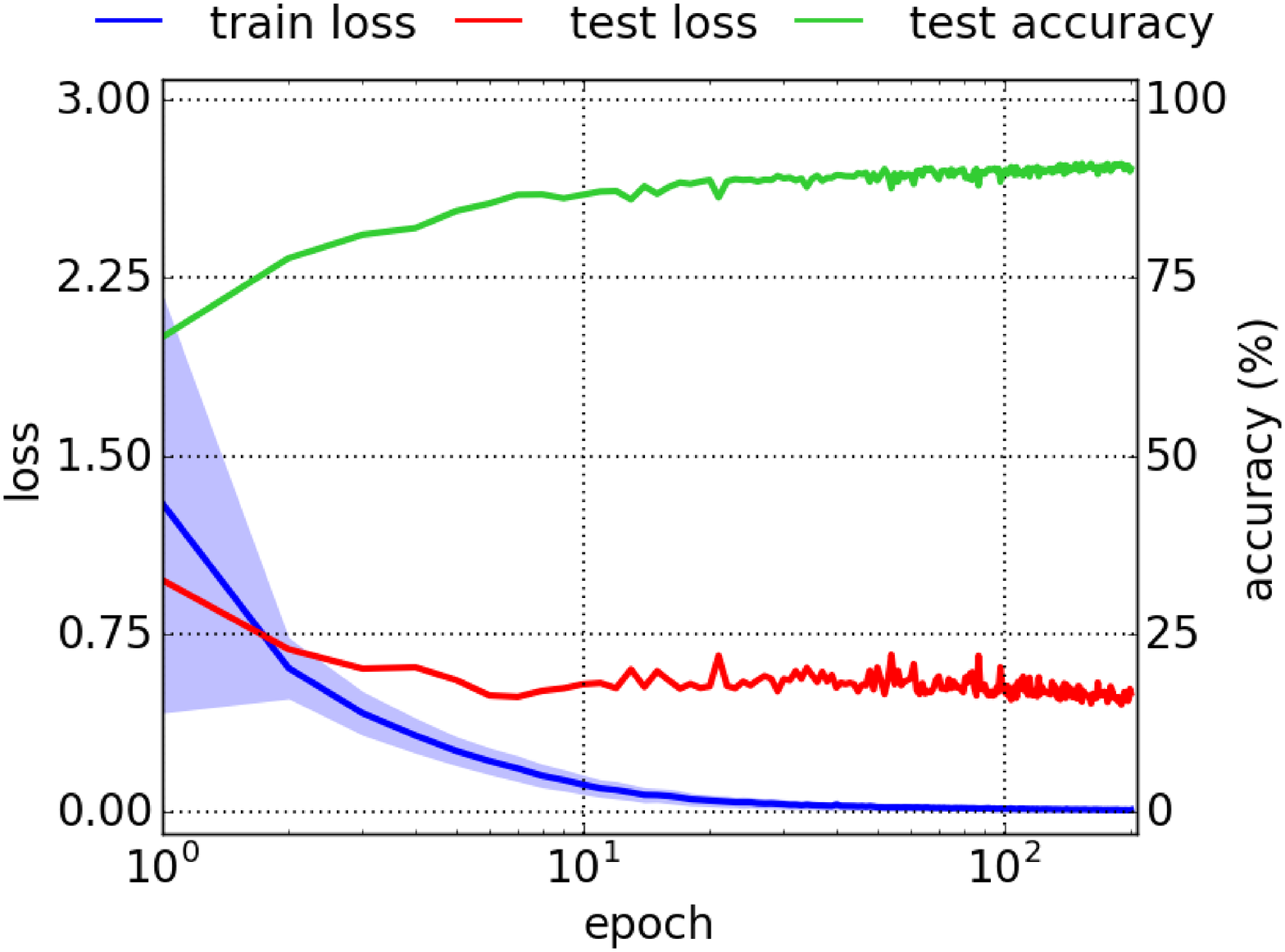} \\
AdaGrad + SGD & AdaGrad + BCSC & AdaGrad + SGD & AdaGrad + BCSC\\
\multicolumn{2}{c}{(a) VGG19~\cite{simonyan2014very}} &  \multicolumn{2}{c}{(b) ResNet18~\cite{he2016deep, he2016identity}}  \\
\end{tabular}
\caption{{\bf Comparison of BCSC with SGD when using with adaptive learning rate (AdaGrad)} Learning curves obtained by SGD with AdaGrad and BCSC with AdaGrad using Cifar10. $M$ = 8 is used for BCSC.}
\label{fig:result_adaptive_learningrate}
\end{figure*}
%
%

{\bf Results with adaptive learning rate}
\hspace{5pt}
In order to demonstrate that the benefits of BCSC are not diminished when using an adaptive learning rate, we compare BCSC with $M$ = 8 and SGD when integrated with the learning rate given by AdaGrad~\cite{duchi2011adaptive} based on the basic models: VGG19 and ResNet18, using the Cifar10 dataset. 
The learning curves are presented in Fig.~\ref{fig:result_adaptive_learningrate} where the training loss and the test loss are noticeably improved with BCSC in comparison to SGD.
The results indicate that BCSC outperforms SGD consistently, regardless of whether an adaptive learning rate scheme by AdaGrad is applied to the algorithm.
%
%
%
%
\begin{table}[htb] 
	\centering
	\caption{Test accuracy with varying degree of dropout (\%)}
	{\small
	\setlength\tabcolsep{3pt} 
	\begin{tabular}{ l | c c c c }
	\hline
	Dropout rate (\%) 	& 0 	& 5 	& 10 	& 15\\
	\hline
	SGD ($M=1$)		& 98.98			& 99.04			& 99.09			& 99.04\\
	BCSC ($M=2$)	& 99.00			& 99.10			& 99.14			& 99.13\\
	BCSC ($M=4$)	& {\bf 99.04}	& 99.10			& 99.17			& 99.16\\
	BCSC ($M=8$)	& 99.02			& {\bf 99.13}	& {\bf 99.17}	& {\bf 99.19}\\
	\hline
	\end{tabular}
	} 
	\label{tab:accuracy_mnist_dropout}
\end{table}
%
%

{\bf Results with dropout}
\hspace{5pt}
In this experiment, we demonstrate that the regularization effects of BCSC persist if additional regularization is employed, for instance using Dropout.
We employ a simple network model, LeNet4, in which we can easily observe the effect of dropout using the MNIST dataset.
Table~\ref{tab:accuracy_mnist_dropout} summarizes the average test accuracy over epoch of BCSC with parameter blocks $M$ = $2, 4, 8$ in comparison to SGD ($M = 1$) at different rates of dropout ($0\%, 5\%, 10\%, 15\%$). 
It is shown that BCSC outperforms SGD regardless of Dropout even though the effectiveness of a larger number of parameter blocks $M$ is shown to be weaker, which is due to the relatively small number of parameters in the network model.
%
%
%
%
\def\fw{73pt}
\def\sp{5pt}
\begin{figure*}[htb]
\centering
\begin{tabular}{c@{}c@{}c@{}c}
\includegraphics[height=\fw]{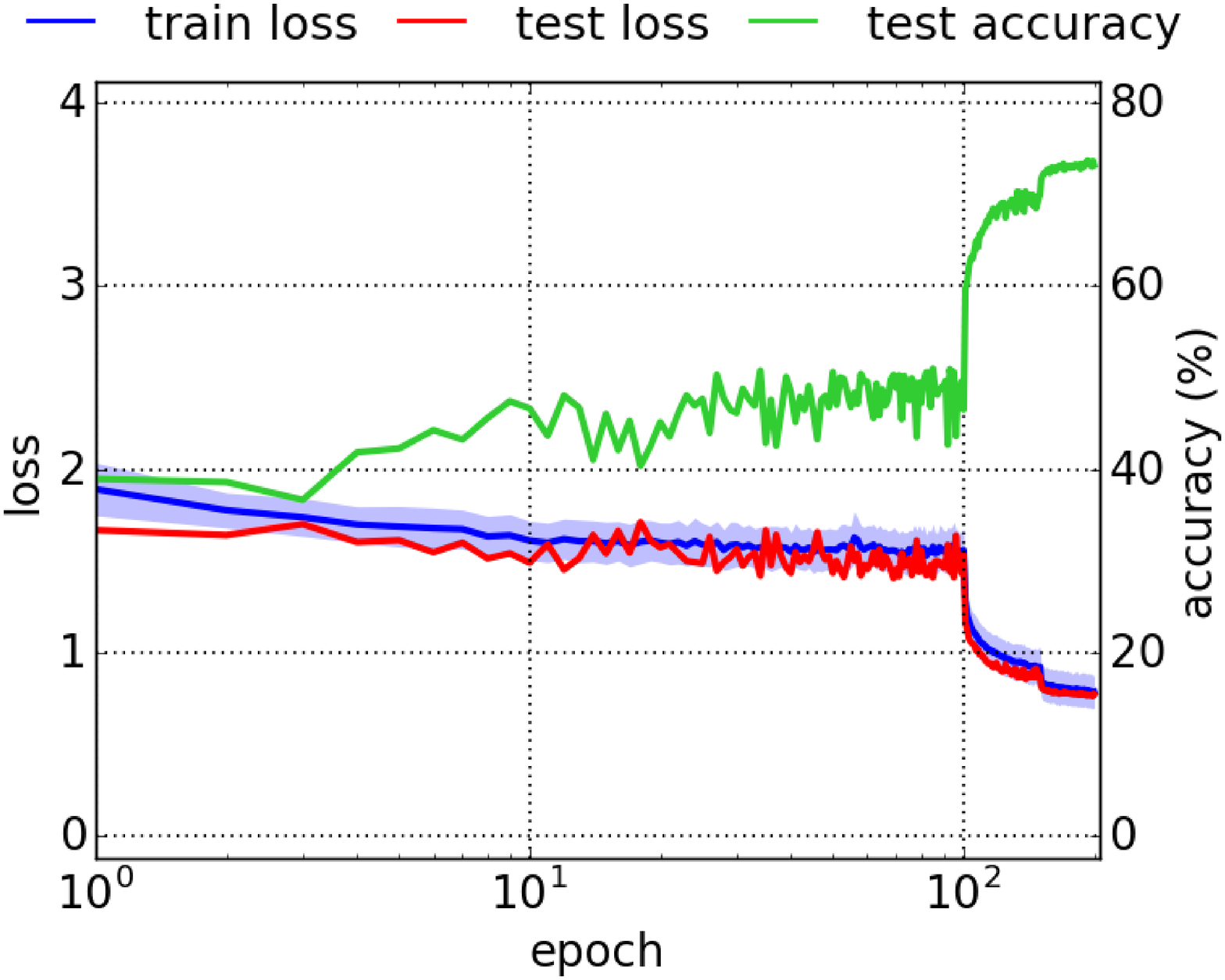} & 
\includegraphics[height=\fw]{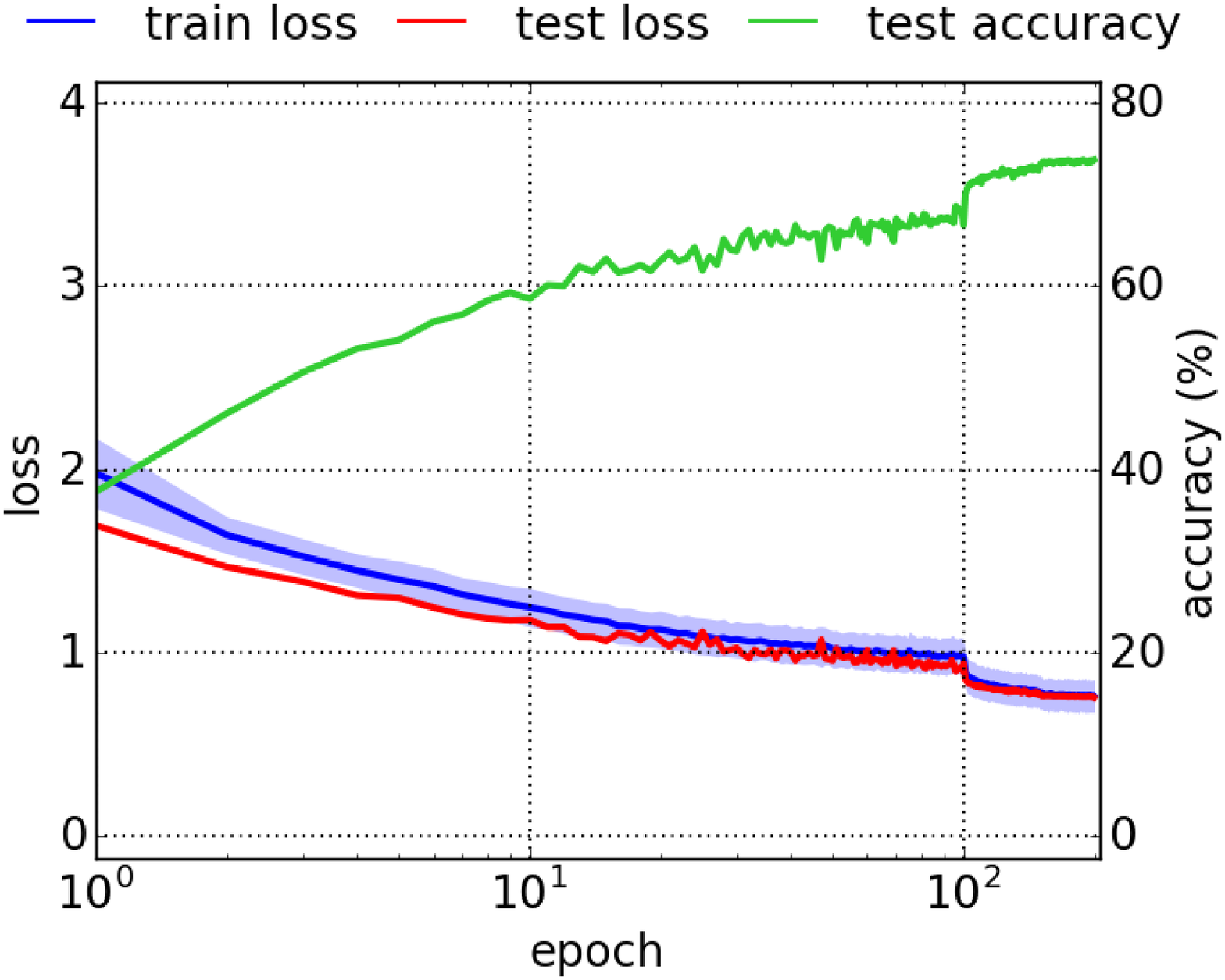} & 
\includegraphics[height=\fw]{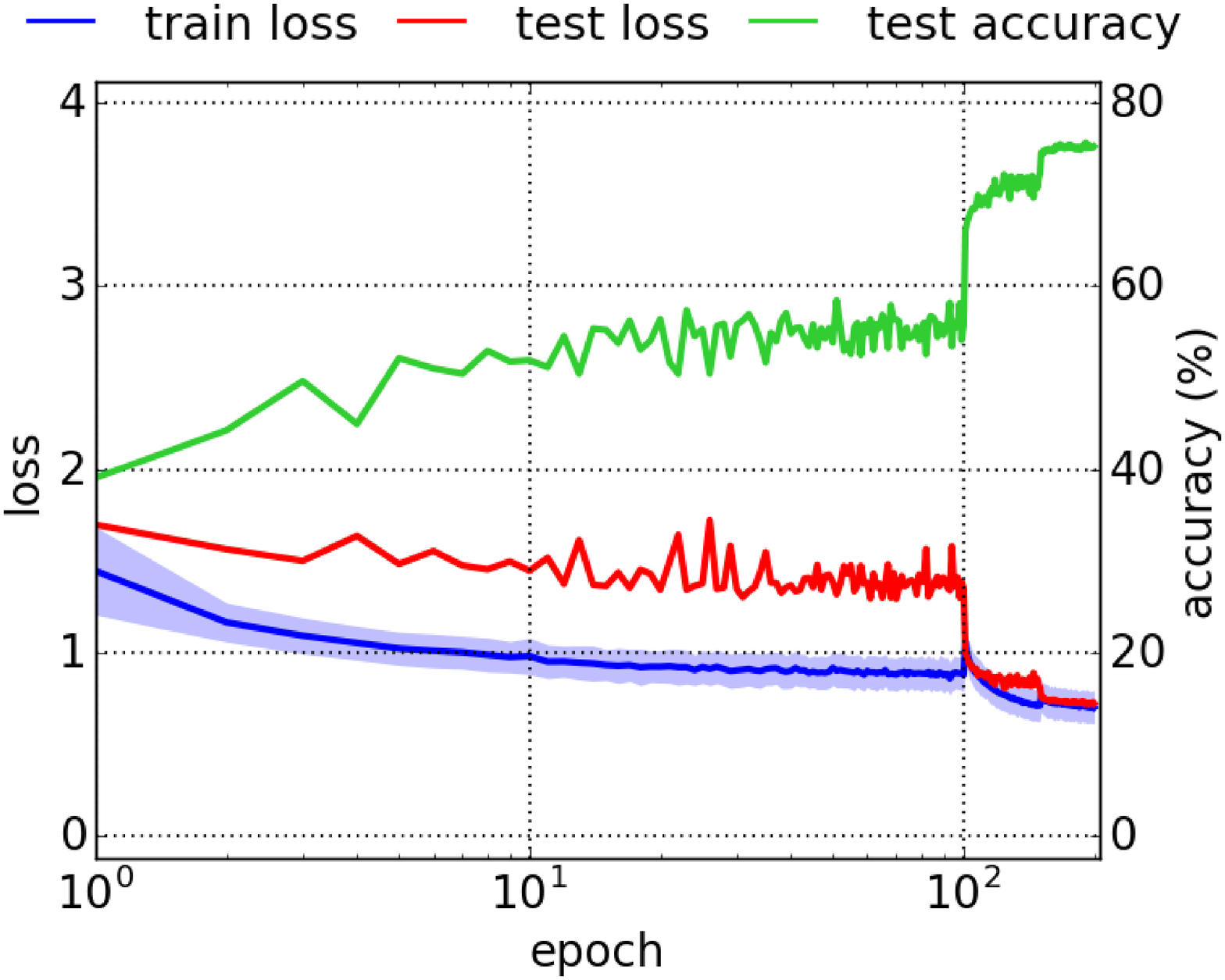} & 
\includegraphics[height=\fw]{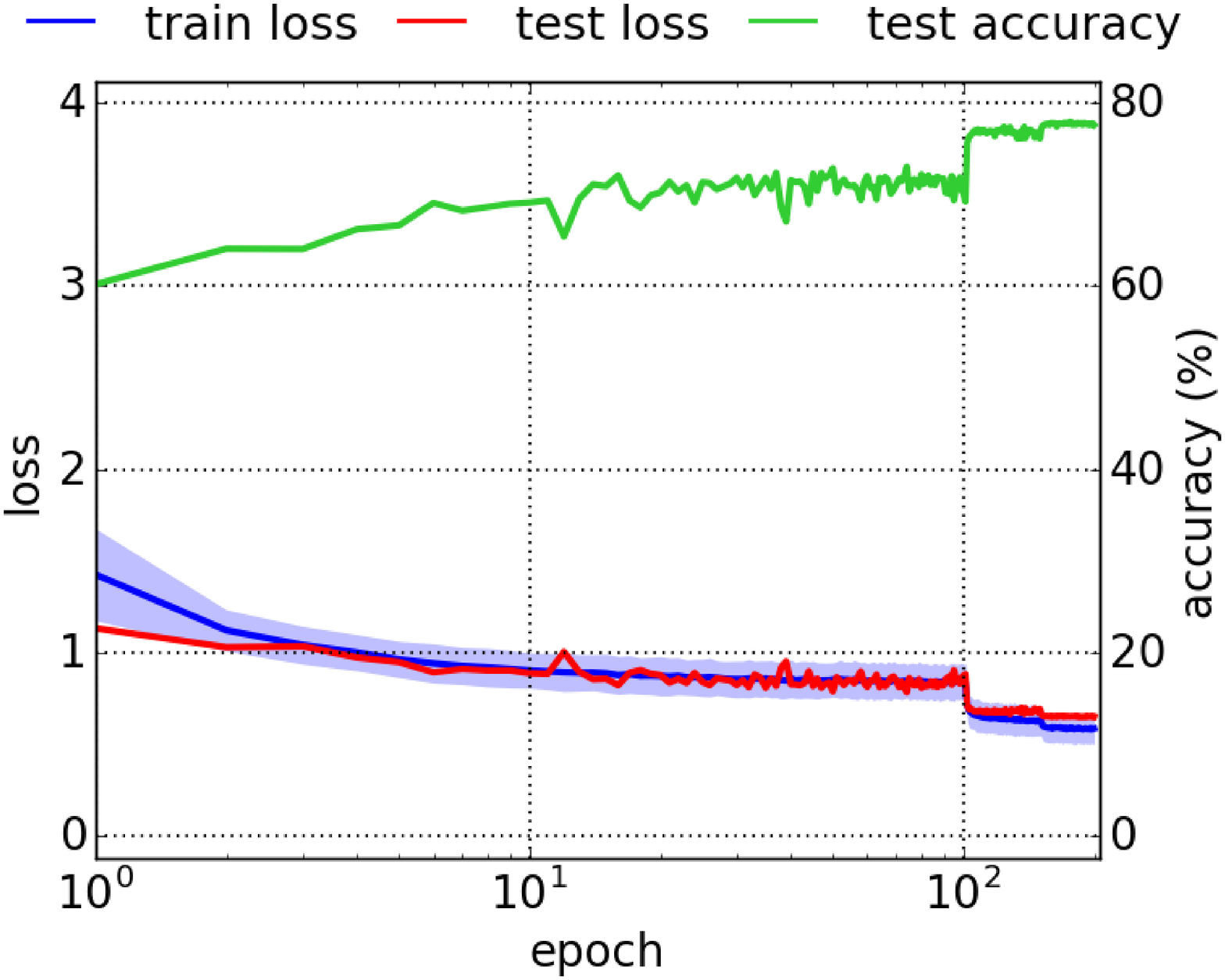} \\
SGD & SBC & RBC & BCSC\\
\multicolumn{4}{c}{(a) LeNet4~\cite{lecun1998gradient}} \\
\\
\includegraphics[height=\fw]{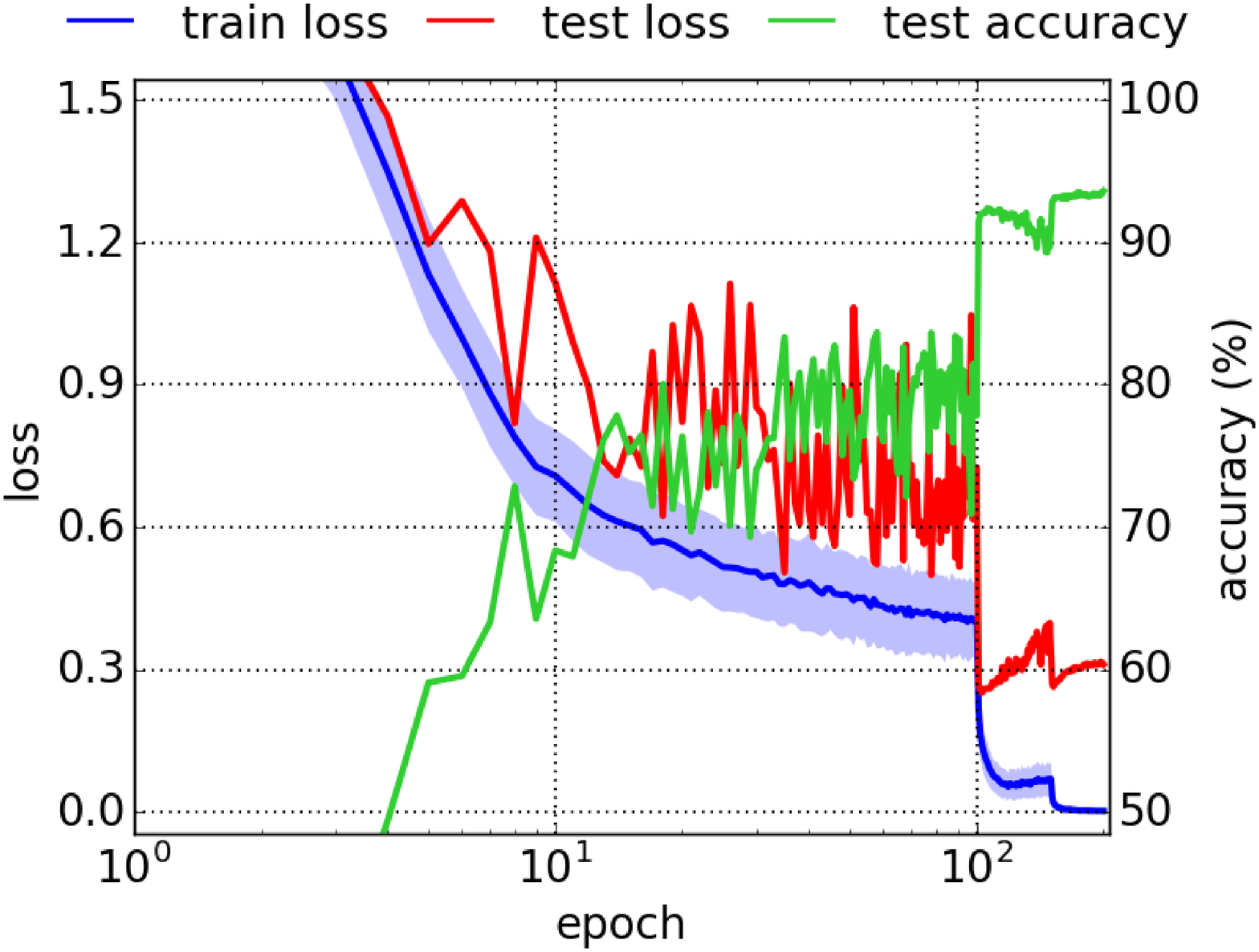} & 
\includegraphics[height=\fw]{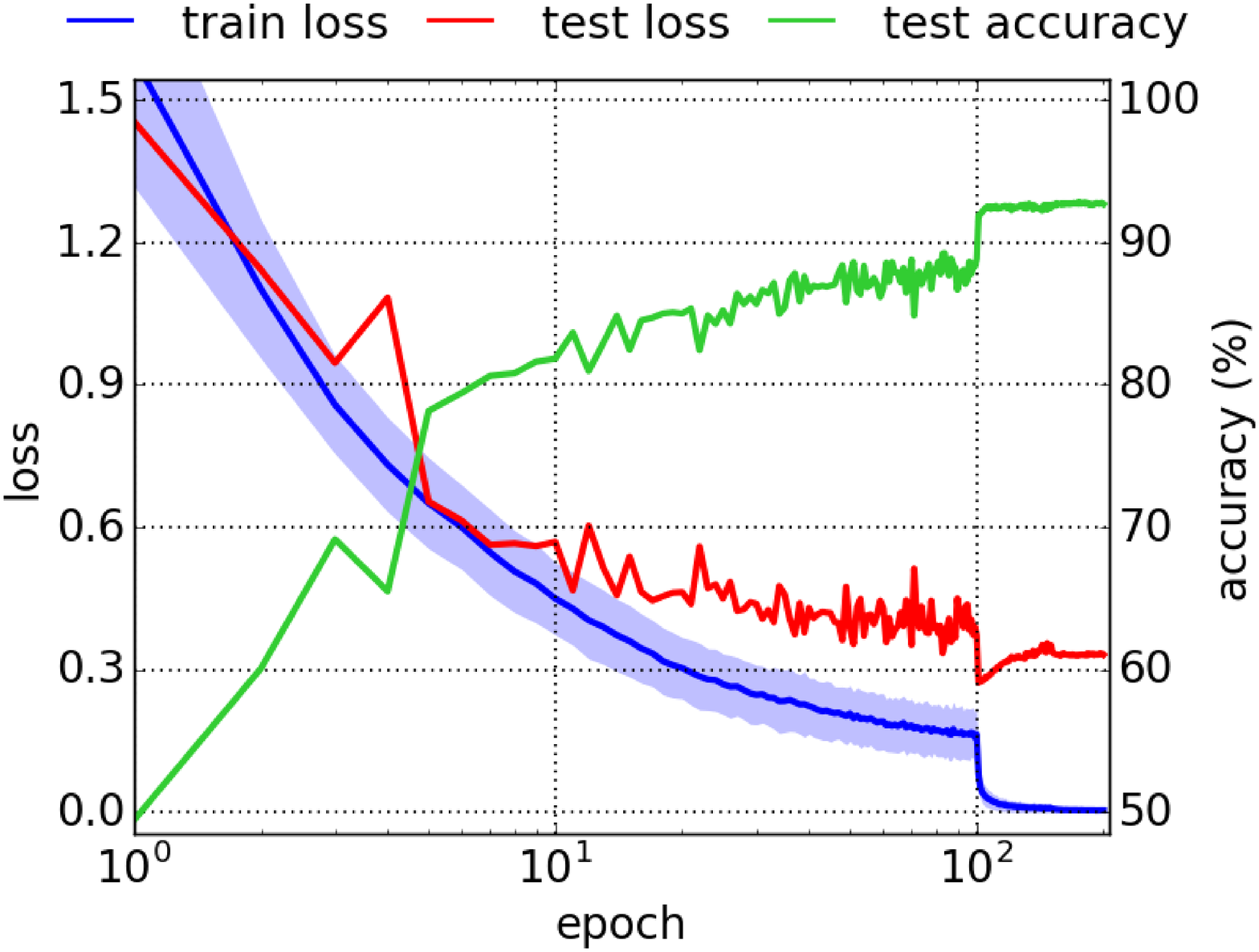} & 
\includegraphics[height=\fw]{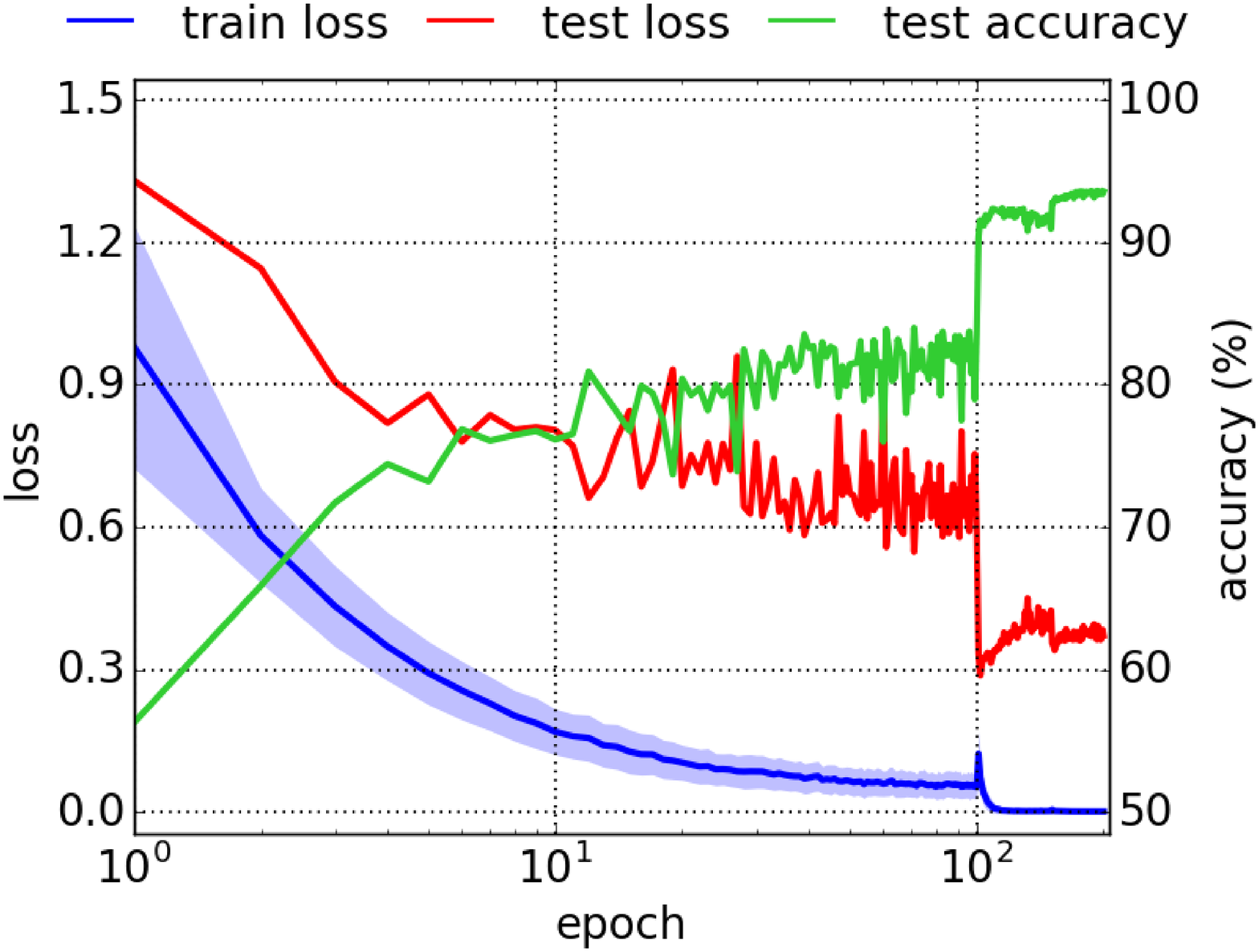} & 
\includegraphics[height=\fw]{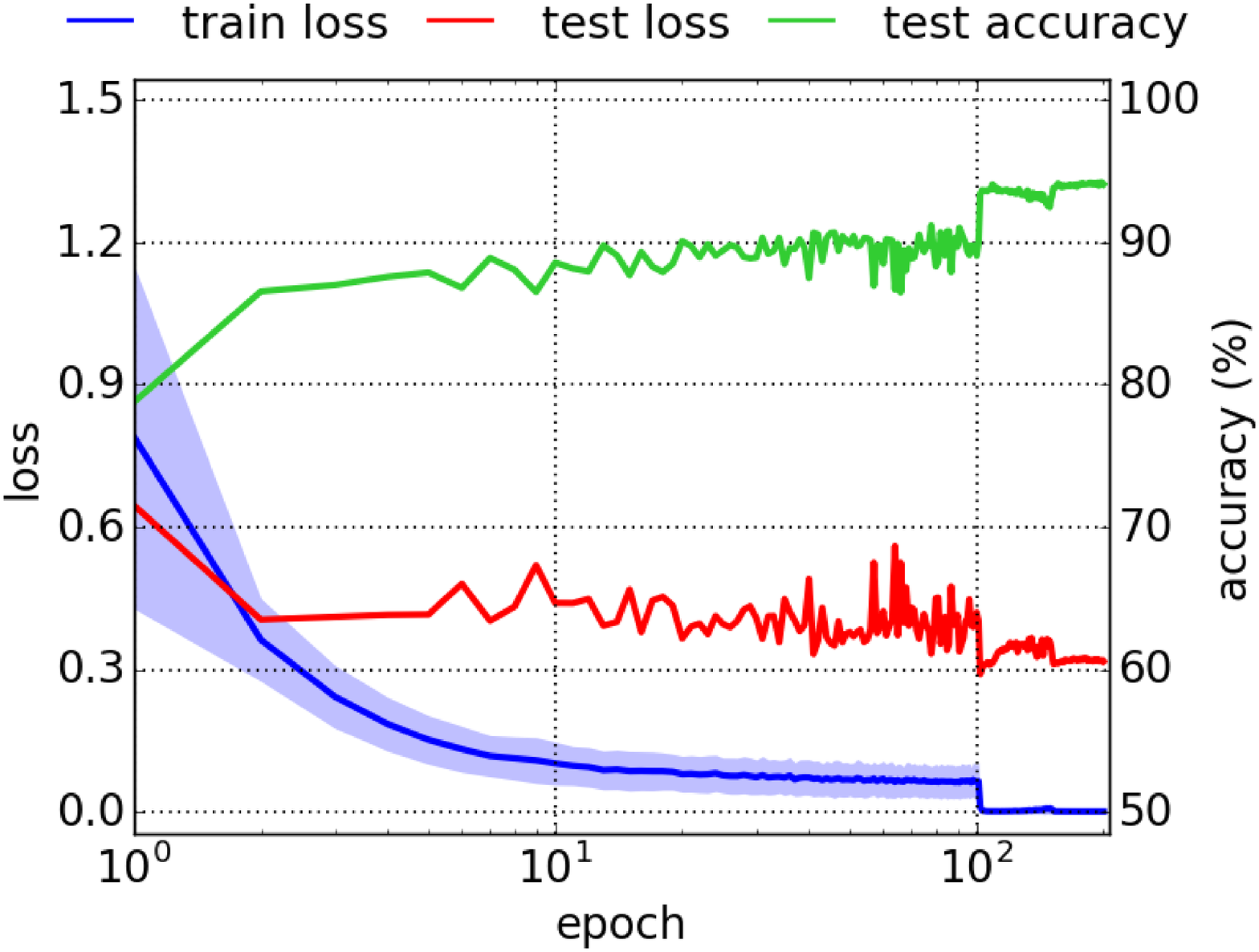} \\
SGD & SBC & RBC & BCSC\\
\multicolumn{4}{c}{(b) VGG19~\cite{simonyan2014very}} \\
\\
\includegraphics[height=\fw]{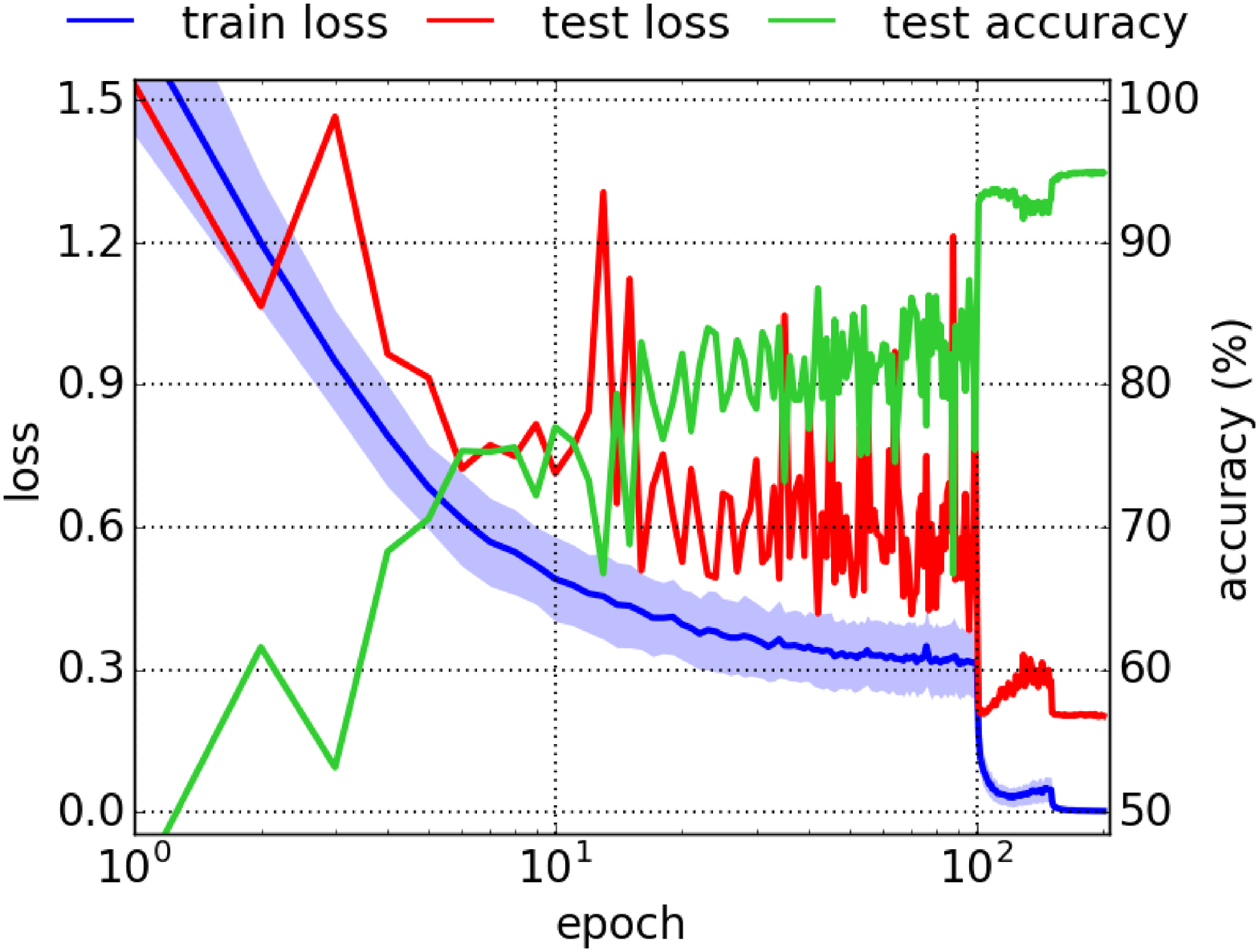} & 
\includegraphics[height=\fw]{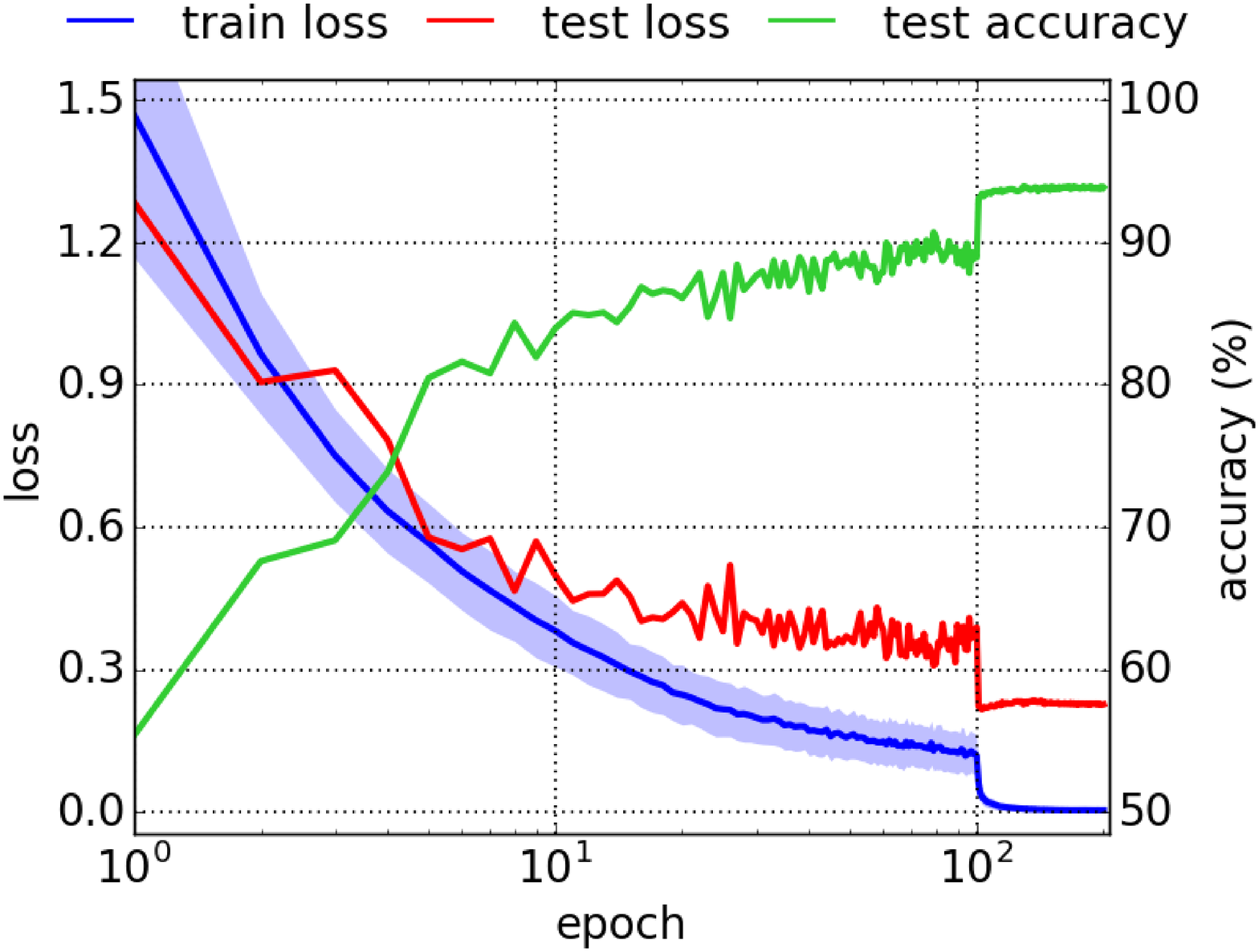} & 
\includegraphics[height=\fw]{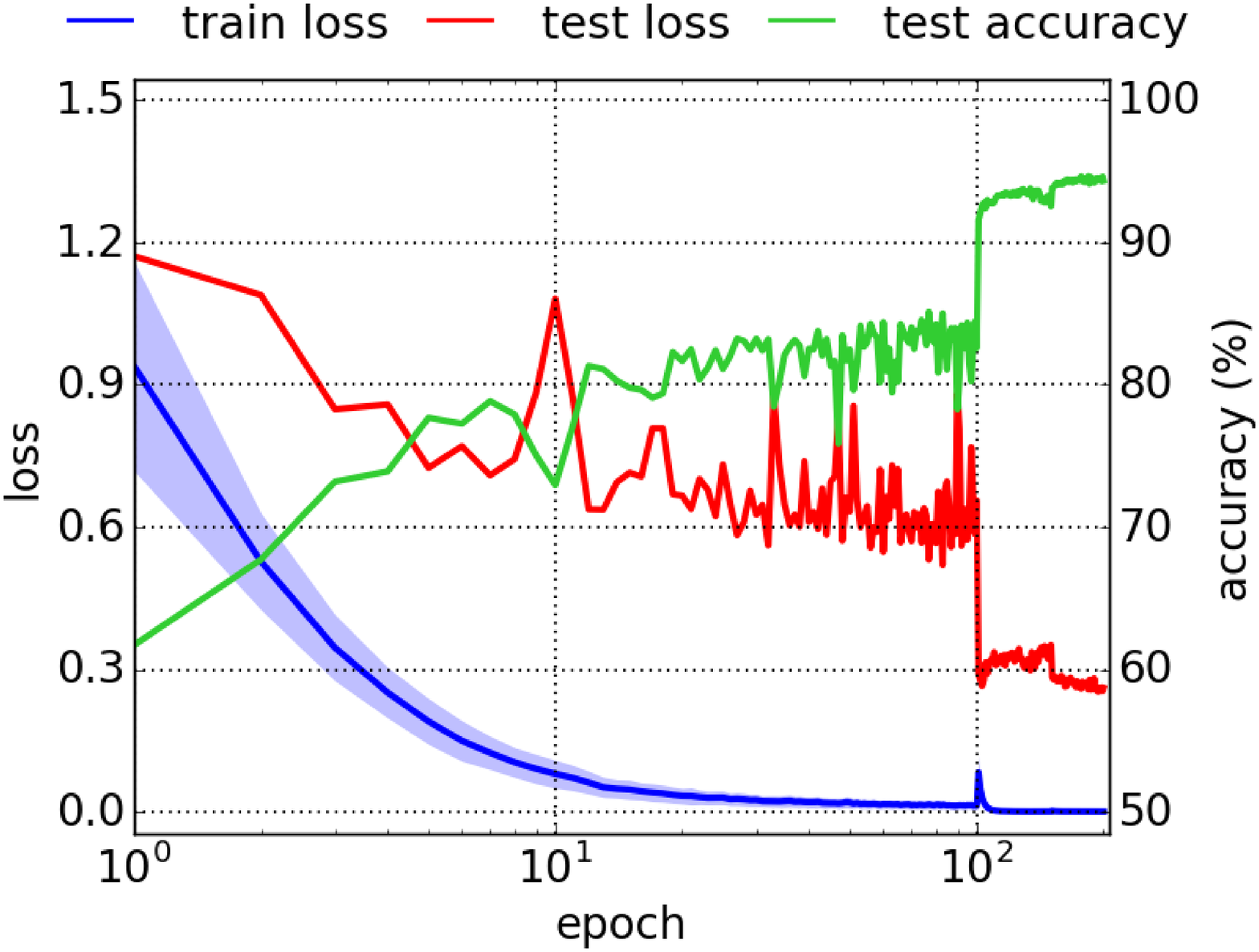} & 
\includegraphics[height=\fw]{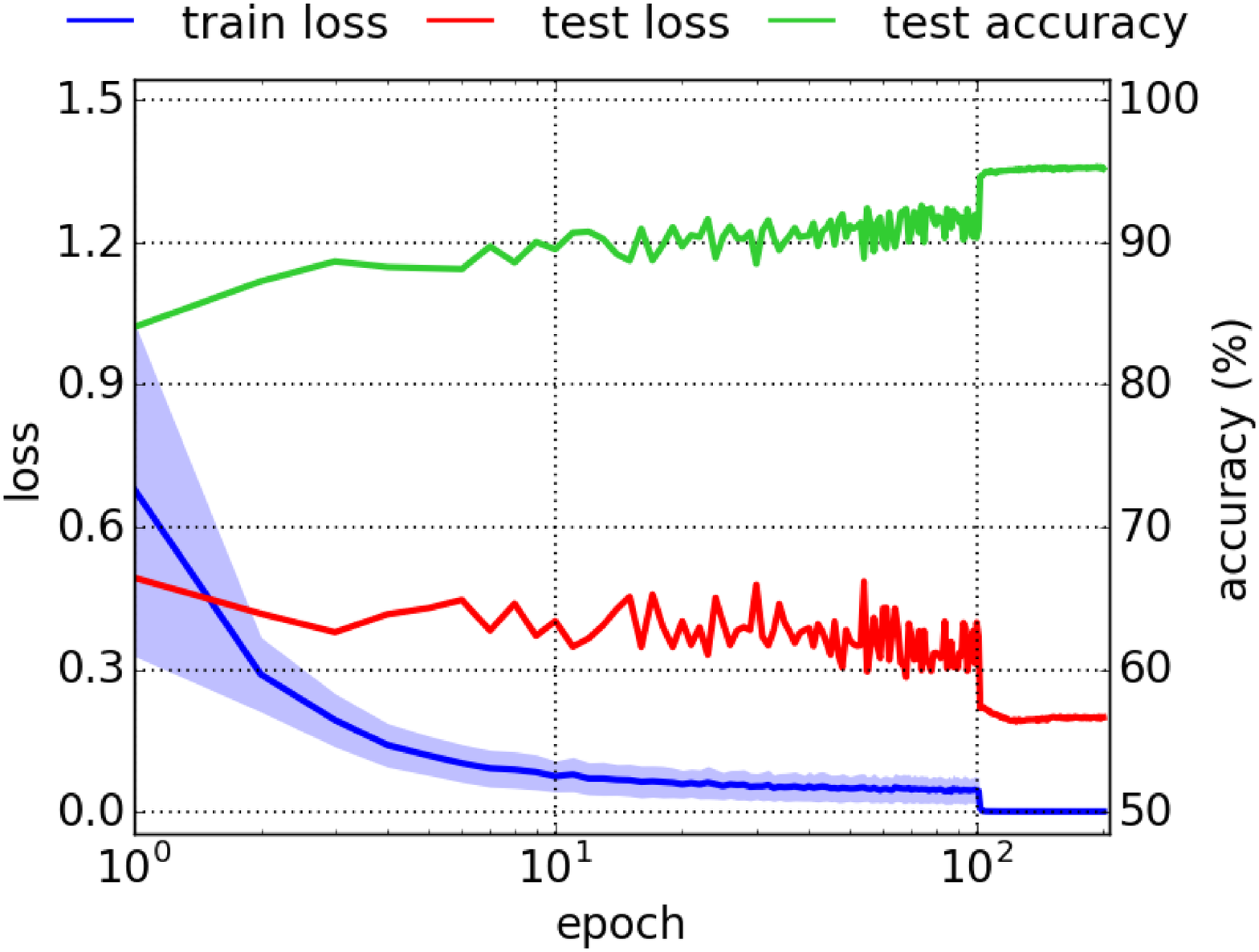} \\
SGD & SBC & RBC & BCSC\\
\multicolumn{4}{c}{(c) ResNet18~\cite{he2016deep, he2016identity}} 
\end{tabular}%
\caption{{\bf Evaluation on Cifar10} Learning curves optimized by (SGD) stochastic gradient descent, (SBC) stochastic randomized block-coordinate descent, (RBC) randomized block-coordinate descent, and (BCSC) our algorithm with $M = 8$.}
\label{fig:result_cifar10_basic}
\end{figure*}
%
%
%
%
%
\begin{table*}[htb] 
	\centering
	\caption{Test accuracy for Cifar10 (\%)}
	{\scriptsize
	\setlength\tabcolsep{2pt} 
	\begin{tabular}{l | c c c c c c | c c c c c c | c c c c c c | c c c c c c }
	\hline
				& \multicolumn{6}{ c |}{(a) First half epochs}& \multicolumn{6}{ c |}{(b) Last half epochs}& \multicolumn{6}{ c |}{(c) All epochs}& \multicolumn{6}{ c }{(d) Final epoch}\\
	 			& AG & AD & SGD & SBC & RBC & BCSC 	& AG & AD & SGD & SGD & SBC & BCSC 	& AG & AD & SGD & SBC & RBC & BCSC 	& AG & AD & SGD & SBC & RBC & BCSC\\
	\hline
	LeNet4		& 52.79	& 65.89	& 46.98	& 64.32	& 54.34	& {\bf 70.49} 		& 62.15	& 69.37	& 70.33	& 72.90	& 72.74	& {\bf 77.17}		& 57.47	& 67.63	& 58.66	& 68.61	& 63.54	& {\bf 73.83}		& 62.12	& 69.64	& 73.24	& 73.80	& 75.25	& {\bf 77.61}\\
	VGG19		& 82.42	& 86.60	& 75.28	& 85.40	& 80.09	& {\bf 89.22}		& 89.07	& 91.55	& 92.33	& 92.58	& 92.57	& {\bf 93.70}		& 85.75	& 89.07	& 83.81	& 88.99	& 86.33	& {\bf 91.46}		& 89.16	& 91.86	& 93.62	& 92.69	& 93.58	& {\bf 94.09}\\
	ResNet18	& 58.43	& 87.58	& 79.43	& 87.01	& 81.34	& {\bf 90.64}		& 78.10	& 91.98	& 93.89	& 93.76	& 93.74	& {\bf 95.12}		& 68.26	& 89.78	& 86.66	& 90.38	& 87.54	& {\bf 92.88}		& 82.20	& 92.40	& 94.90	& 93.87	& 94.34	& {\bf 95.19}\\
	\hline
	\end{tabular}
	} 
	\label{tab:accuracy_cifar10}
\end{table*}
%
%

%
%
%
%
%
\def\fw{70pt}
\begin{figure*}[htb]
\centering
\begin{tabular}{c@{}c@{}c@{}c}
\includegraphics[height=\fw]{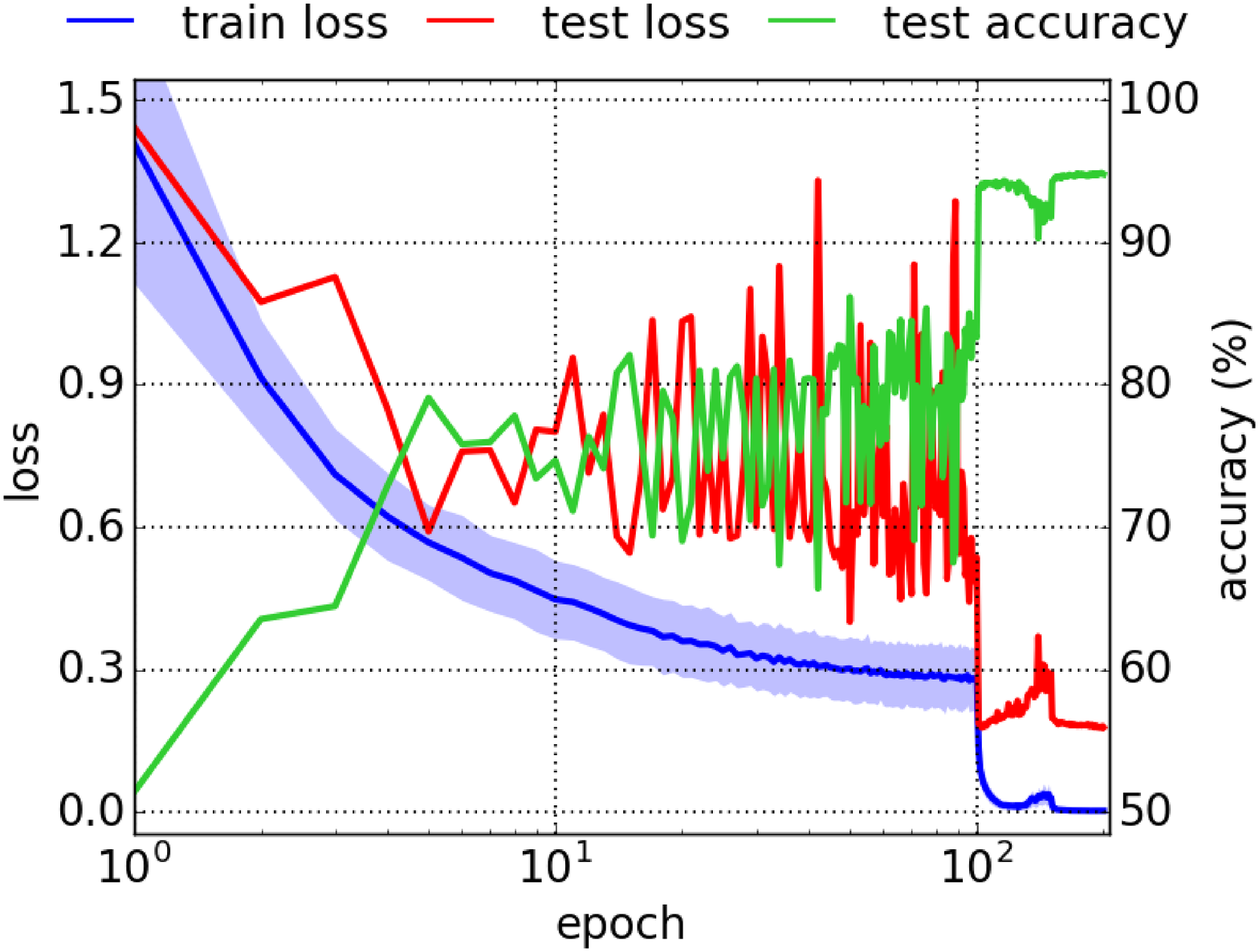} & 
\includegraphics[height=\fw]{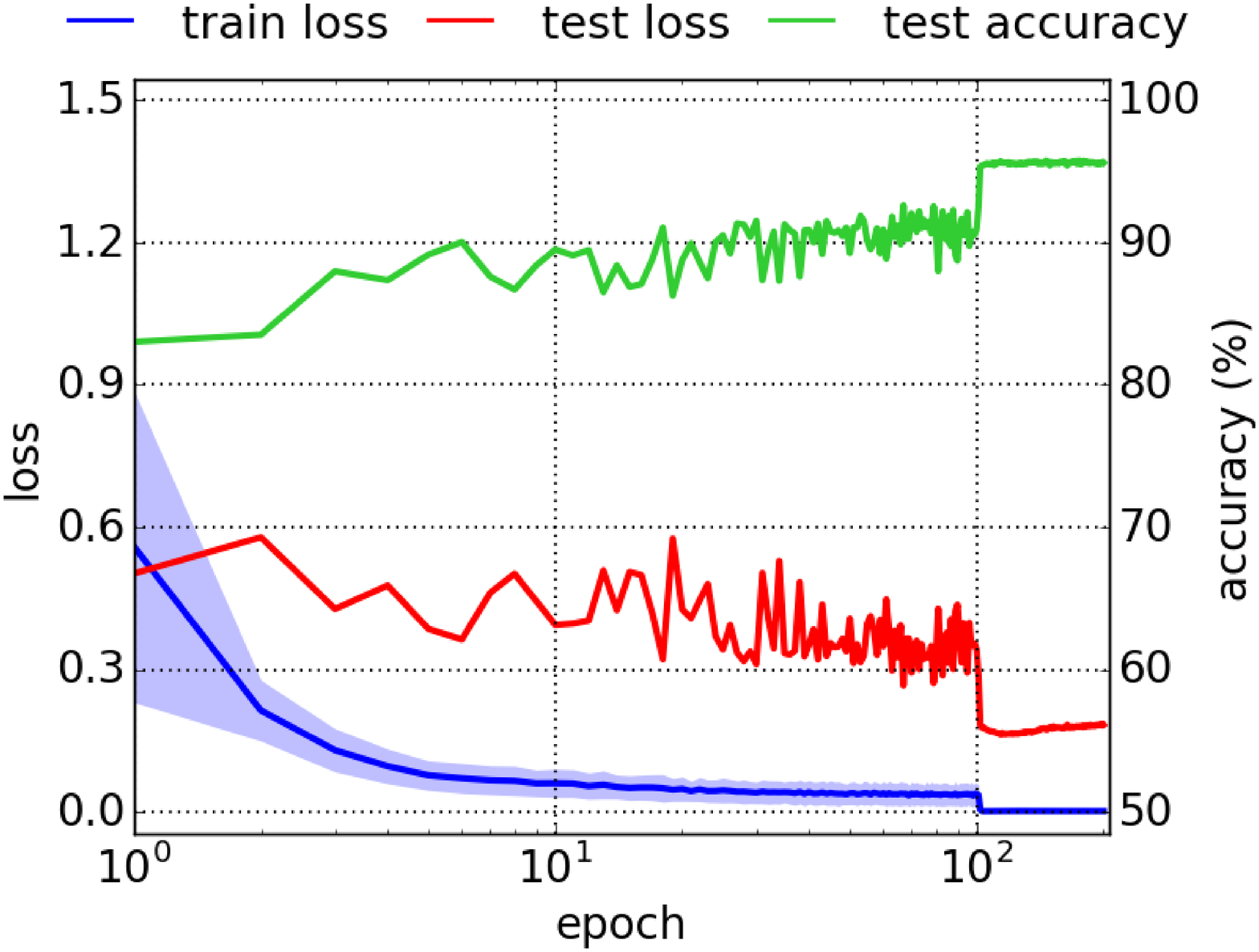} & 
\includegraphics[height=\fw]{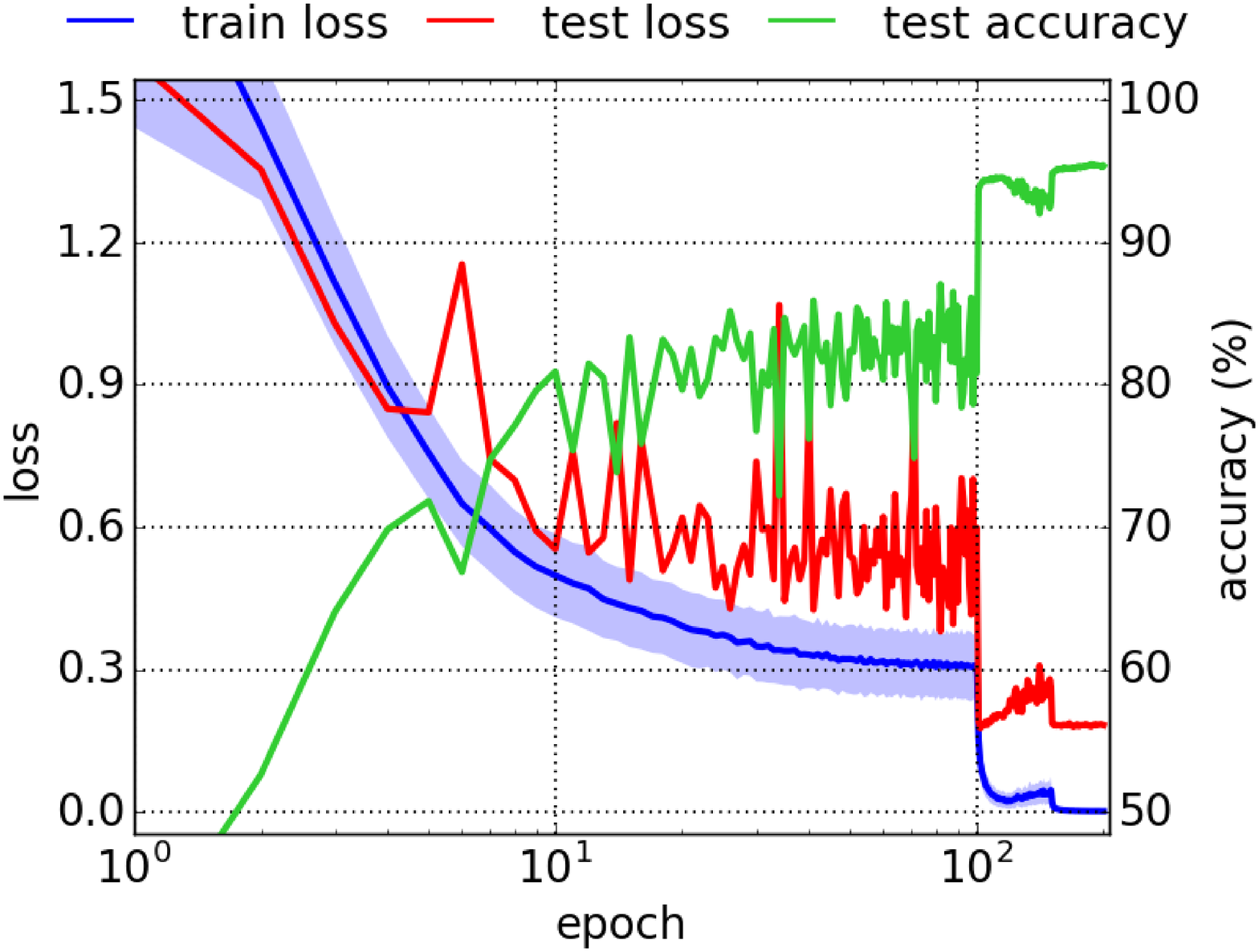} & 
\includegraphics[height=\fw]{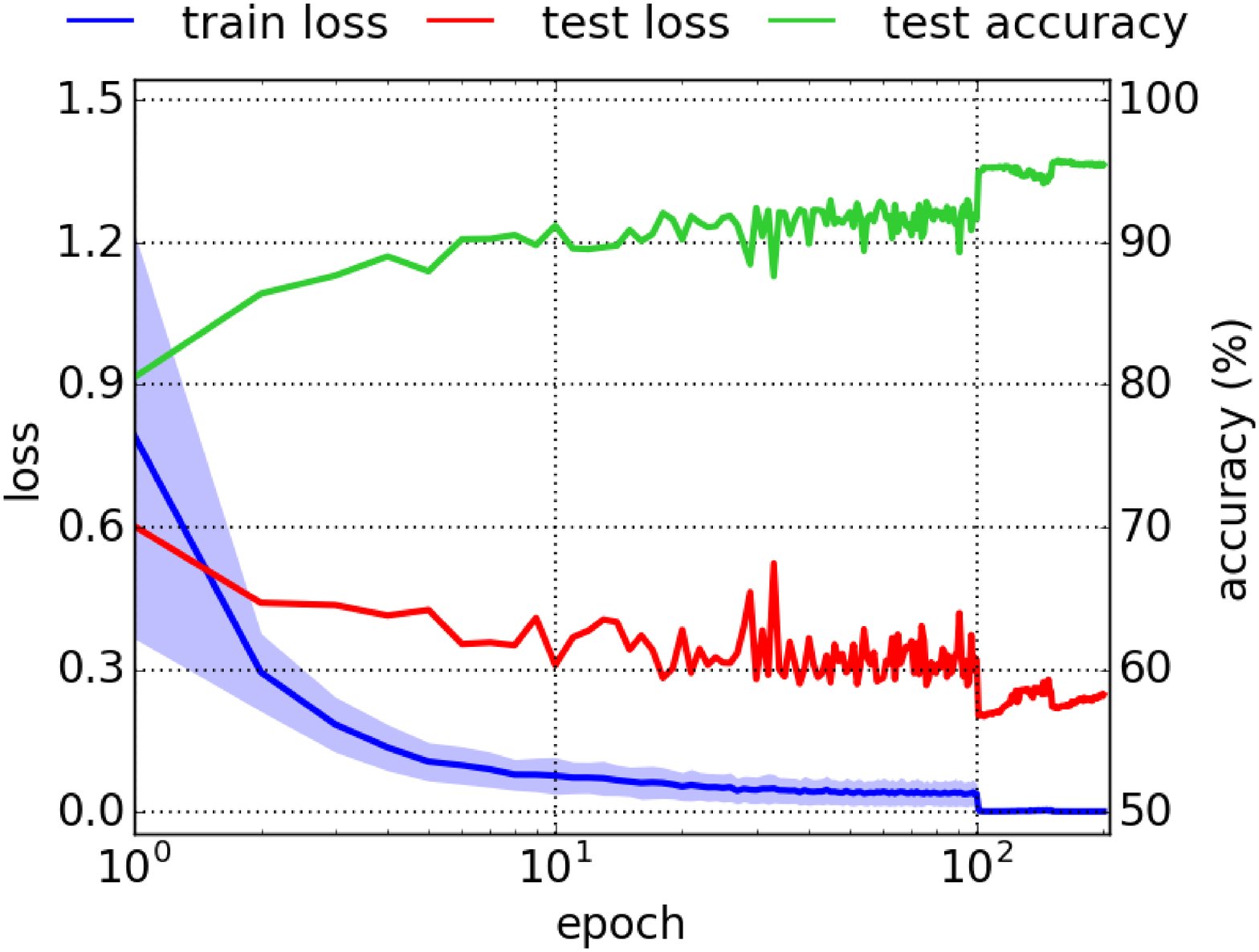} \\
SGD & BCSC & SGD & BCSC\\
\multicolumn{2}{c}{(a) GoogLeNet~\cite{szegedy2015going}} &  \multicolumn{2}{c}{(b) DPN92~\cite{chen2017dual}}  \\
\end{tabular}%
\caption{{\bf Deep models on Cifar10} Learning curves optimized by SGD and BCSC with $M$ = $8$.}
\label{fig:result_cifar10_deep}
\end{figure*}
%
%

{\bf Results with Deep models on Cifar10}
\hspace{5pt}
We compare the performance of our BCSC against other state-of-the-art optimization methods including AdaGrad (AG), AdaDelta (AD), stochastic gradient descent (SGD), stochastic randomized block-coordinate descent (SBC), and randomized block-coordinate descent (RBC).
In this comparative analysis, we provide the learning curves and the accuracy table based on the network models including LeNet4, VGG19, ResNet18, GoogLeNet and DPN92 using the Cifar10 dataset. 
The experimental results for BCSC are obtained with $M$ = $8$, which is chosen as an example, but the results with other values for $M$ agree with the effectiveness and robustness of the number of parameter blocks as demonstrated by previous experiments.
The learning curves obtained by different optimization algorithms, SGD, SBC, RBC and BCSC, based on different network models, LeNet4, VGG19 and ResNet18, are presented in Fig.~\ref{fig:result_cifar10_basic} where BCSC outperforms all the other algorithms in accuracy, stability and convergence speed regardless of the network models.
For more extensive comparison, the learning curves are obtained by SGD and BCSC based on deeper network models, GoogLeNet and DPN92, using the Cifar10 dataset and they are presented in Fig.~\ref{fig:result_cifar10_deep} where the performance of BCSC is shown to be significantly better than SGD.
In addition to the comparison by the learning curve, we provide quantitative evaluation of the test accuracy  computed within (a) the first half epochs, (b) the last half epochs, (c) all the epochs and (d) the final epoch in Table~\ref{tab:accuracy_cifar10} and~\ref{tab:accuracy_cifar10_deep}.
These experimental results indicate that our BCSC algorithm outperforms all five  state-of-the-art optimization methods irrespective of the architecture and the depth of the models not only by the final accuracy, but also by the convergence speed.
%
%
%
%
%
%
\def\fw{68pt}
\begin{figure*}[htb] 
\centering
\begin{tabular}{c@{}c@{}c@{}c@{}c@{}c}
\includegraphics[height=\fw]{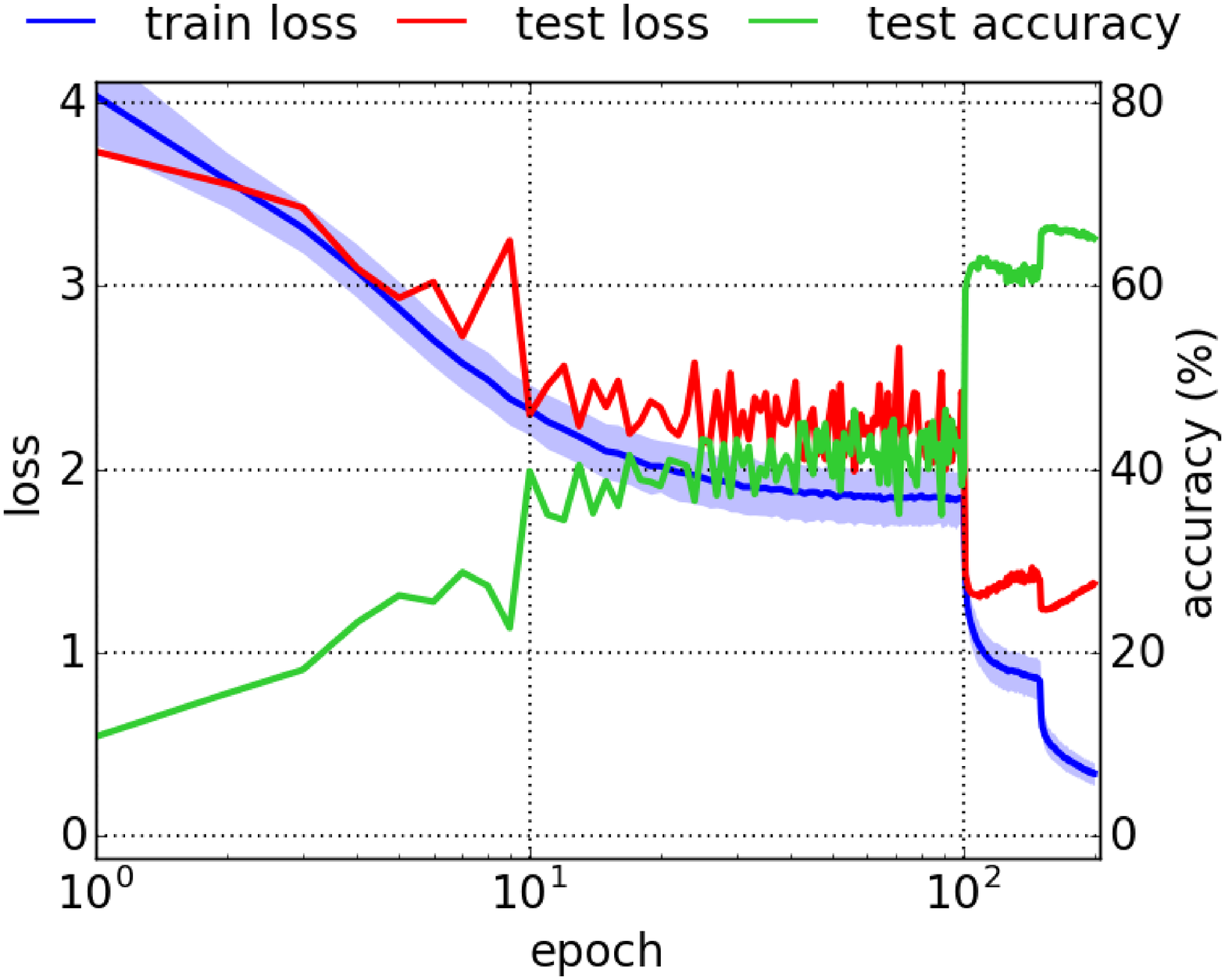} & 
\includegraphics[height=\fw]{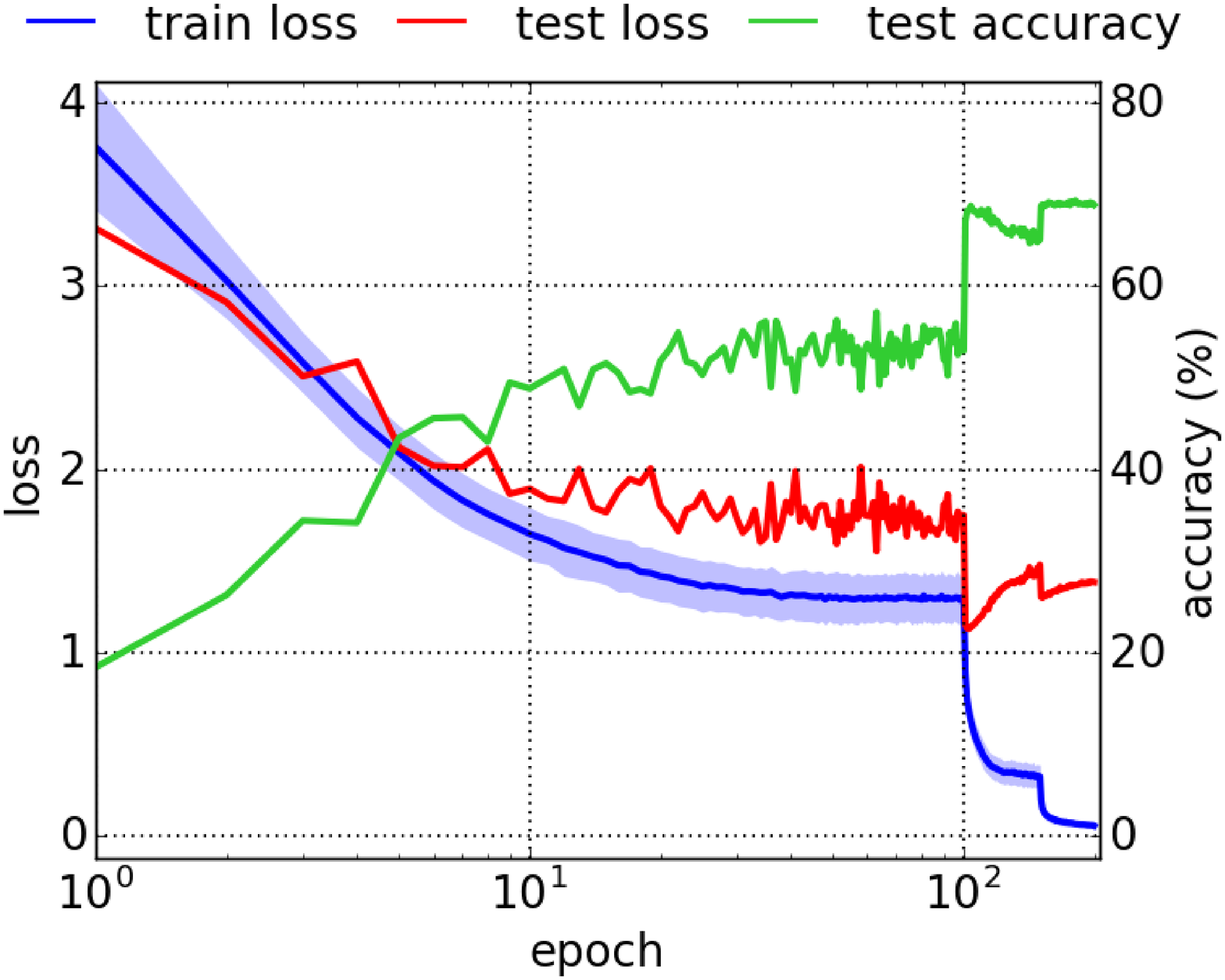} & 
\includegraphics[height=\fw]{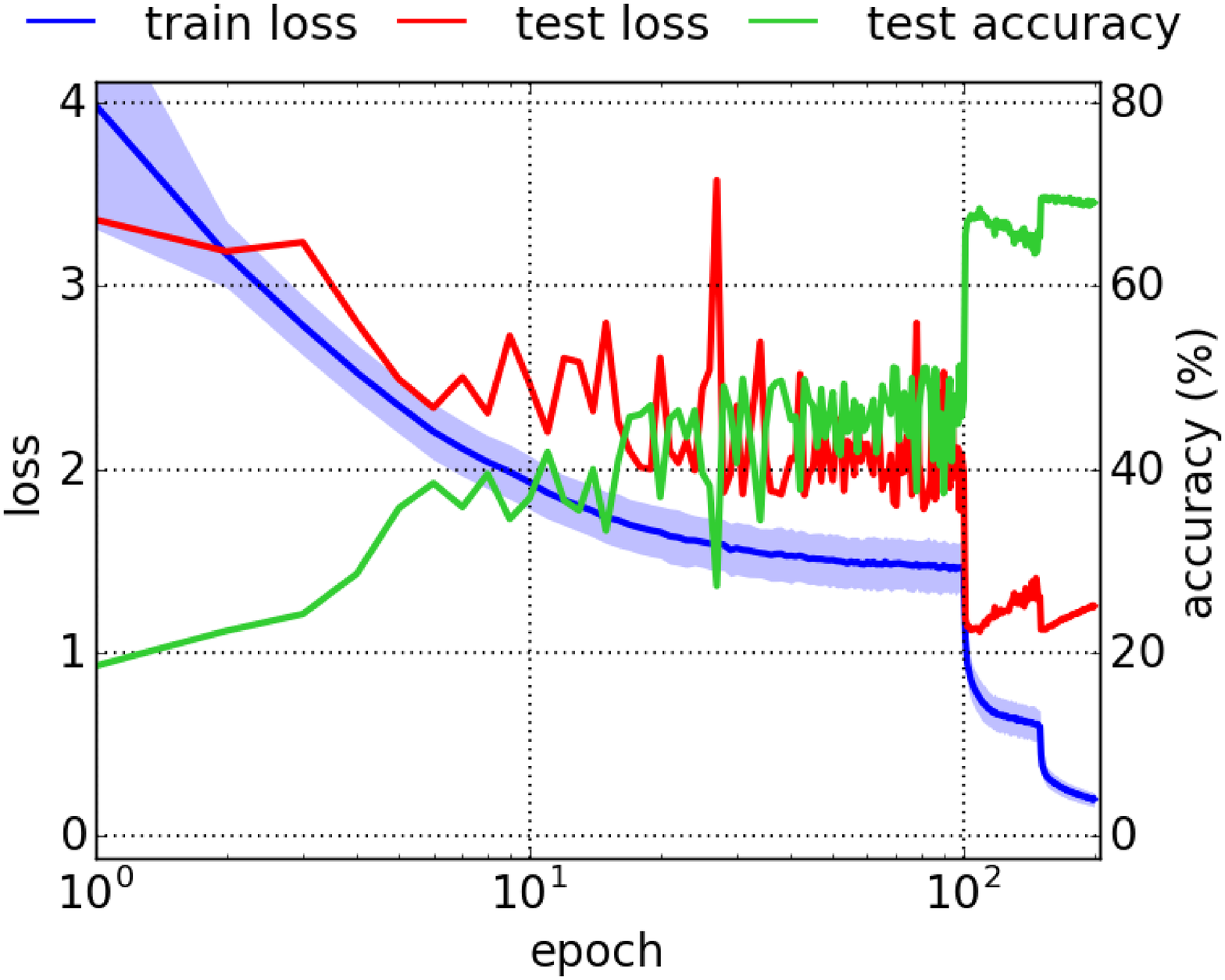} & 
\includegraphics[height=\fw]{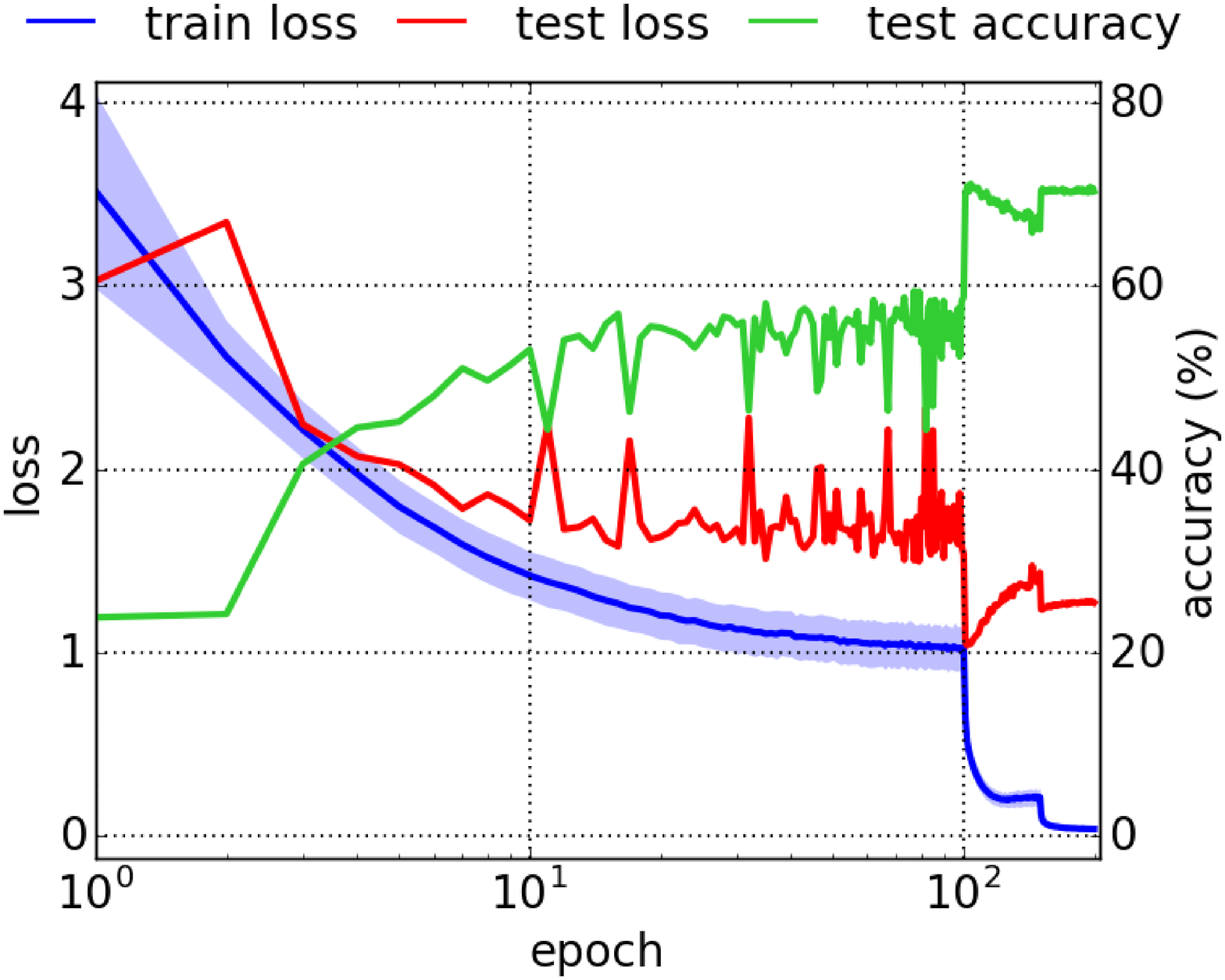} & 
\includegraphics[height=\fw]{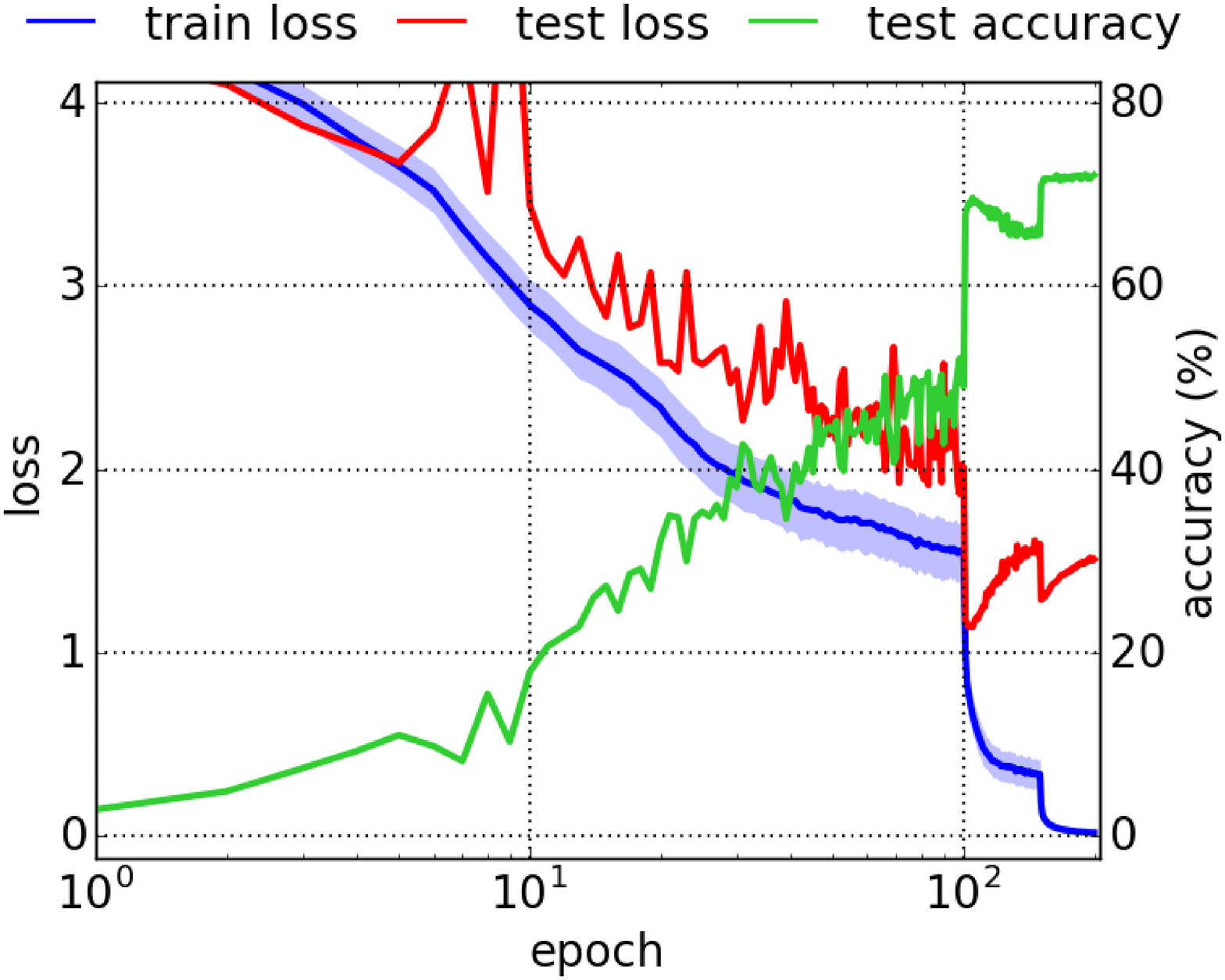} & 
\includegraphics[height=\fw]{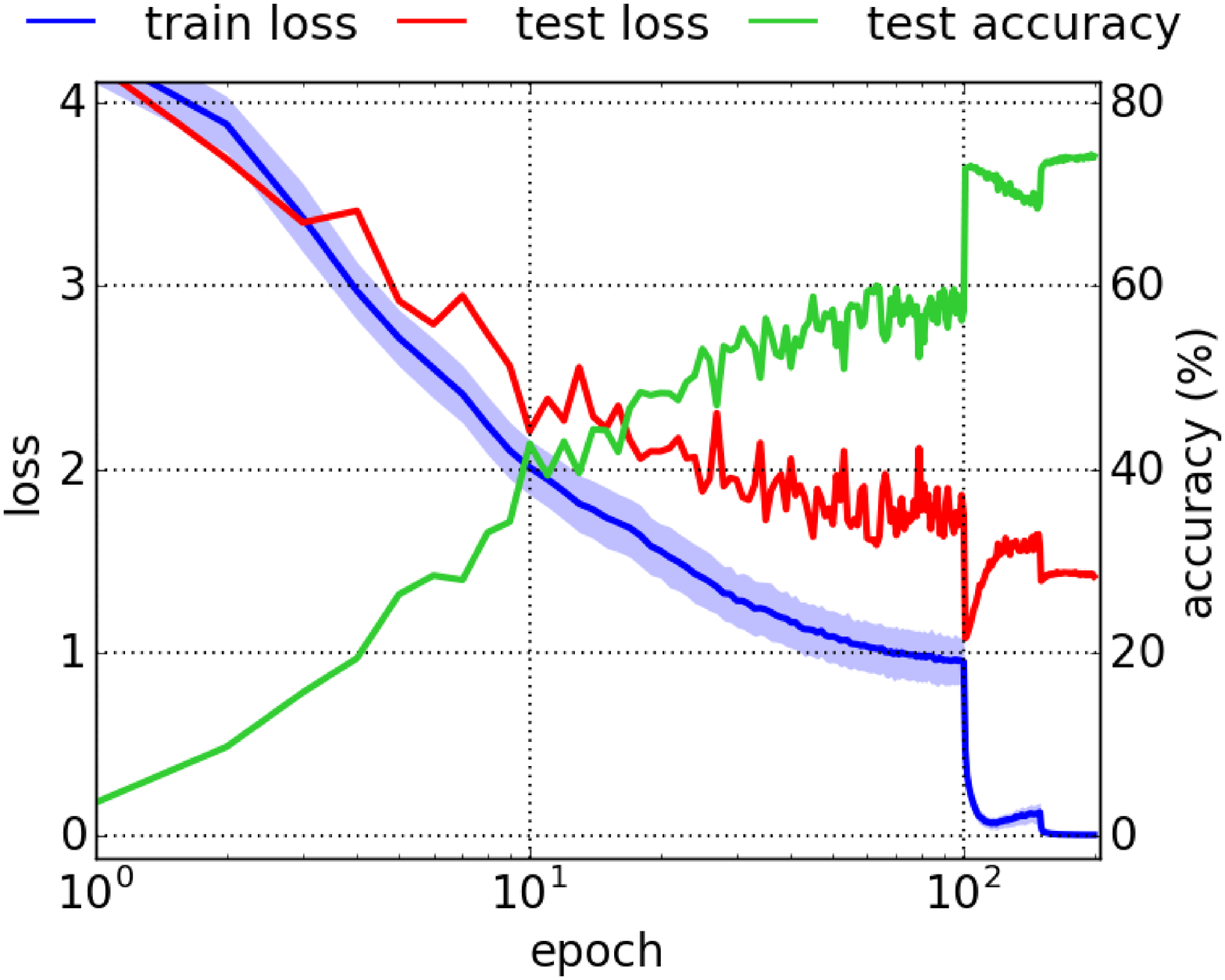} \\
SGD & BCSC & SGD & BCSC & SGD & BCSC\\
\multicolumn{2}{c}{(a) MobileNet~\cite{howard2017mobilenets}} &  \multicolumn{2}{c}{(b) ShuffleNet~\cite{zhang2017shufflenet}} &  \multicolumn{2}{c}{(c) VGG19~\cite{simonyan2014very}} \\
\\
\includegraphics[height=\fw]{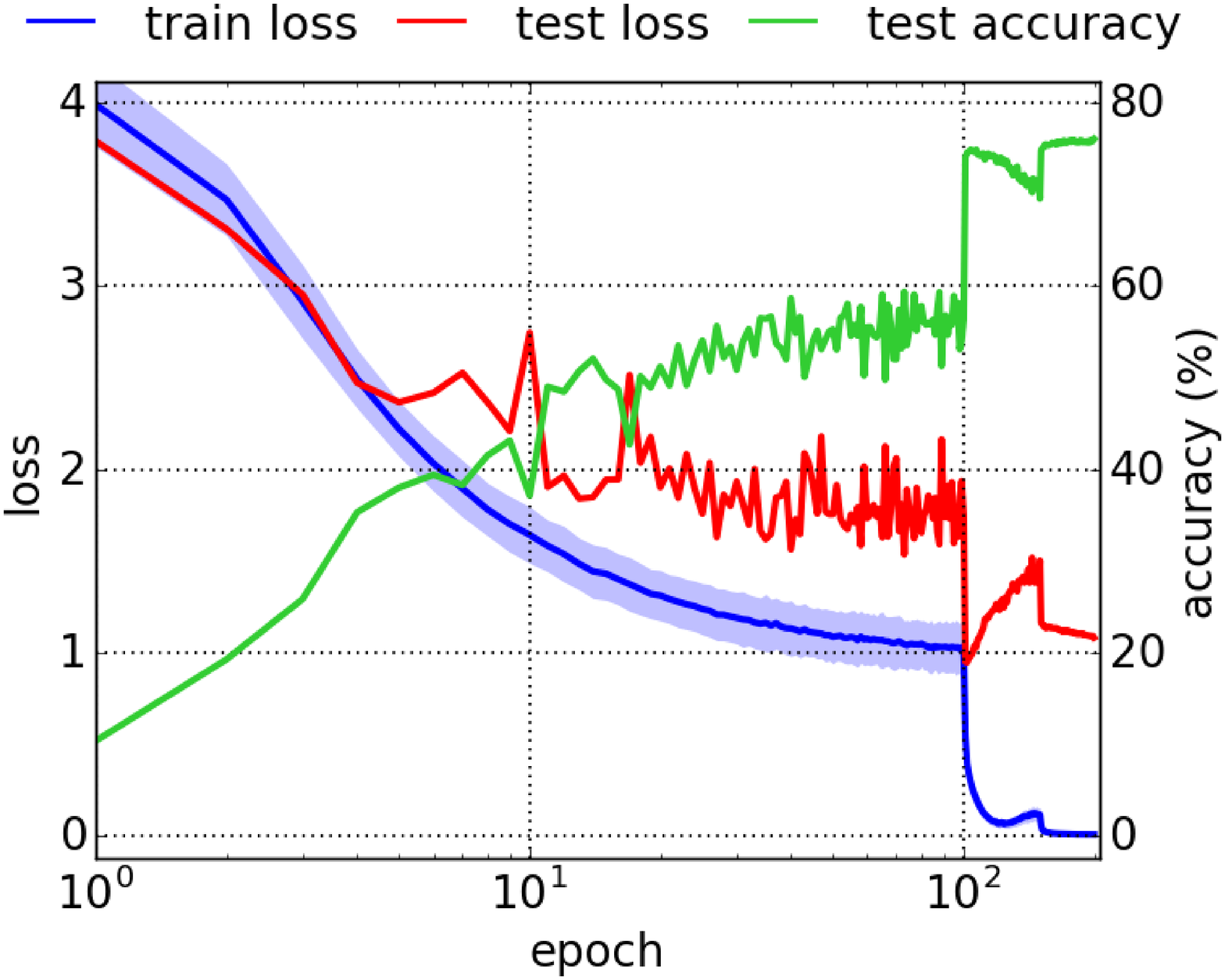} & 
\includegraphics[height=\fw]{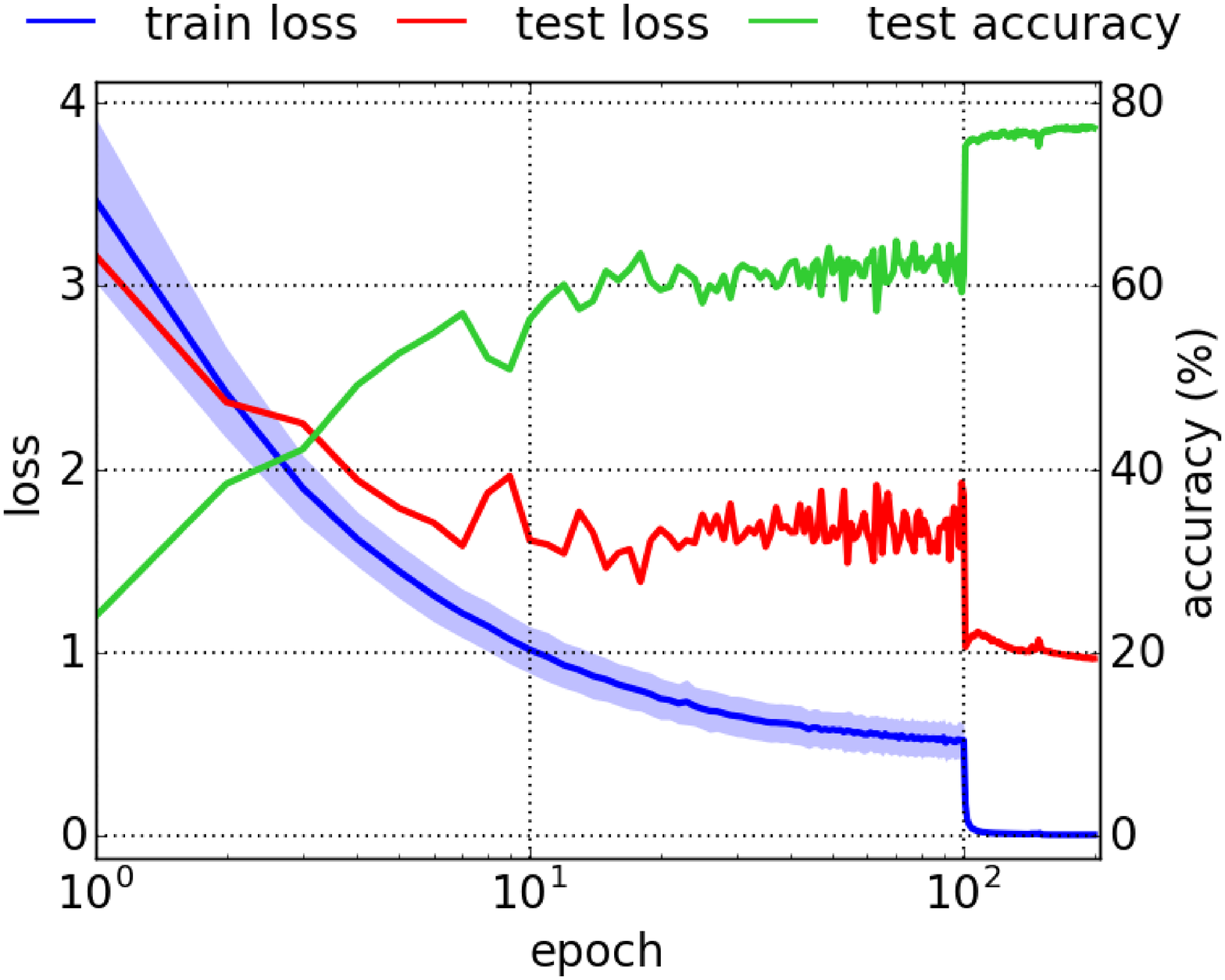} & 
\includegraphics[height=\fw]{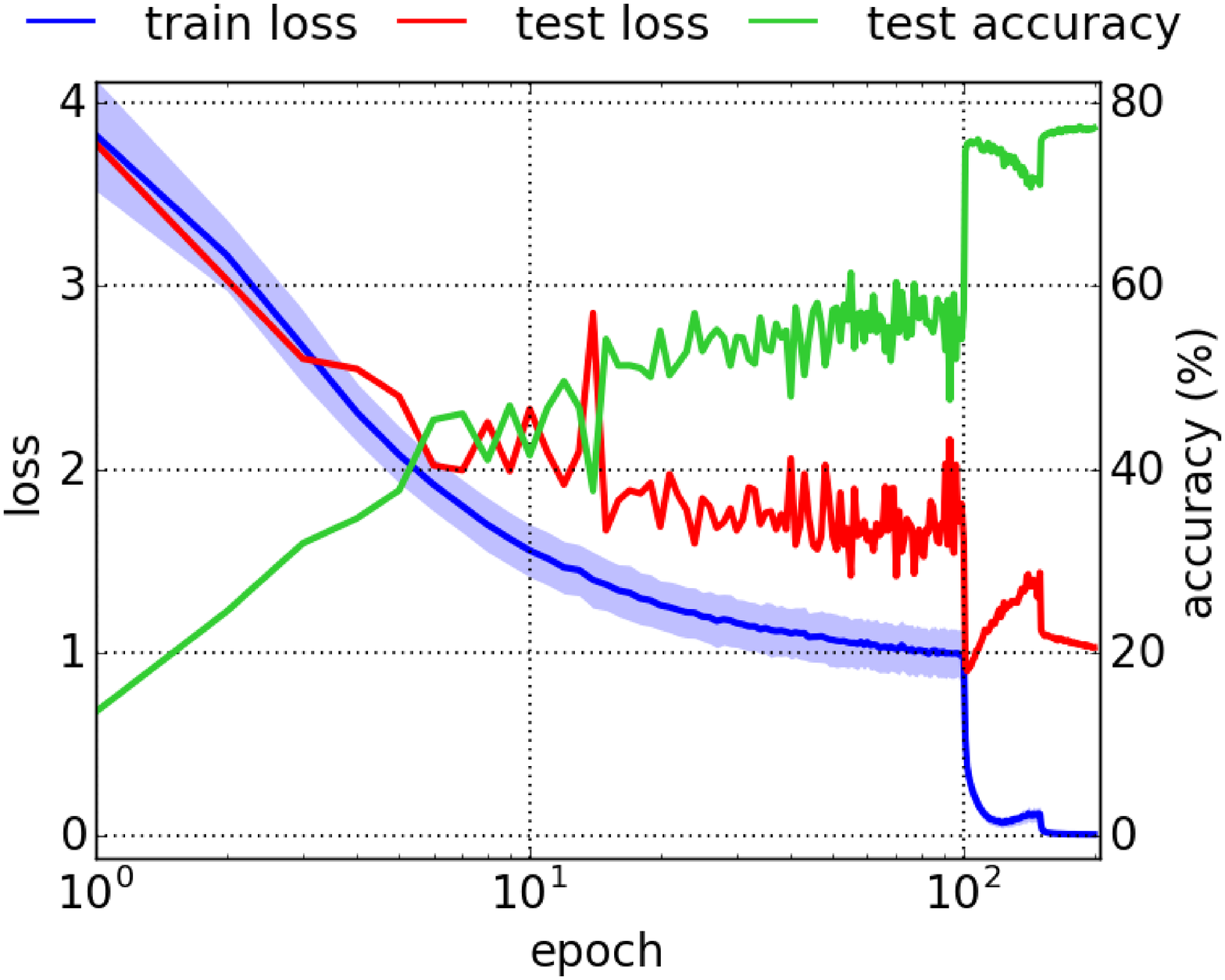} & 
\includegraphics[height=\fw]{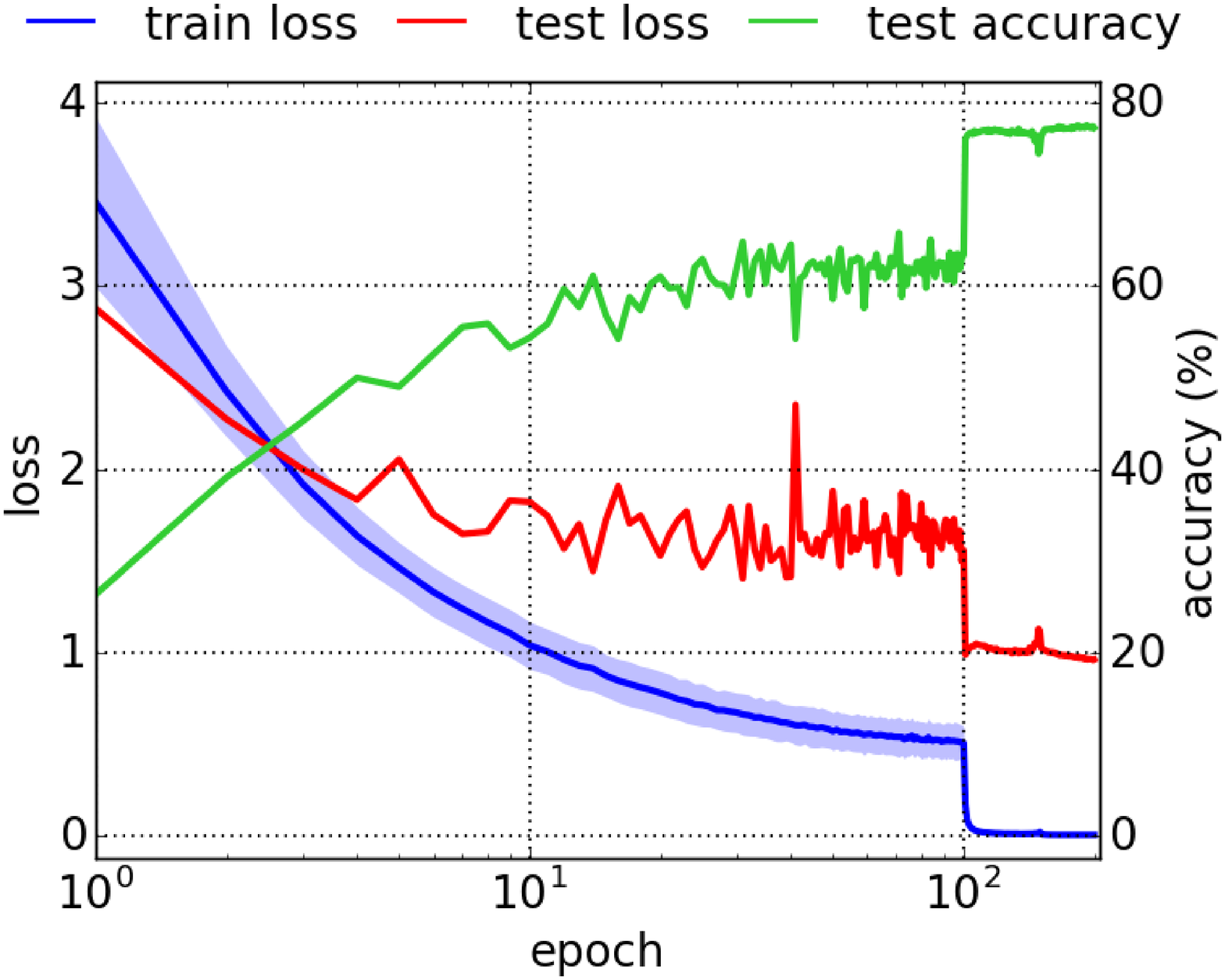} & 
\includegraphics[height=\fw]{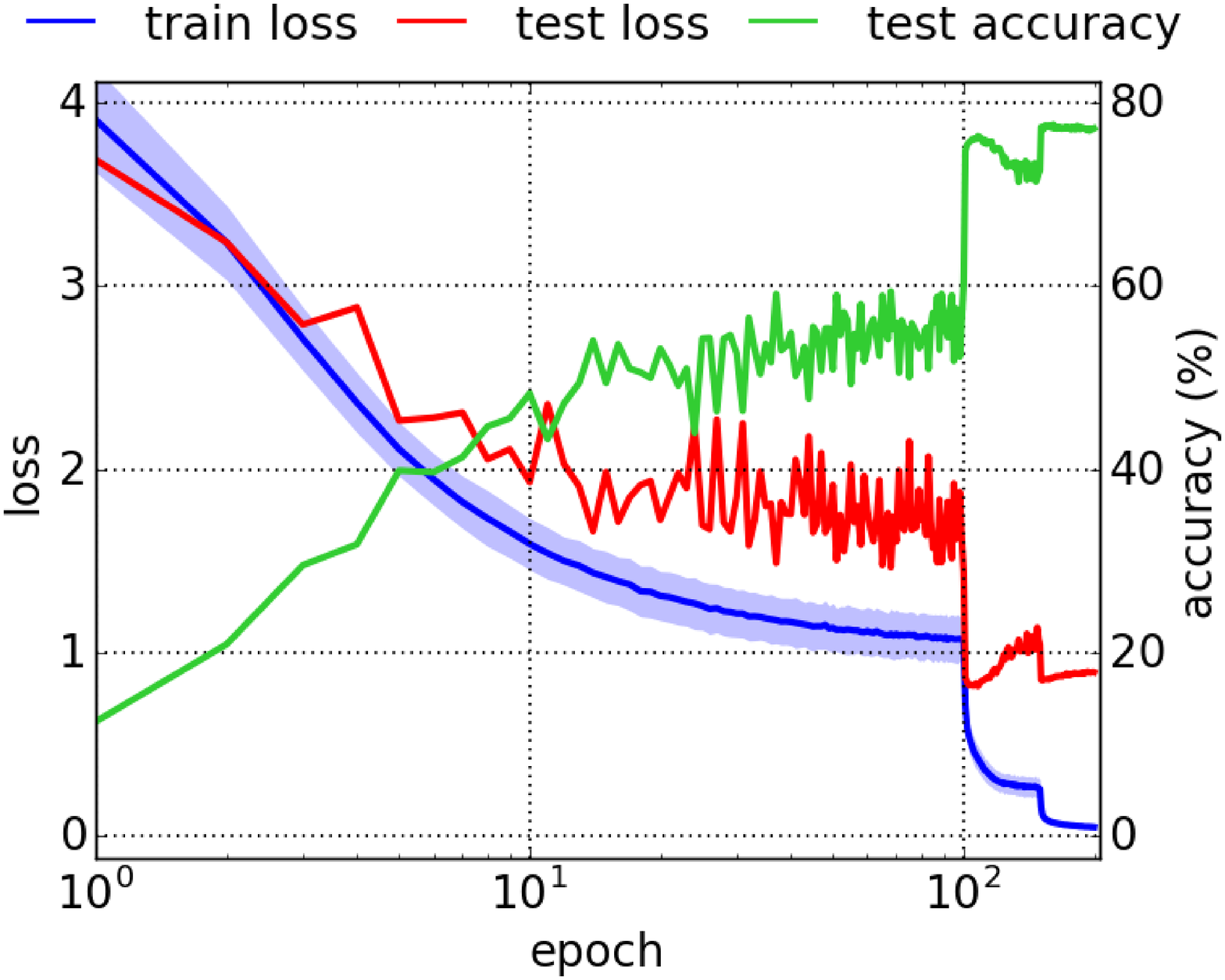} & 
\includegraphics[height=\fw]{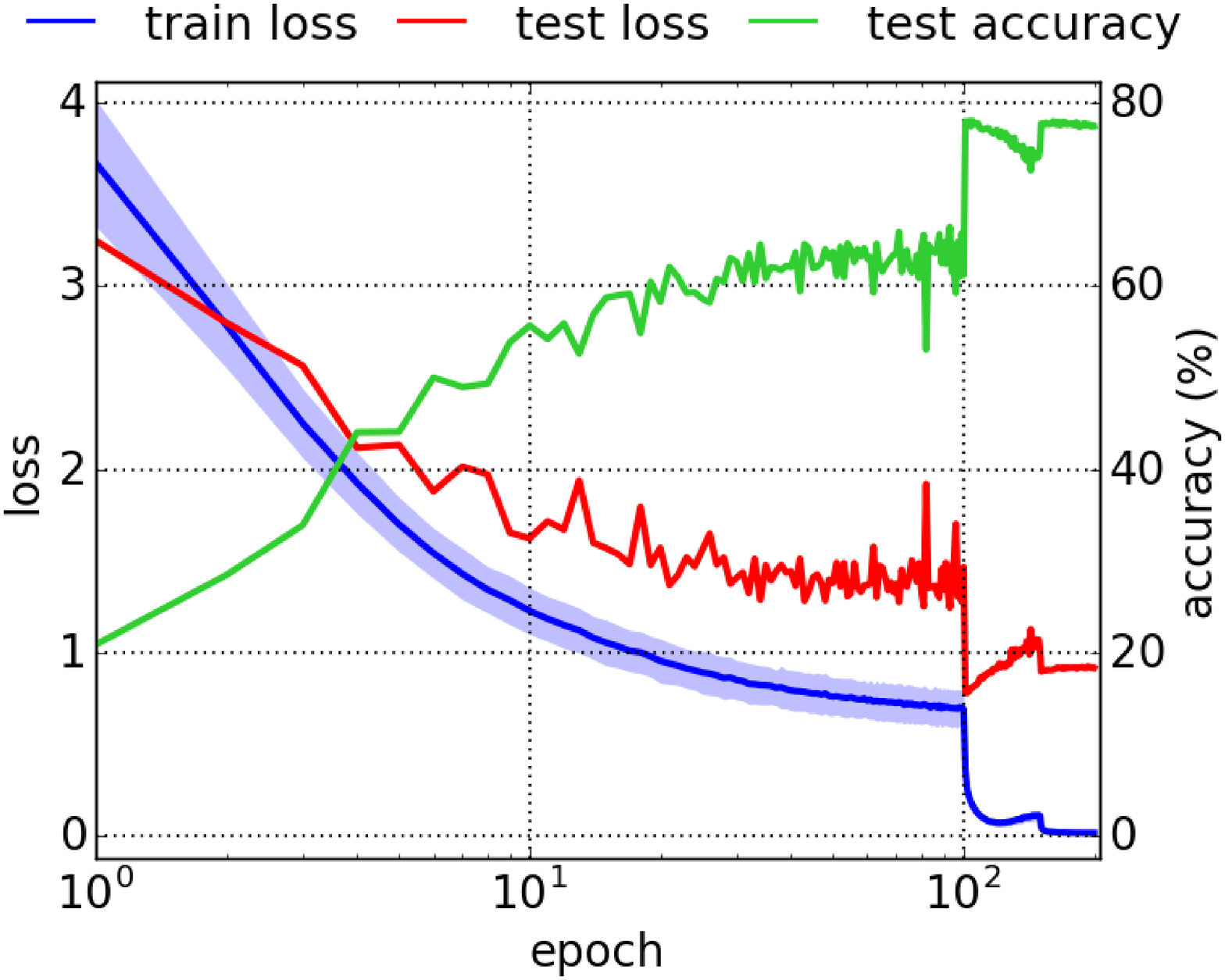} \\
SGD & BCSC & SGD & BCSC & SGD & BCSC\\
\multicolumn{2}{c}{(d) ResNet18~\cite{he2016deep, he2016identity}} &  \multicolumn{2}{c}{(e) SENet18~\cite{hu2017squeeze}} &  \multicolumn{2}{c}{(f) DenseConv~\cite{huang2017densely}} \\
\\
\includegraphics[height=\fw]{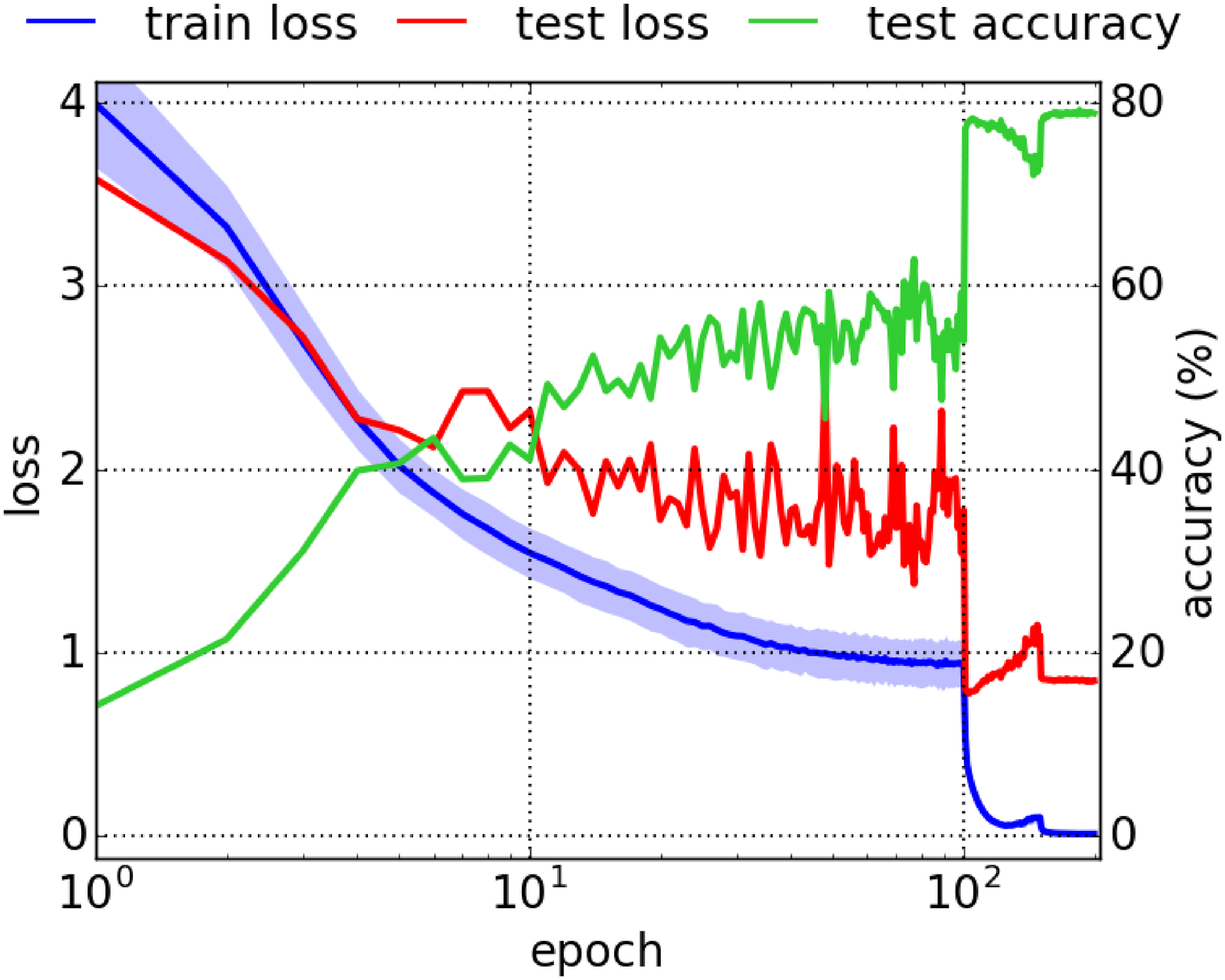} & 
\includegraphics[height=\fw]{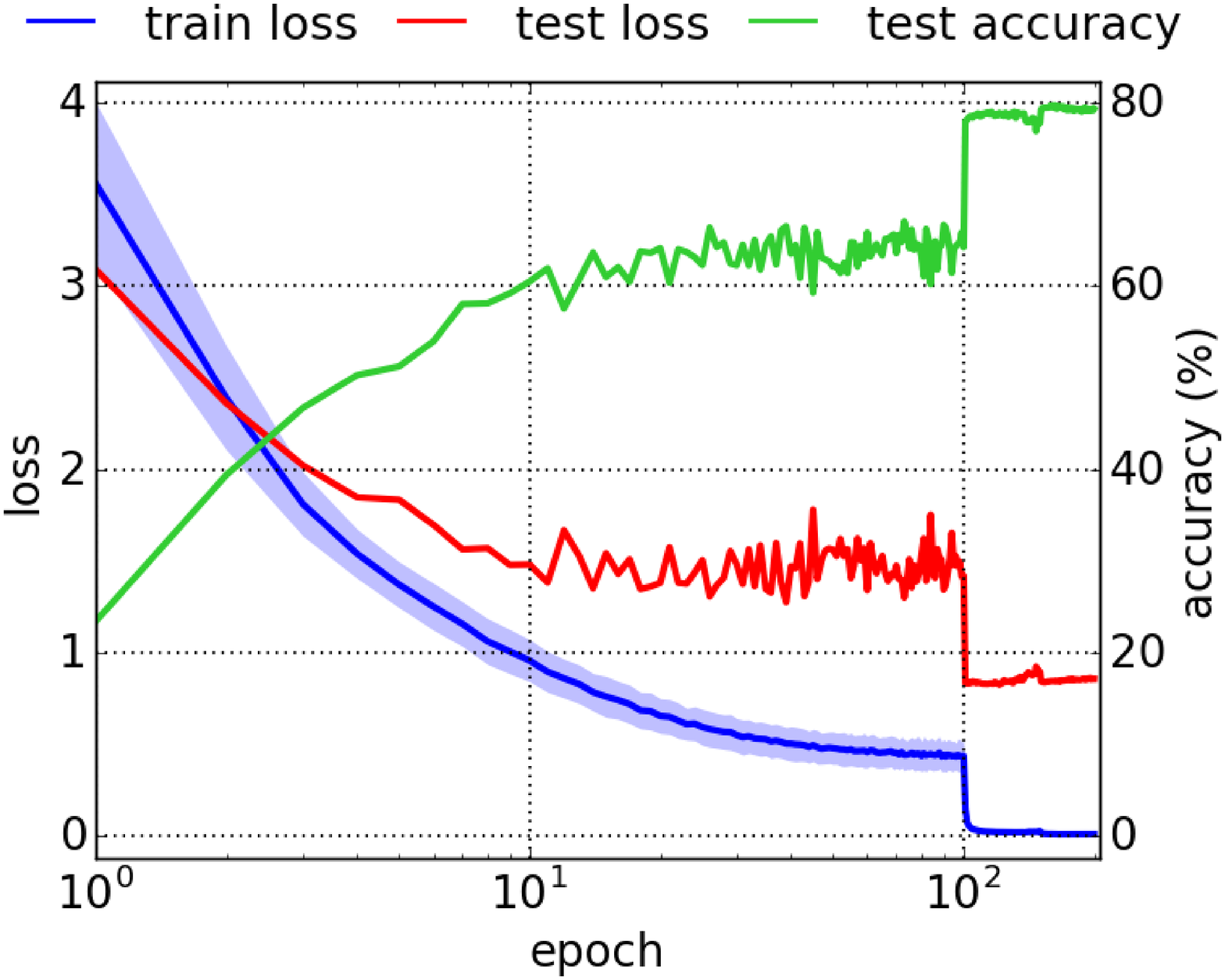} & 
\includegraphics[height=\fw]{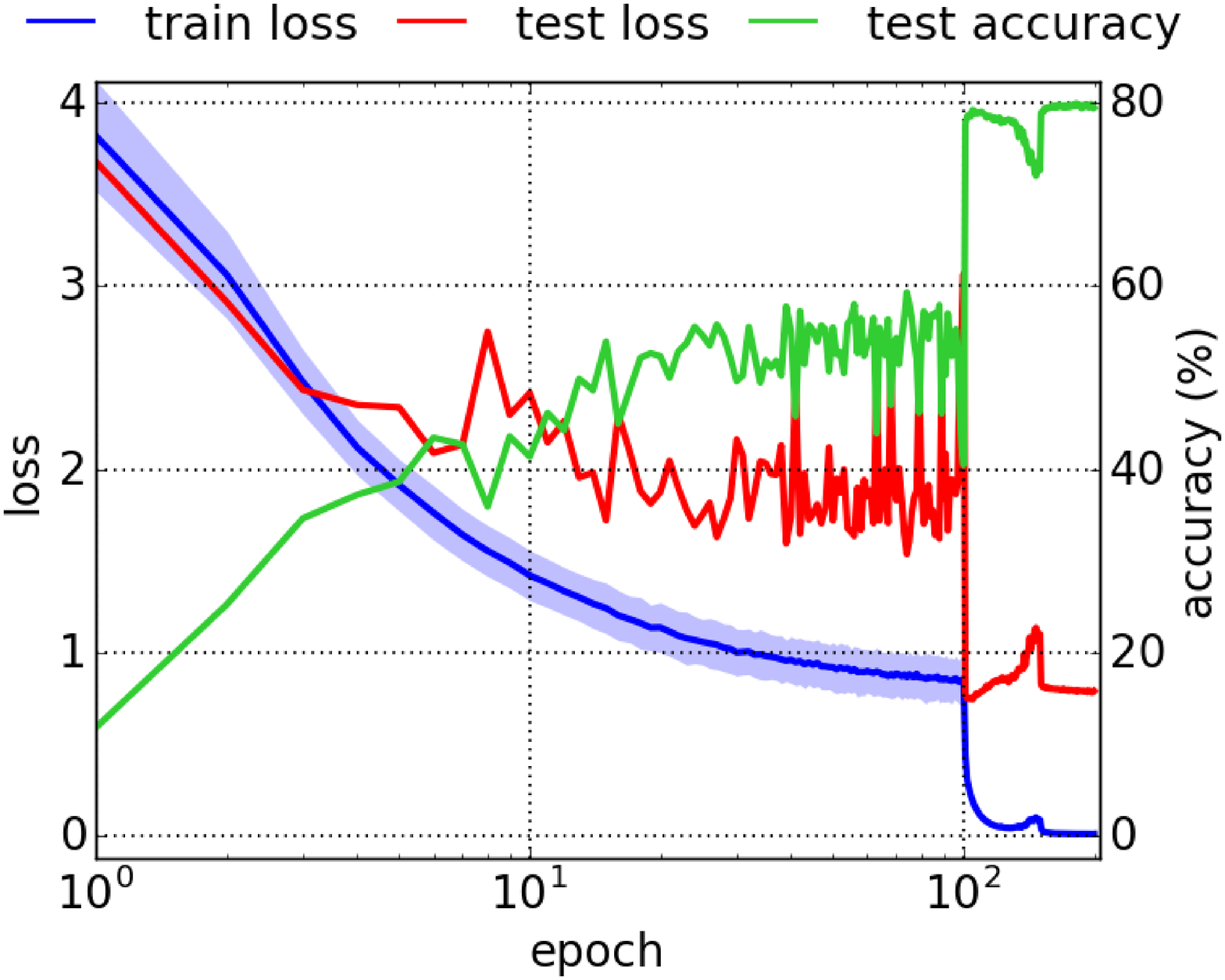} & 
\includegraphics[height=\fw]{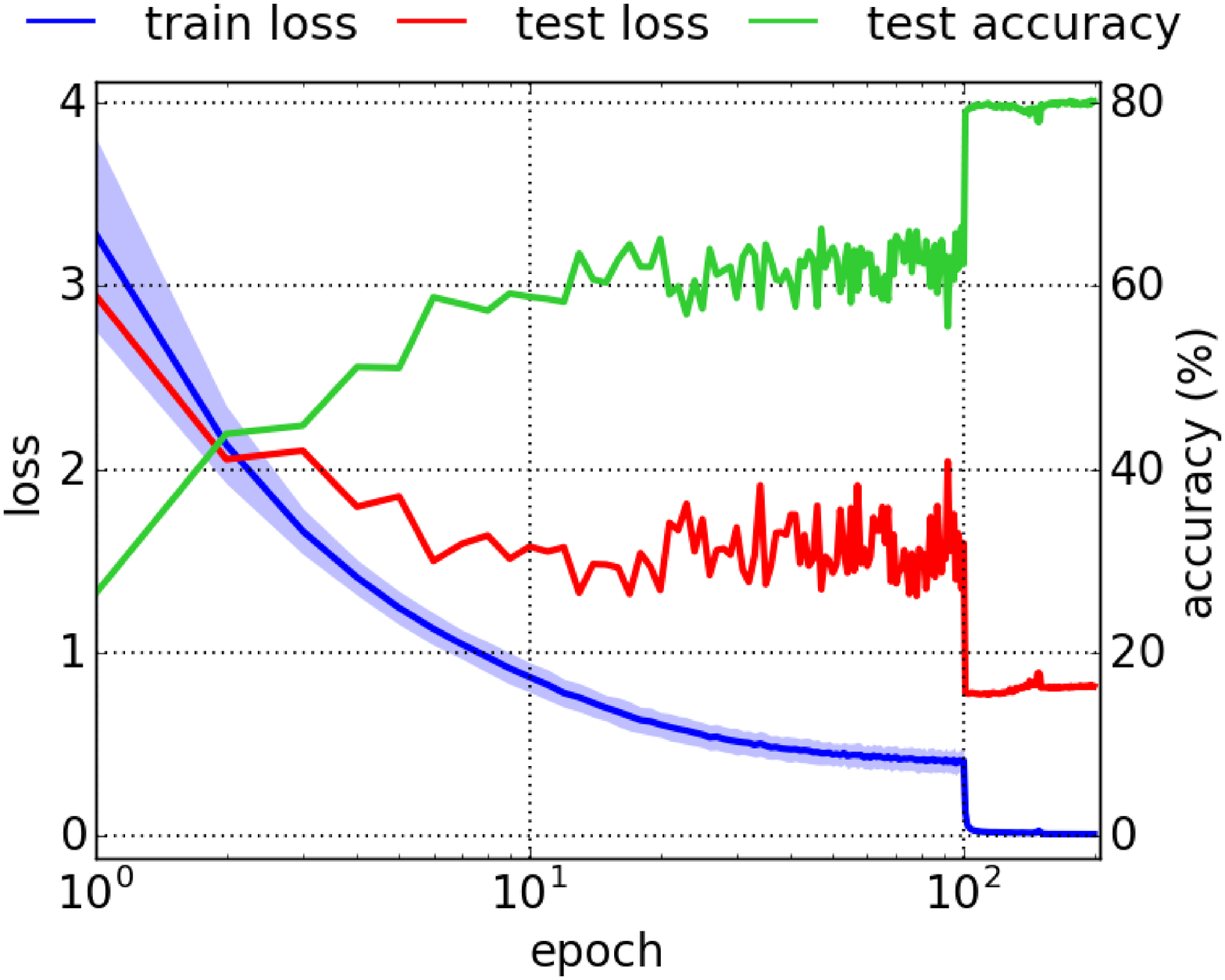} & 
\includegraphics[height=\fw]{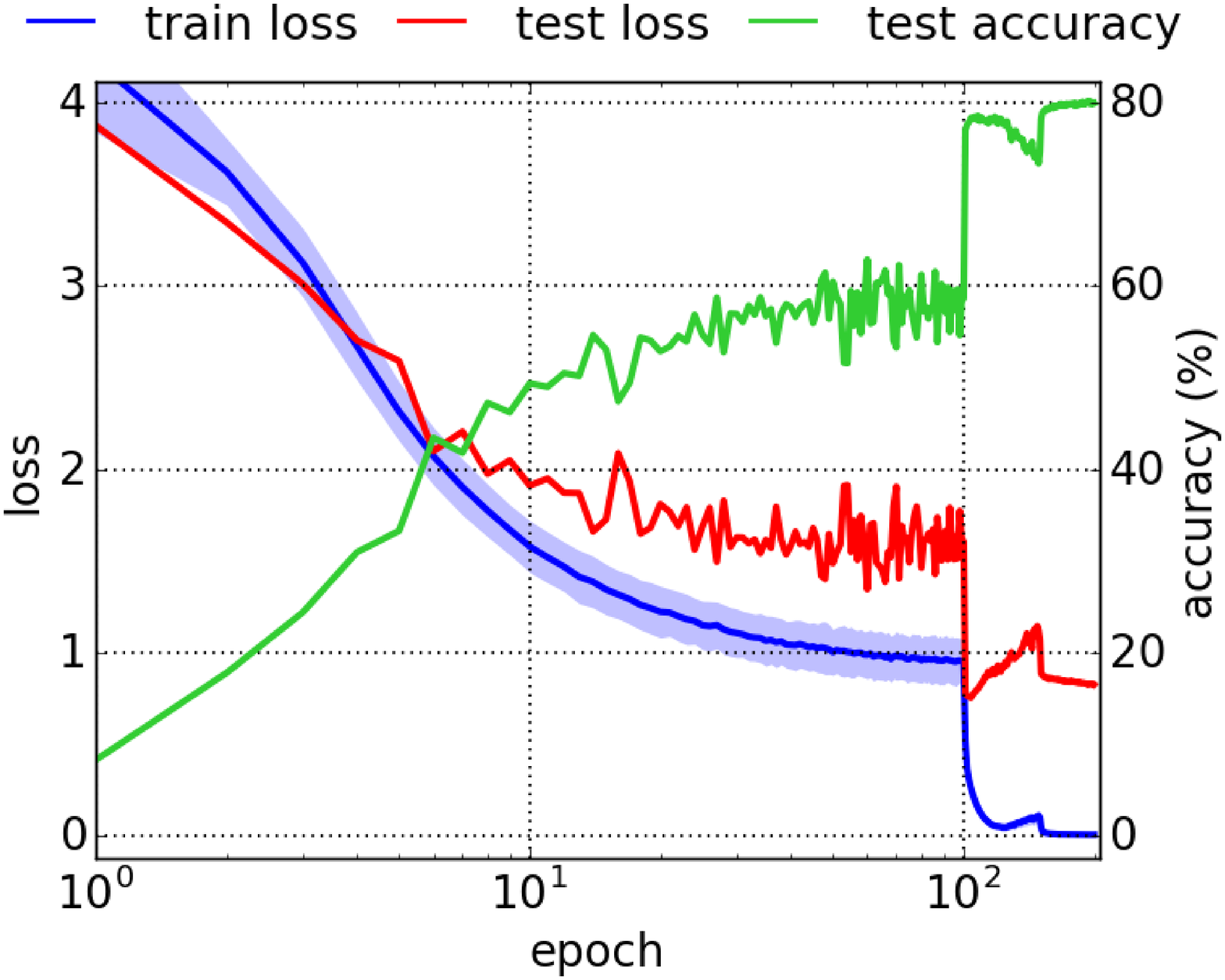} & 
\includegraphics[height=\fw]{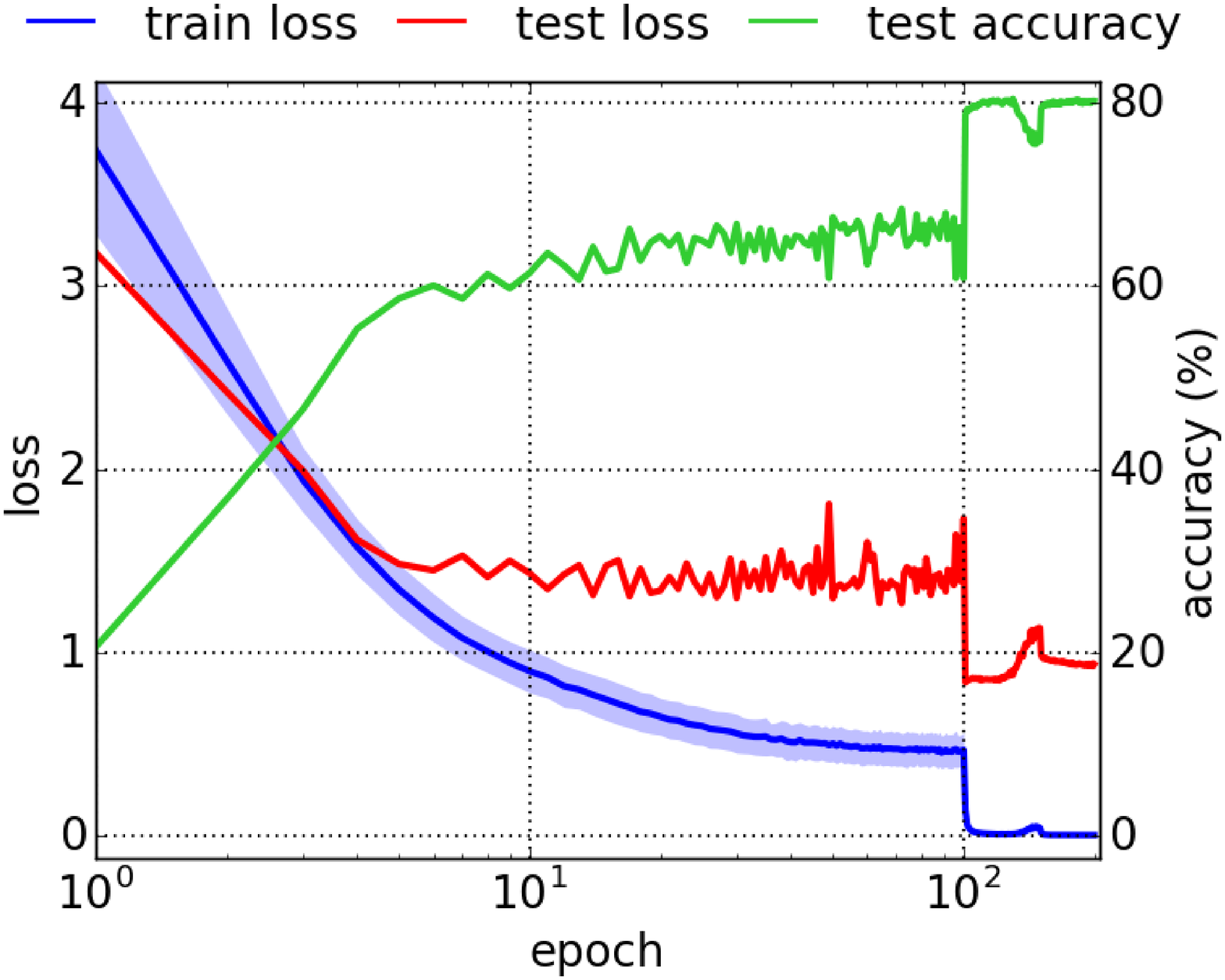} \\
SGD & BCSC & SGD & BCSC & SGD & BCSC\\
\multicolumn{2}{c}{(g) ResNeXt29~\cite{xie2016aggregated}} &  \multicolumn{2}{c}{(h) GoogLeNet~\cite{szegedy2015going}} &  \multicolumn{2}{c}{(i) DPN92~\cite{chen2017dual}}
\end{tabular}
\caption{{\bf Deep models on Cifar100} Learning curves optimized by SGD and BCSC with $M$ = $2$.}
\label{fig:result_cifar100}

\end{figure*}
%
%

\vspace{10pt}
{\bf Results with Deep models on Cifar100}
\hspace{5pt}
We now further validate the performance of BCSC in comparison with SGD using the more challenging Cifar100 based on the models including LeNet4, MobileNet, ShuffleNet, VGG19, ResNet18, SENet18, DenseConv, ResNeXt29, GoogLeNet, and DPN92.
In this experiment, we use $M$ = $2$ due to the heavy computational cost required to optimize deep models using the Cifar100 dataset. 
The learning curves are obtained by the algorithms, SGD and BCSC with $M$ = $2$, based on different network models, and they are presented in Fig.~\ref{fig:result_cifar100} where better, faster, and more stable results are observed with BCSC irrespective of the architecture albeit the minimum partition number is used.
The quantitative evaluation of BCSC in comparison to SGD is provided in Table~\ref{tab:accuracy_cifar100} where the testing accuracy is computed within (a) the first half epochs, (b) the last half epochs, (c) all the epochs and (d) the final epoch.
These experiments further confirm that BCSC outperforms standard SGD irrespective of the architecture of the models in accuracy, stability, and convergence speed.
It is also noted that the effectiveness of the algorithm can be demonstrated even with the minimum number of groupings in the model parameters. 
The performance of BCSC is consistently improved with increasing number of parameter blocks in particular with deep network models where the number of parameters is large.
\section{Discussion}  \label{sec:discussion}
We have presented a first-order optimization algorithm for large scale problems in deep learning when both the number of training data and the number of model parameters are large, and when the training data is polluted with outliers. 
The proposed algorithm, named BCSC, is based on the intuition that different subsets of data being used for updating different subsets of parameters is beneficial in handling outliers. 
The experimental results based on the state-of-the-art network models with the standard datasets indicate that the proposed dual stochastic process with the block-cyclic constraint leads to improved robustness to outliers in the training phase.
In addition, it has been empirically demonstrated that our algorithm outperforms the state-of-the-arts in optimizing a number of recent deep models in terms of accuracy, stability and convergence speed.
Our algorithm can be naturally extended to distributed and parallel computation, so as to mitigate the added computational cost due to the dual stochastic process.
Additional variants to the sampling and circulant schemes, as well as hyper-parameter tuning and determination of the optimal parameter-batch sizes, are also subject of future work. 
%

%
%
%
%
\begin{table}[htb] 
   \vspace{35pt}
	\centering
	\caption{Test accuracy of deep models for Cifar10 (\%)}
	{\footnotesize
	\setlength\tabcolsep{3pt} 
	\begin{tabular}{ l | c c | c c | c c | c c }
	\hline
	Epoch		& \multicolumn{2}{ c |}{(a) First half}& \multicolumn{2}{ c |}{(b) Last half}& \multicolumn{2}{ c |}{(c) All}& \multicolumn{2}{ c }{(d) Final}\\
	 			& SGD 	& BCSC 		& SGD 	& BCSC 		& SGD 	& BCSC 		& SGD 	& BCSC\\ 				
	\hline
	GoogLeNet	& 77.66	& {\bf 89.97}		& 94.04	& {\bf 95.56}		& 85.85	& {\bf 92.77}		& 94.78	& {\bf 95.61}\\
	DPN92		& 80.53	& {\bf 91.15}		& 94.50	& {\bf 95.24}		& 87.51	& {\bf 93.20}		& 95.38	& {\bf 95.46}\\
	\hline
	\end{tabular}
	} 
	\label{tab:accuracy_cifar10_deep}
   \vspace{35pt}
\end{table}
%
%

%
%
%
\begin{table}[htb] 
	\centering
	\caption{Test accuracy for Cifar100 (\%)}
	{\scriptsize
	\setlength\tabcolsep{3pt} 
	\begin{tabular}{ l | c c | c c | c c | c c }
	\hline
	Epoch		& \multicolumn{2}{ c |}{(a) First half}& \multicolumn{2}{ c |}{(b) Last half}& \multicolumn{2}{ c |}{(c) All}& \multicolumn{2}{ c }{(d) Final}\\
	 			& SGD 	& BCSC 				& SGD 	& BCSC 				& SGD 	& BCSC 				& SGD 	& BCSC\\ 				
	\hline
	LeNet4		& 15.16 & {\bf 22.94}		& 36.74	& {\bf 39.75}		& 25.95	& {\bf 31.35}		& 41.26	& {\bf 42.35}\\
	MobileNet	& 39.36	& {\bf 51.47}		& 63.80	& {\bf 67.79}		& 51.58	& {\bf 59.63}		& 65.21	& {\bf 68.88}\\
	ShuffleNet	& 43.57	& {\bf 53.75}		& 67.77	& {\bf 69.57}		& 55.67	& {\bf 61.66}		& 69.12	& {\bf 70.36}\\
	VGG19		& 38.47	& {\bf 51.48}		& 69.48	& {\bf 72.38}		& 53.98	& {\bf 61.93}		& 72.14	& {\bf 74.19}\\
	ResNet18	& 52.14	& {\bf 60.21}		& 74.35	& {\bf 76.79}		& 63.24	& {\bf 68.50}		& 76.08	& {\bf 77.28}\\
	SENet18		& 52.90	& {\bf 60.09}		& 75.38	& {\bf 76.98}		& 64.14	& {\bf 68.53}		& 77.28	& {\bf 77.30}\\
	DenseConv 	& 51.91	& {\bf 60.02}		& 75.68	& {\bf 76.84}		& 63.79	& {\bf 68.43}		& 77.22	& {\bf 77.46}\\
	ResNeXt29	& 52.65	& {\bf 62.39}		& 77.52	& {\bf 78.97}		& 65.09	& {\bf 70.68}		& 78.88	& {\bf 79.37}\\
	GoogLeNet	& 51.14	& {\bf 60.97}		& 78.33	& {\bf 79.68}		& 64.73	& {\bf 70.33}		&  79.51	& {\bf 80.20}\\
	DPN92		& 54.58	& {\bf 63.88}		& 78.30	& {\bf 79.48}		& 66.44	& {\bf 71.68}		& 79.98	& {\bf 80.23}\\
	\hline
	\end{tabular}
	}
	\label{tab:accuracy_cifar100}
\end{table}
%
%

\end{document}